\documentclass{article}

\PassOptionsToPackage{numbers, compress}{natbib}
\usepackage[final]{nips_2016}

\usepackage[utf8]{inputenc} % allow utf-8 input
\usepackage[T1]{fontenc}    % use 8-bit T1 fonts
\usepackage[bookmarks=false]{hyperref}       % hyperlinks
\usepackage{url}            % simple URL typesetting
\usepackage{booktabs}       % professional-quality tables
\usepackage{amsfonts}       % blackboard math symbols
\usepackage{nicefrac}       % compact symbols for 1/2, etc.
\usepackage{microtype}      % microtypography
\usepackage{natbib}

\usepackage{amsmath}
\usepackage{amsthm}
\usepackage{amssymb}
\usepackage{graphicx}
\usepackage{caption}
\usepackage{subcaption}
\usepackage{epstopdf}
\usepackage{wrapfig}
\usepackage{xcolor}
\usepackage{algorithmic}
\usepackage{algorithm}
\usepackage{a4wide} % sohau added this
\hypersetup{
    colorlinks,
    linkcolor={red!50!black},
    citecolor={blue!50!black},
    urlcolor={blue!80!black}
}

\title{A Locally Adaptive Normal Distribution}

\author{
Georgios Arvanitidis, Lars Kai Hansen and S{\o}ren Hauberg\\
Technical University of Denmark, Lyngby, Denmark\\
DTU Compute, Section of Cognitive Systems\\
\texttt{\{gear,lkai,sohau\}@dtu.dk}\\
}

\date{\vspace{-5ex}}

\newtheorem{definition}{Definition}
\newtheorem{remark}{Remark}

\DeclareMathOperator*{\argmin}{\arg\!\min}

\newcommand{\bs}[1]{\boldsymbol{#1}}
\renewcommand{\b}[1]{\mathbf{#1}}
\newcommand{\dif}[1]{\mathrm{#1}}
\newcommand{\grad}[1]{\nabla_{#1}}
\newcommand{\norm}[1]{\left\lVert#1\right\rVert}
\newcommand{\inner}[2]{\langle#1,#2\rangle}
\newcommand{\tangent}[1]{\mathcal{T}_{\boldsymbol{#1}}\M}
\newcommand{\abs}[1]{\left| #1 \right|}
\newcommand{\parder}[2]{\frac{\partial#1}{\partial#2}}

\newcommand{\Log}[2]{\text{Log}_{#1}(#2)}
\newcommand{\Exp}[2]{\text{Exp}_{#1}(#2)}

\def\t{\intercal}
\def\D{\mathcal{D}}
\def\R{\mathbb{R}}  % the real numbers symbol
\def\E{\mathbb{E}}  % the Expectation symbol
\def\N{\mathcal{N}} % the Normal distribution symbol
\def\M{\mathcal{M}} % the Manifold symbol
\def\Z{\mathcal{Z}} % define the Manifold symbol
\def\S{\mathcal{S}} % define the Manifold symbol
\def\C{\mathcal{C}} % define the Manifold symbol
\definecolor{minColOne}{rgb}{0, 0.4470, 0.7410}
\definecolor{minColTwo}{rgb}{0.8500, 0.3250, 0.0980}
\definecolor{minColThree}{rgb}{0.4660, 0.6740, 0.1880}

\begin{document}

\maketitle

\begin{abstract}
The multivariate normal density is a monotonic function of the distance to the mean, and its ellipsoidal shape is due to the underlying Euclidean metric. We suggest to replace this metric with a locally adaptive, smoothly changing (Riemannian) metric that favors regions of high local density. The resulting \emph{locally adaptive normal distribution (LAND)} is a generalization of the normal distribution to the ``manifold'' setting, where data is assumed to lie near a potentially low-dimensional manifold embedded in $\mathbb{R}^D$. The LAND is parametric, depending only on a mean and a covariance, and is the maximum entropy distribution under the given metric. The underlying metric is, however, non-parametric. We develop a maximum likelihood algorithm to infer the distribution parameters that relies on a combination of gradient descent and Monte Carlo integration. We further extend the LAND to mixture models, and provide the corresponding EM algorithm. We demonstrate the efficiency of the LAND to fit non-trivial probability distributions over both synthetic data, and EEG measurements of human sleep.
\end{abstract}

\section{Introduction}\label{sec:intro}
The multivariate normal distribution is a fundamental building block in many machine learning algorithms, and its well-known density can compactly be written as
\begin{equation}
  \label{eq:normal_pdf_general}
  p(\b{x} ~|~ \bs{\mu}, \bs{\bs{\Sigma}}) \propto \exp\left(-\frac{1}{2} \text{dist}^2_{\bs{\Sigma}}(\bs{\mu},\b{x}) \right),
\end{equation}
where $\text{dist}^2_{\bs{\Sigma}}(\bs{\mu},\b{x})$ denotes the Mahalanobis distance for covariance matrix $\bs{\Sigma}$. This distance measure corresponds to the length of the straight line connecting $\bs{\mu}$ and $\b{x}$, and consequently the normal distribution is often used to model \emph{linear} phenomena. When data lies near a nonlinear manifold embedded in $\R^D$ the normal distribution becomes inadequate due to its linear metric. We investigate if a useful distribution can be constructed by replacing the linear distance function with a nonlinear counterpart. This is similar in spirit to Isomap \cite{tenenbaum:global:2000} that famously replace the linear distance with a geodesic distance measured over a neighborhood graph spanned by the data, thereby allowing for a nonlinear model. This is, however, a discrete distance measure that is only well-defined over the training data. For a generative model, we need a continuously defined metric over the entire $\R^D$.

Following \citet{hauberg:nips:2012} we learn a smoothly changing metric that favors regions of high density i.e., geodesics tend to move near the data. Under this metric, the data space is interpreted as a $D$-dimensional Riemannian manifold. This ``manifold learning'' does not change dimensionality, but merely provides a local description of the data. The Riemannian view-point, however, gives a strong mathematical foundation upon which the proposed distribution can be developed. Our work, thus, bridges work on statistics on Riemannian manifolds \cite{Pennec:JMIV:06,Miaomiao:nips:2013} with manifold learning \cite{tenenbaum:global:2000}.

\begin{figure}[!ht]
  \centering
  \begin{subfigure}[b]{0.31\textwidth}
    \includegraphics[width=\textwidth]{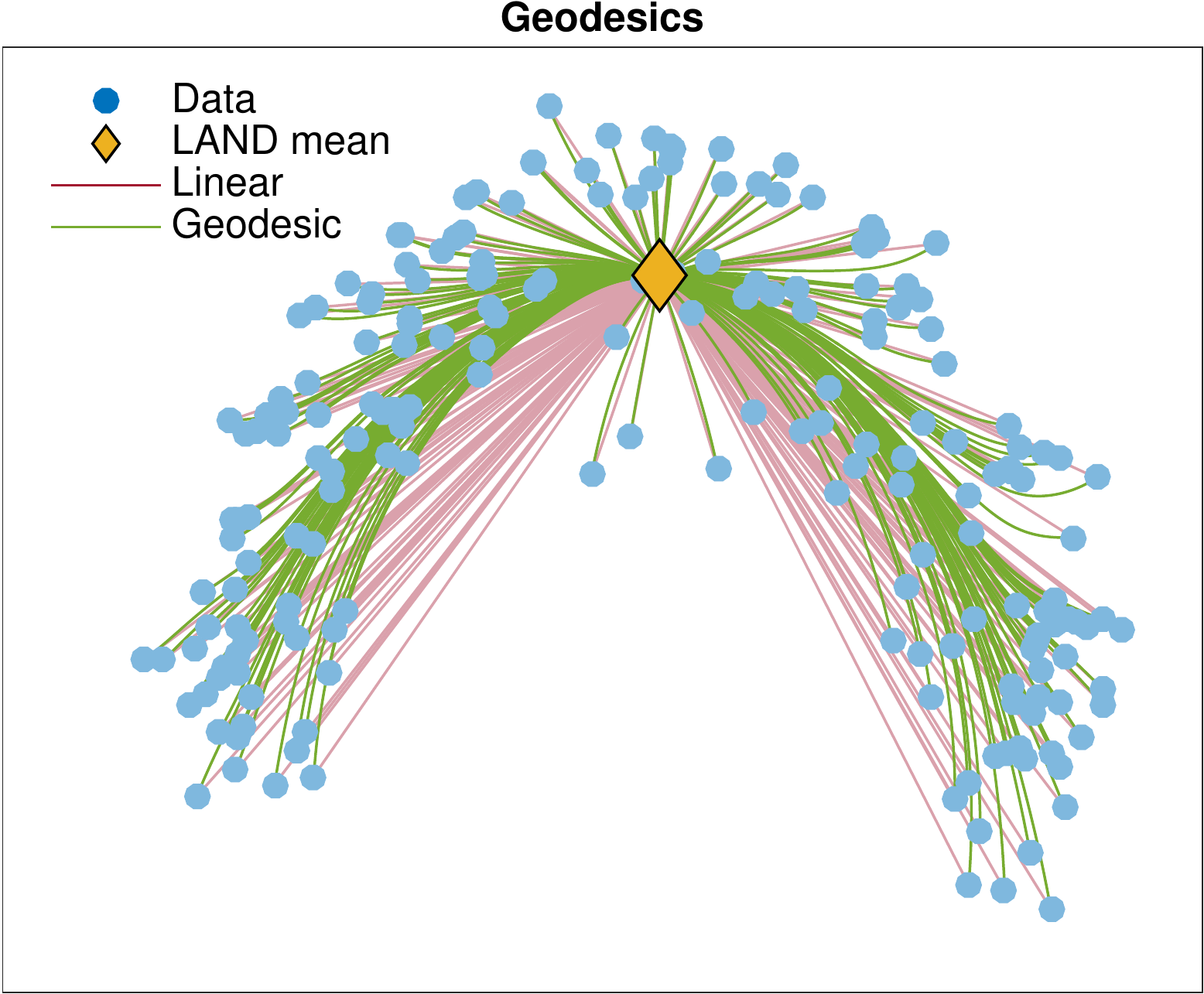}
  \end{subfigure}
  \quad
  \begin{subfigure}[b]{0.31\textwidth}
    \includegraphics[width=\textwidth]{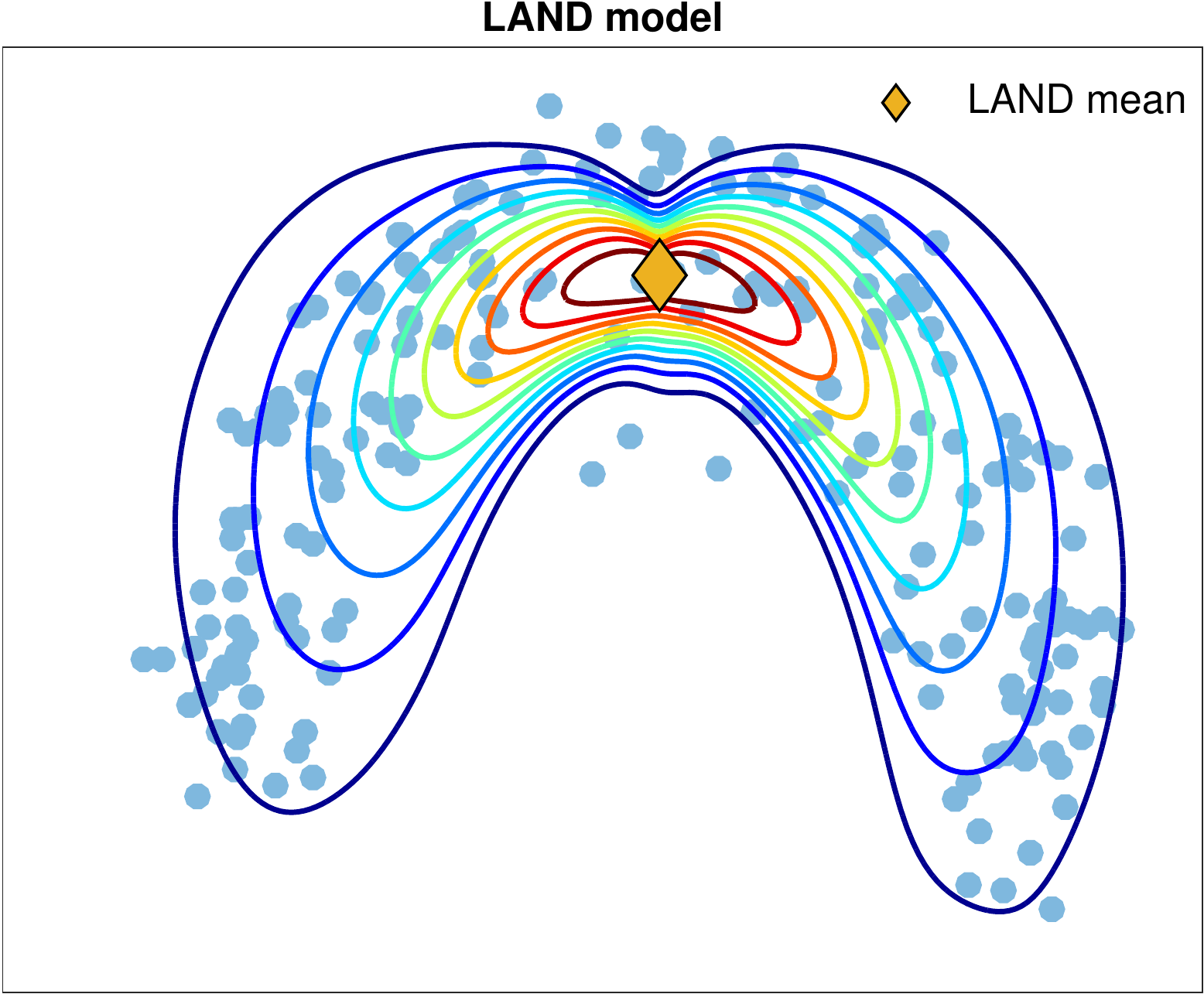}
  \end{subfigure}
  \quad
  \begin{subfigure}[b]{0.31\textwidth}
    \includegraphics[width=\textwidth]{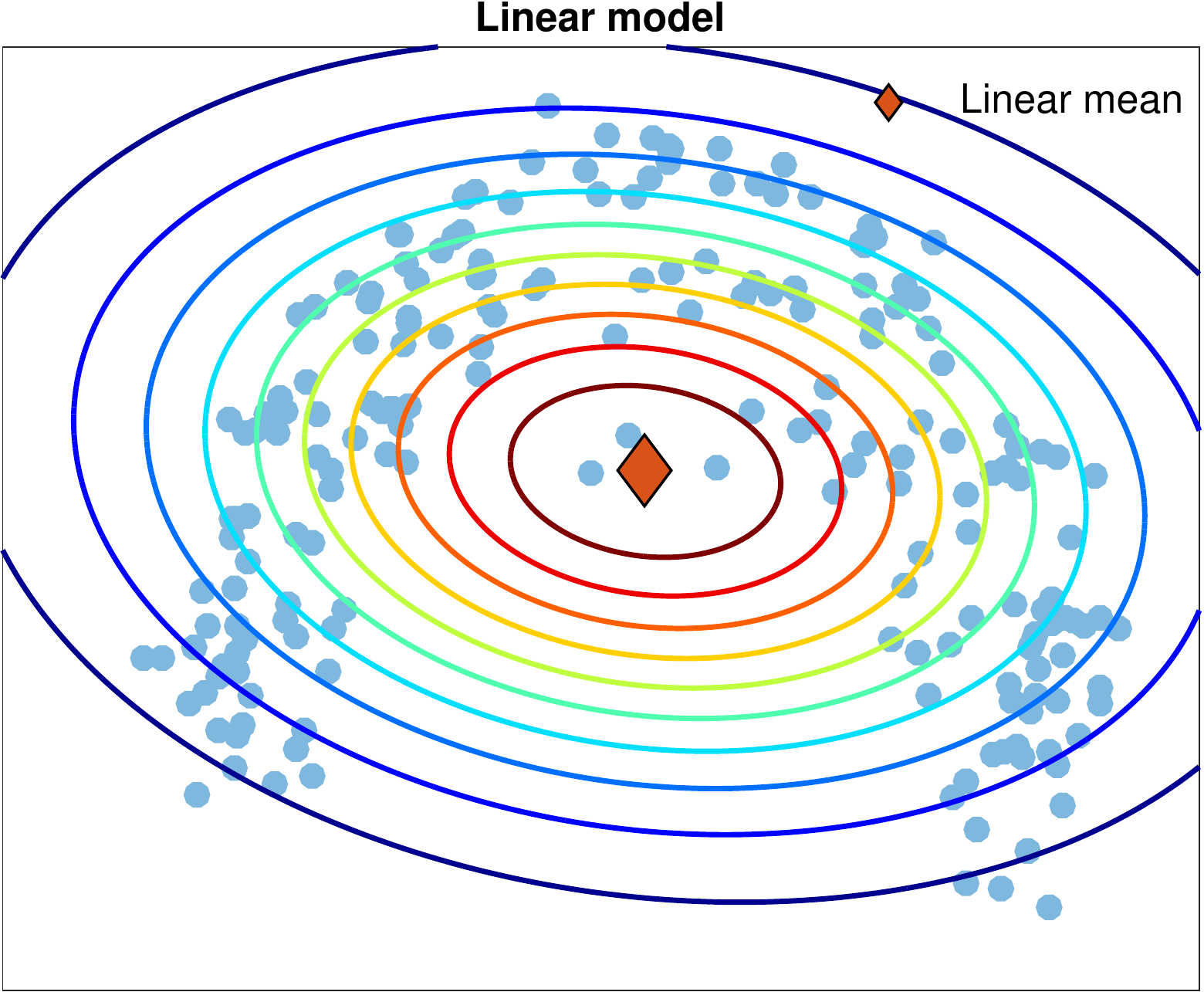}
  \end{subfigure}
    
  \caption{Illustration of the LAND using MNIST images of the digit 1 projected onto the first 2 principal components.  \textit{Left}: comparison of the geodesic and the linear distance. \textit{Center}: the proposed locally adaptive normal distribution. \textit{Right}: the Euclidean normal distribution.}
\end{figure}

We develop a \emph{locally adaptive normal distribution (LAND)} as follows: First, we construct a metric that captures the nonlinear structure of the data and enables us to compute geodesics; from this, an 
\newline

unnormalized density is trivially defined. Second, we propose a scalable Monte Carlo integration scheme for normalizing the density with respect to the measure induced by the metric. Third, we develop a gradient-based algorithm for maximum likelihood estimation on the learned manifold.
We further consider a mixture of LANDs and provide the corresponding EM algorithm.
The usefulness of the model is verified on both synthetic data and EEG measurements of human sleep stages.

\textbf{Notation}: all points $\b{x}\in\R^D$ are considered as column vectors, and they are denoted with bold lowercase characters. $\S_{++}^D$ represents the set of symmetric $D\times D$ positive definite matrices. The learned Riemannian manifold is denoted $\M$, and its tangent space at $\b{x}\in\M$ is denoted $\tangent{\b{x}}$. 

\section{A Brief Summary of Riemannian Geometry}\label{sec:geometry}
We start our exposition with a brief review of \emph{Riemannian manifolds} \cite{docarmo:1992}. These smooth manifolds are naturally equipped with a distance measure, and are commonly used to model physical phenomena such as dynamical or periodic systems, and many problems that have a smooth behavior. 
%We focus on the Riemannian manifolds \cite{docarmo:1992}, since they are naturally equipped with a distance measure.

\begin{definition}
A smooth manifold $\M$ together with a Riemannian metric $\b{M}:\M \rightarrow \mathcal{S}_{++}^D$ is called a Riemannian manifold. 
The Riemannian metric $\b{M}$ encodes a smoothly changing inner product $\inner{\b{u}}{\b{M}(\b{x})\b{v}}$ on the tangent space $\b{u},\b{v} \in \tangent{\b{x}}$ of each point $\b{x}\in\M$. 
\end{definition}

\begin{remark}
The Riemannian metric $\b{M}(\b{x})$ acts on tangent vectors, and may, thus, be interpreted as a standard Mahalanobis metric restricted to an infinitesimal region around $\b{x}$.
\end{remark}

The local inner product based on $\b{M}$ is a suitable model for capturing local behavior of data, i.e.\ \emph{manifold learning}.
From the inner product, we can define \emph{geodesics} as length-minimizing curves connecting two points $\b{x},\b{y}\in\M$, i.e.
\begin{align}
  \hat{\bs{\gamma}} %= \argmin_{\bs{\gamma}}(\text{length}(\bs{\gamma}))  
               = \argmin_{\bs{\gamma}} \int_{0}^{1} \sqrt{ \inner{\bs{\gamma}'(t)}{\b{M}(\bs{\gamma}(t))\bs{\gamma}'(t)}}\dif{d}t, \quad \text{s.t.} \quad \bs{\gamma}(0) = \b{x},~\bs{\gamma}(1) = \b{y}.
\end{align}
Here $\b{M}(\bs{\gamma}(t))$ is the metric tensor at $\bs{\gamma}(t)$, and the tangent vector $\bs{\gamma}'$ denotes the derivative (velocity) of $\bs{\gamma}$. The distance between $\b{x}$ and $\b{y}$ is defined as the length of the geodesic. A standard result from differential geometry is that the geodesic can be found as the solution to a system of $2^{\text{nd}}$ order ordinary differential equations (ODEs) \cite{docarmo:1992, hauberg:nips:2012}:
\begin{wrapfigure}{r}{0.33\textwidth}
  \vspace{-5pt}
  \centering
  \includegraphics[width=0.32\textwidth]{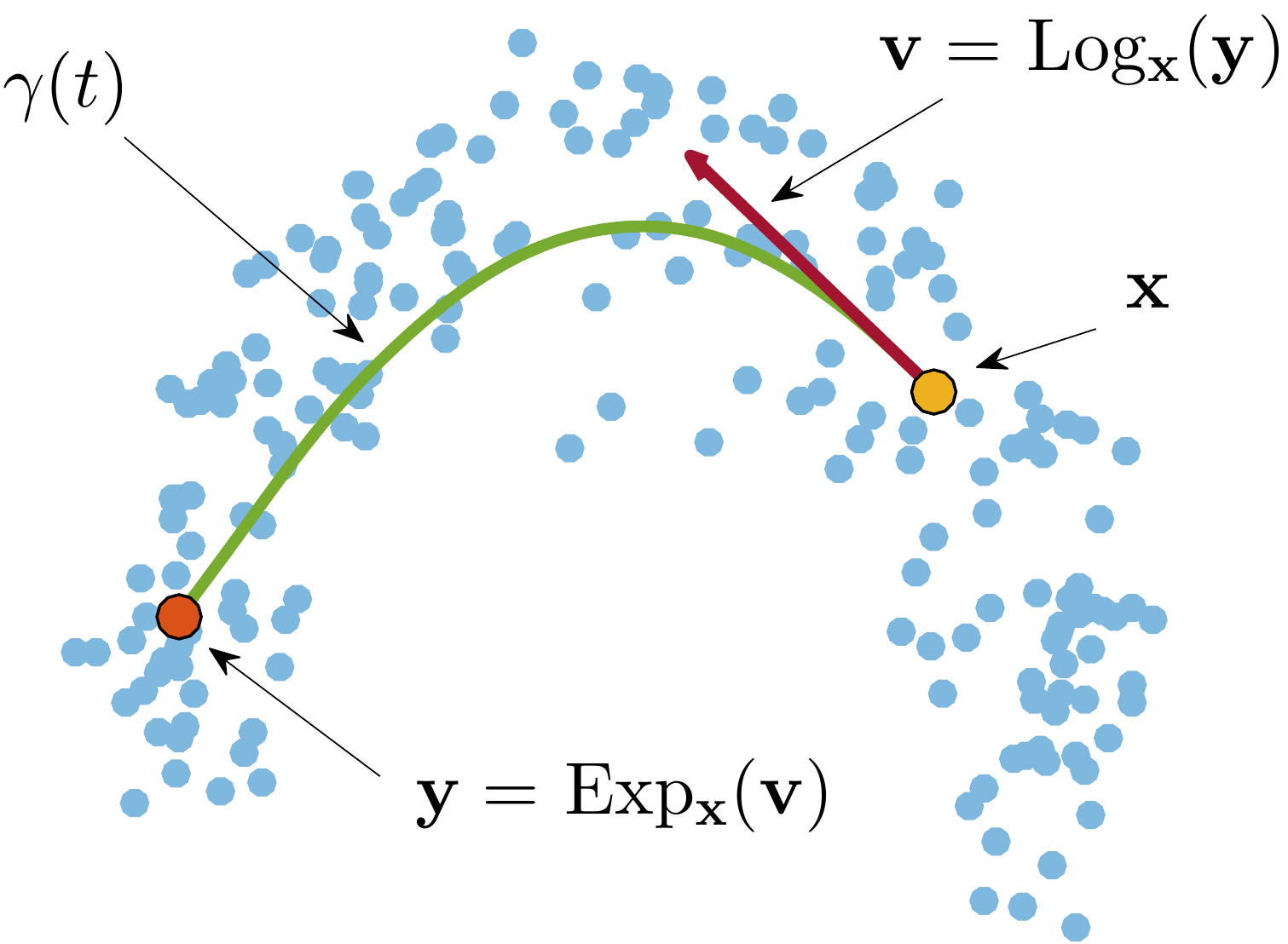}
  \caption{An illustration of the exponential and logarithmic maps.}
  \label{fig:operations_illustration}
  \vspace{-40pt}
\end{wrapfigure}
\begin{align}
\bs{\gamma}''(t) = -\frac{1}{2}\b{M}^{-1}(\bs{\gamma}(t))\left[\parder{\text{vec}[{\b{M}(\bs{\gamma}(t))}]}{\bs{\gamma}(t)}\right]^\t \left(\bs{\gamma}'(t) \otimes \bs{\gamma}'(t) \right)
\label{eq:geodesic_ode} 
\end{align}
subject to $\bs{\gamma}(0) = \b{x}, ~\bs{\gamma}(1) = \b{y}$. Here $\text{vec}[\cdot]$ stacks the columns of a matrix into a vector and $\otimes$ is the Kronecker product. 

This differential equation allows us to define basic operations on the manifold.
The \textit{exponential map} at a point $\b{x}$ takes a tangent vector $\b{v} \in \tangent{\b{x}}$ to $\b{y} = \Exp{\b{x}}{\b{v}}\in\M$ such that the curve $\bs{\gamma}(t) = \Exp{\b{x}}{t \cdot \b{v}}$ is a geodesic originating at $\b{x}$ with initial velocity $\b{v}$ and length $\| \b{v} \|$.
The inverse mapping, which takes $\b{y}$ to $\tangent{\b{x}}$ is known as the \textit{logarithm map} and is denoted $\Log{\b{x}}{\b{y}}$. By definition $\| \Log{\b{x}}{\b{y}} \|$ corresponds to the geodesic distance from $\b{x}$ to $\b{y}$.
These operations are illustrated in Fig.~\ref{fig:operations_illustration}.
The exponential and the logarithmic map can be computed by solving Eq.~\ref{eq:geodesic_ode} numerically, as an \textit{initial value problem (IVP)} or a \textit{boundary value problem (BVP)} respectively. In practice the IVPs are substantially faster to compute than the BVPs.

The Mahalanobis distance is naturally extended to Riemannian manifolds as $\text{dist}^2_{\bs{\Sigma}}(\b{x},\b{y}) = \inner{\Log{\b{x}}{\b{y}}}{\bs{\Sigma}^{-1}\Log{\b{x}}{\b{y}}}$. From this, \citet{Pennec:JMIV:06} considered the Riemannian normal distribution
\begin{align}
\label{eq:pennec_normal_distribution}
  p_{\M}(\b{x}~|~ \bs{\mu},\bs{\Sigma}) = \frac{1}{\C} \exp\left(-\frac{1}{2} \inner{ \Log{\bs{\mu}}{\b{x}}}{\b{\bs{\Sigma}^{-1}}\Log{\bs{\mu}}{\b{x}}}  \right), \quad \b{x}\in\M
\end{align}
and showed that it is the manifold-valued distribution with maximum entropy subject to a known mean and covariance. This distribution is an instance of Eq.~\ref{eq:normal_pdf_general} and is the distribution we consider in this paper. Next, we consider standard ``intrinsic least squares'' estimates of $\bs{\mu}$ and $\bs{\Sigma}$.

%\begin{wrapfigure}{r}{0pt}
%% FOR RESIZE HORIZONTIACAL THE IMAGE CHANGE THE at end TO very near end FOR THE TANGENT LABEL
%    \centering
%%     \vspace{-40pt} 
%\begin{tikzpicture}[>=stealth,blackarr/.style={->,minColThree,shorten >= 3pt}]
%\draw [thick, minColOne](-2,0) .. controls (-1,1) and  (2,3) .. (2,0) node [black, at end, left] (TextNode) {$\M$};
%\draw [thick, minColTwo](-2,1.6) -- (2,1.6) node [black,  at end, right, above] (TextNode) {$\tangent{\b{x}}$};
%\coordinate (A) at (-1.5,0.5) ;
%\coordinate (B) at (-1.5,1.6) ;
%\draw[blackarr] (A) to [bend right = 50] node [black, near end ,right]{\small $\text{Log}_{\mu}$} (B) ;
%\draw[blackarr] (B) to [bend right = 50] node [black, midway,left]{\small $\text{Exp}_{\mu}$} (A) ;
%\filldraw [gray] (A) circle (2pt) node [black, below right] (TextNode) {$\b{x}$};
%\filldraw [gray] (0.9,1.6) circle (2pt) node [black,below] (TextNode) {$\b{\mu}$};
%\filldraw [gray] (-1.5,1.6) circle (2pt) node [black,above] (TextNode) {$\b{v}$};
%\end{tikzpicture}
%  \caption{An illustration of the two operations exponential and logarithmic map.}
%% \vspace{-20pt} 
%\end{wrapfigure}

\subsection{Intrinsic Least Squares Estimators}

Let the data be generated from an unknown probability distribution $q_{\M}(\b{x})$ on a manifold. Then it is common \cite{Pennec:JMIV:06} to define the \textit{intrinsic mean} of the distribution as the point that minimize the variance
\begin{align}
  \hat{\bs{\mu}} = \argmin_{\bs{\mu}\in\M} \int_{\M} \text{dist}^2(\bs{\mu},\b{x})q_{\M}(\b{x})\dif{d}\M(\b{x}),
  \label{eq:def_mu}
\end{align}
where $\dif{d}\M(\b{x})$ is the measure (or infinitesimal volume element) induced by the metric. Based on the mean, a covariance matrix can be defined
\begin{align}
  \hat{\bs{\Sigma}} = \int_{\D(\hat{\bs{\mu}})} \Log{\hat{\bs{\mu}}}{\b{x}}\Log{\hat{\bs{\mu}}}{\b{x}}^\t q_{\M}(\b{x})\dif{d}\M(\b{x}),
  \label{eq:def_cov}
\end{align}
where $\D(\hat{\bs{\mu}})$ is the domain over which $\tangent{\hat{\bs{\bs{\mu}}}}$ is well-defined.
For the manifolds we consider, the domain $\D(\hat{\bs{\mu}})$ is $\R^D$.
Practical estimators of $\hat{\bs{\mu}}$ rely on gradient-based optimization to find a local minimizer of Eq.~\ref{eq:def_mu}, which is well-defined \cite{karcher:1977}.
For finite data $\{\b{x}_n\}_{n=1}^N$, the descent direction is proportional to $\hat{\b{v}} = \sum_{n=1}^N \Log{\bs{\mu}}{\b{x}_n} \in \tangent{\bs{\mu}}$, and the updated mean is a point on the geodesic curve $\bs{\gamma}(t)=\Exp{\bs{\mu}}{t\cdot \hat{\b{v}}}$. After estimating the mean, the empirical covariance matrix is estimated as $\hat{\bs{\Sigma}} = \frac{1}{N-1} \sum_{n=1}^N \Log{\hat{\bs{\mu}}}{\b{x}_n} \Log{\hat{\bs{\mu}}}{\b{x}_n}^\t$. It is worth noting that even though these estimators are natural, they are not maximum likelihood estimates for the Riemannian normal distribution \eqref{eq:pennec_normal_distribution}.

In practice, the intrinsic mean often falls in regions of low data density \cite{hauberg:tpami:princurve}. For instance, consider data distributed uniformly on the equator of a sphere, then the optima of Eq.~\ref{eq:def_mu} is either of the poles. Consequently, the empirical covariance is often overestimated.
%this will influence negatively the empirical covariance. Also, we can not rule out the fact that the empirical covariance estimators usually overestimate the given data.

\section{A Locally Adaptive Normal Distribution}

We now have the tools to define a locally adaptive normal distribution (LAND): we replace the linear Euclidean distance with a locally adaptive Riemannian distance and study the corresponding Riemannian normal distribution \eqref{eq:pennec_normal_distribution}. By learning a Riemannian manifold and using its structure to estimate distributions of the data, we provide a new and useful link between Riemannian statistics and manifold learning.

\subsection{Constructing a Metric \label{sec:metric_learning}}

In the context of manifold learning, \citet{hauberg:nips:2012} suggest to model the local behavior of the data manifold via a locally-defined Riemannian metric. Here we propose to use a local covariance matrix to represent the local structure of the data. We only consider diagonal covariances for computational efficiency and to prevent the overfitting. The locality of the covariance is defined via an isotropic Gaussian kernel of size $\sigma$. Thus, the metric tensor at $\b{x}\in\M$ is defined as the inverse of a local diagonal covariance matrix with entries
\begin{align}
  {M}_{dd}(\b{x}) = \left( \sum_{n=1}^N w_n(\b{x}) (x_{nd} - x_d)^2 + \rho \right)^{-1},
%\bs{\Sigma}(\b{x}) = \sum_{n=1}^N w_n \text{diag}\left[(\b{x}_n - \b{x})(\b{x}_n - \b{x})^{\t}\right] + \rho, 
\quad \text{with} \quad w_n(\b{x}) = \exp\left(-\frac{\norm{\b{x}_n - \b{x}}^2_2}{2\sigma^2} \right).
\end{align}
Here $x_{nd}$ is the $d^{\text{th}}$ dimension of the $n^{\text{th}}$ observation, and $\rho$ a regularization parameter to avoid singular covariances.
%, and the final metric is $\b{M}(\b{x}) = \bs{Q}(\b{x})^{-1}$. 
This defines a smoothly changing (hence Riemannian) metric that captures the local structure of the data.
It is easy to see that if $\b{x}$ is outside of the support of the data, then the metric tensor is large. Thus, geodesics are ``pulled'' towards the data where the metric is small. Note that the proposed metric is not invariant to linear transformations.While we restrict our attention to this particular choice, other learned metrics are equally applicable, c.f. \cite{Tosi:UAI:2014,hauberg:nips:2012}.

\subsection{Estimating the Normalization Constant}

The normalization constant of Eq.~\ref{eq:pennec_normal_distribution} is by definition
\begin{equation}
\label{eq:norm_const_int}
 \C(\bs{\mu},\bs{\Sigma}) = \int_{\M} \exp\left(-\frac{1}{2}\inner{\Log{\bs{\mu}}{\b{x}}}{\bs{\Sigma}^{-1}\Log{\bs{\mu}}{\b{x}}}\right) \dif{d}\M(\b{x}),
\end{equation}
where $\dif{d}\M(\b{x})$ denotes the measure induced by the Riemannian metric.
The constant $\C(\bs{\mu},\bs{\Sigma})$ depends not only on the covariance matrix, but also on the mean of the distribution, and the curvature of the manifold (captured by the logarithm map). For a general learned manifold, $\C(\bs{\mu},\bs{\Sigma})$ is inaccessible in closed-form and we resort to numerical techniques. We start by rewriting Eq.~\ref{eq:norm_const_int} as
\begin{equation}
  \label{eq:norm_const_int_tangent}
  \C(\bs{\mu},\bs{\Sigma}) = \int_{\tangent{\bs{\mu}}} \sqrt{\abs{\b{M}(\Exp{\bs{\mu}}{\b{v}})}} \exp\left(-\frac{1}{2}\inner{\b{v}}{\bs{\Sigma}^{-1}\b{v}}\right)  \dif{d} \b{v}.
\end{equation}
In effect, we integrate the distribution over the tangent space $\tangent{\bs{\mu}}$ instead of directly over the manifold. This transformation relies on the fact that the volume of an infinitely small area on the manifold can be computed in the tangent space if we take the deformation of the metric into account \cite{Pennec:JMIV:06}. This deformation is captured by the measure which, in the tangent space, is $\dif{d}\M(\b{x}) = \sqrt{\abs{\b{M}(\Exp{\bs{\mu}}{\b{v}})}} \dif{d}\b{v}$. For notational simplicity we define the function $m(\bs{\mu},\b{v}) = \sqrt{\abs{\b{M}(\Exp{\bs{\mu}}{\b{v}})}}$, which intuitively captures the cost for a point to be outside the data support ($m$ is large in low density areas and small where the density is high).

\begin{wrapfigure}{r}{0.3\textwidth}
  \vspace{-12pt}
  \centering
  \includegraphics[width=0.29\textwidth]{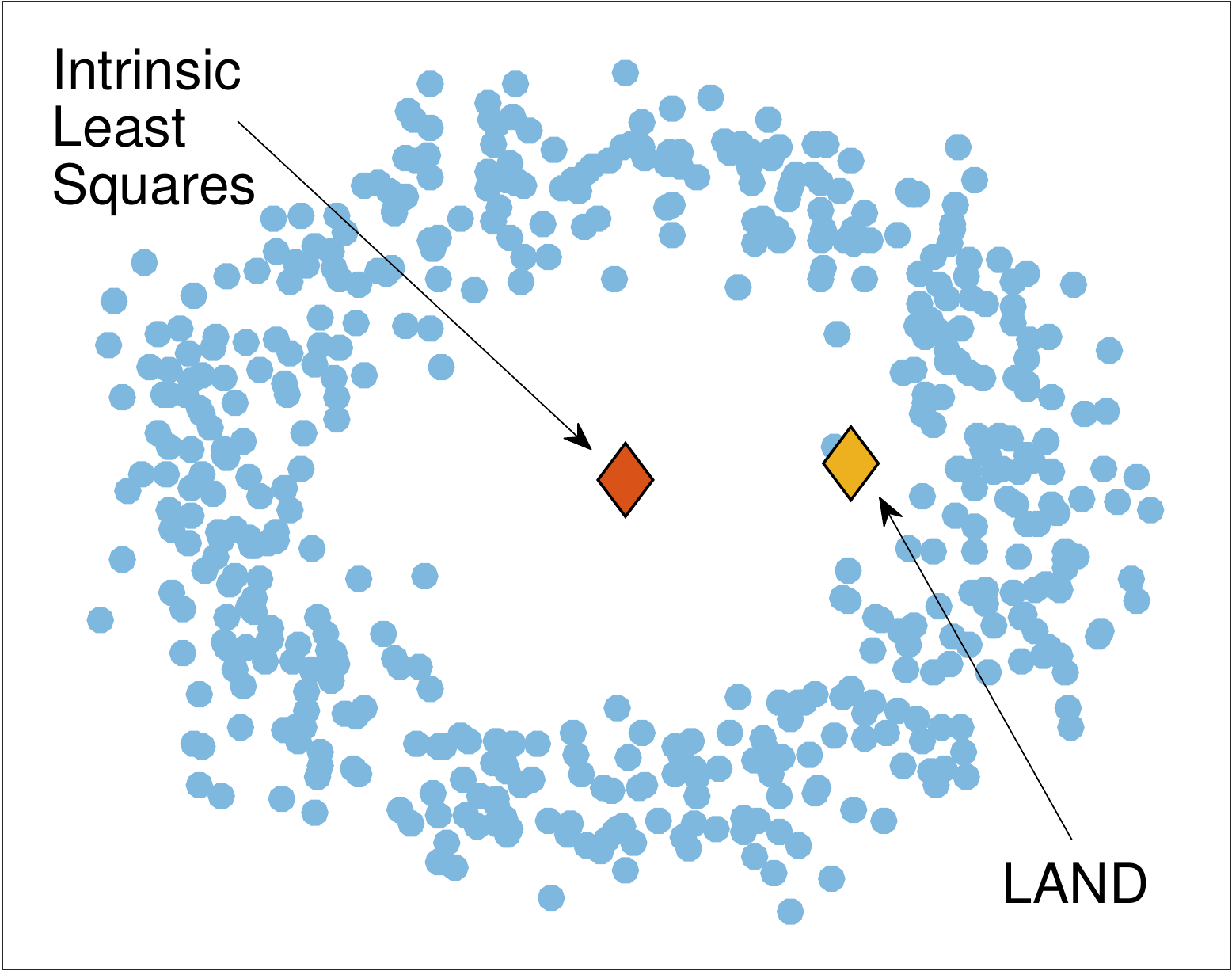}
  \caption{Comparison of LAND and intrinsic least squares means.}
  \label{fig:means_comparison}
  \vspace{-15pt}
\end{wrapfigure}
We estimate the normalization constant \eqref{eq:norm_const_int_tangent} using Monte Carlo integration. We first multiply and divide the integral with the normalization constant of the Euclidean normal distribution $\Z =  \sqrt{ (2\pi)^D \abs{\bs{\Sigma}} }$. Then, the integral becomes an expectation estimation problem $\C(\bs{\mu},\bs{\Sigma}) = \Z \cdot \E_{\N(0,\bs{\Sigma})}[m(\bs{\mu},\b{v})]$, which can be estimated numerically as 
\begin{equation}
  \label{eq:norm_const_est}
  \C(\bs{\mu},\bs{\Sigma}) \simeq \frac{\Z}{S} \sum_{s=1}^S m(\bs{\mu},\b{v}_s),\quad \text{where} \quad \b{v}_s \sim \N(0,\bs{\Sigma})
\end{equation}
and $S$ is the number of samples on $\tangent{\bs{\mu}}$. The computationally expensive element is to evaluate $m$, which in turn requires evaluating $\Exp{\bs{\mu}}{\b{v}}$. This amounts to solving an IVP numerically, which is fairly fast. Had we performed the integration directly on the manifold \eqref{eq:norm_const_int} we would have had to evaluate the logarithm map, which is a much more expensive BVP. The tangent space integration, thus, scales better.

\subsection{Inferring Parameters}

Assuming an independent and identically distributed dataset $\{\b{x}_n\}_{n=1}^N$, we can write their joint distribution as $p_{\M}(\b{x}_1,\dots,\b{x}_N) = \prod_{n=1}^N p_{\M}(\b{x}_n ~|~ \bs{\mu},\bs{\Sigma})$. We find parameters $\bs{\mu}$ and $\bs{\Sigma}$ by maximum likelihood, which we implement by minimizing the mean negative log-likelihood
\begin{equation}
  \label{eq:mle_obj}
  \{\hat{\bs{\mu}},\hat{\bs{\Sigma}}\} =\argmin_{\substack{\bs{\mu}\in\M\\ \bs{\Sigma}\in\S_{++}^D}} \phi\left(\bs{\mu},\bs{\Sigma}\right) =\argmin_{\substack{\bs{\mu}\in\M\\ \bs{\Sigma}\in\S_{++}^D}} \frac{1}{2N}\sum_{n=1}^N \inner{\Log{\bs{\mu}}{\b{x}_n}}{\bs{\Sigma}^{-1}\Log{\bs{\mu}}{\b{x}_n}} + \log\left(\C(\bs{\mu},\bs{\Sigma})\right).
\end{equation}
The first term of the objective function $\phi:\M\times \S_{++}^D$ is a data-fitting term, while the second can be seen as a force that both pulls the mean closer to the high density areas and shrinks the covariance. Specifically, when the mean is in low density areas, as well as when the covariance gives significant 
\begin{wrapfigure}{r}{0.475\textwidth}
    \vspace{-10pt}
    \begin{minipage}{0.475\textwidth}
\begin{algorithm}[H]
	\caption{LAND maximum likelihood}
	\label{alg:mle}
      \algsetup{indent=1em}
        \begin{algorithmic}[1]
         	\REQUIRE{the data $\{\b{x}_n\}_{n=1}^N$, stepsizes $\alpha_{\bs{\mu}},\alpha_{\b{A}}$}
         	\ENSURE {the estimated $\hat{\bs{\mu}},~\hat{\bs{\Sigma}},~\hat{\C}(\hat{\bs{\mu}},\hat{\bs{\Sigma}})$}
         	\STATE{initialize $\bs{\mu}^0,\bs{\Sigma}^0$ and $t\leftarrow0$}
			 \REPEAT
			 	\STATE{estimate $\C(\bs{\mu}^t,\bs{\Sigma}^t)$ using Eq.~\ref{eq:norm_const_est} }
			 \STATE{compute $d_{\bs{\mu}}\phi(\bs{\mu}^t,\bs{\Sigma}^t)$ using Eq.~\ref{eq:grad_mu}}
			 \STATE{$\bs{\mu}^{t+1} \leftarrow \Exp{\bs{\mu}^t}{\alpha_{\bs{\mu}}d_{\bs{\mu}}\phi(\bs{\mu}^t,\bs{\Sigma}^t) }$}
			\STATE{estimate $\C(\bs{\mu}^{t+1},\bs{\Sigma}^t)$ using Eq.~\ref{eq:norm_const_int_tangent} }
			\STATE{compute $\grad{\b{A}}\phi(\bs{\mu}^{t+1},\bs{\Sigma}^t)$ using Eq.~\ref{eq:grad_A}}
			\STATE{$\b{A}^{t+1} \leftarrow \b{A}^t - \alpha_{\b{A}} \grad{\b{A}}\phi(\bs{\mu}^{t+1},\bs{\Sigma}^t)$}
			\STATE{$\bs{\Sigma}^{t+1} \leftarrow [(\b{A}^{t+1})^\t \b{A}^{t+1}]^{-1}$} 
			\STATE{$t \leftarrow t+1$}
%			 \UNTIL{stopping criterion}
			\UNTIL{$\norm{\phi(\bs{\mu}^{t+1},\bs{\Sigma}^{t+1}) - \phi(\bs{\mu}^{t},\bs{\Sigma}^{t})}_2^2 \leq \epsilon$}
        \end{algorithmic}
\end{algorithm}
    \end{minipage}
    \vspace{-10pt}
\end{wrapfigure}
probability to those areas, the value of $m(\bs{\mu},\b{v})$ will by construction  be large. Consequently, $\C(\bs{\mu},\bs{\Sigma})$ will increase and these solutions will be penalized. In practice, we find that the maximum likelihood LAND mean generally avoids low density regions, which is in contrast to the standard intrinsic least squares mean \eqref{eq:def_mu}, see Fig.~\ref{fig:means_comparison}.

%outside of the support of the data distances to the data become large by the construction of the learned metric. Consequently, the covariance becomes large, which is what the second term of $\phi$ penalizes.   \gear{The first term of the objective function $\phi:\M\times \S_{++}^D$ is a data-fitting term, while the second term penalizes solutions that violate the manifold constraints. Specifically, when the mean is in low density areas, by the construction of the metric the corresponding $\C(\bs{\mu},\bs{\Sigma}$ increases. The behaviour is similar when the covariance matrix gives significant probability to low density areas. Consequently, the minimizers of the objective prefer solutions that meet the manifold constraints. In practice, we find that the maximum likelihood LAND mean is generally within the support of the data, which is in contrast to the standard intrinsic mean \eqref{eq:def_mu}}

%
In practice we optimize $\phi$ using block coordinate descent: we optimize the mean keeping the covariance fixed and vice versa. Unfortunately, both of the sub-problems are non-convex, and unlike the linear normal distribution, they lack a closed-form solution. Since the logarithm map is a differentiable function, we can use gradient-based techniques to infer $\bs{\mu}$ and $\bs{\Sigma}$. Below we give the descent direction for $\bs{\mu}$ and $\bs{\Sigma}$ and the corresponding optimization scheme is given in Algorithm~\ref{alg:mle}. Initialization is discussed in Appx. \ref{appendix:initialization}.

\textbf{Optimizing $\bs{\mu}$}: the objective function is differentiable with respect to $\bs{\mu}$ \cite{docarmo:1992}, and using that $\parder{}{\bs{\mu}}\inner{\Log{\bs{\mu}}{\b{x}}}{\bs{\Sigma}^{-1}\Log{\bs{\mu}}{\b{x}}} = -2\bs{\Sigma}^{-1}\Log{\bs{\mu}}{\b{x}}$, we get the gradient
\begin{equation}
  \label{eq:grad_mu}
  \grad{\bs{\mu}} \phi(\bs{\mu},\bs{\Sigma}) =  -\bs{\Sigma}^{-1} \left[ \frac{1}{N}\sum_{n=1}^N \Log{\bs{\mu}}{\b{x}_n} - \frac{\Z}{\C(\bs{\mu},\bs{\Sigma})\cdot S}\sum_{s=1}^S m(\bs{\mu},\b{v}_s) \b{v}_s \right].
\end{equation}
It is easy to see that this gradient is highly dependent on the condition number of $\bs{\Sigma}$. We find that this, at times, makes the gradient unstable, and choose to use the steepest descent direction instead of the gradient direction. This is equal to $d_{\bs{\mu}}\phi(\bs{\mu},\bs{\Sigma}) = -\bs{\Sigma} \grad{\bs{\mu}}\phi(\bs{\mu},\bs{\Sigma})$ (see Appx. \ref{appendix:gradient_mu}).

\textbf{Optimizing $\bs{\Sigma}$}: since the covariance matrix by definition is constrained to be in the space $\S_{++}^D$, a common trick is to decompose the matrix as $\bs{\Sigma}^{-1} = \b{A}^\t \b{A}$, and optimize the objective with respect to $\b{A}$. The gradient of this factor is (see Appx. \ref{appendix:gradient_Sigma} for a derivation)
\begin{equation}
\label{eq:grad_A}
\grad{\b{A}}\phi(\bs{\mu},\bs{\Sigma}) = \b{A}\left[ \frac{1}{N} \sum_{n=1}^N \Log{\bs{\mu}}{\b{x}_n}\Log{\bs{\mu}}{\b{x}_n}^\t - \frac{\Z}{\C(\bs{\mu},\bs{\Sigma})\cdot S} \sum_{s=1}^S m(\bs{\mu},\b{v}_s)\b{v}_s\b{v}_s^\t \right].
\end{equation}
Here the first term fits the given data by increasing the size of the covariance matrix, while the second term regularizes the covariance towards a small matrix. 

\subsection{Mixture of LANDs}

At this point we can find maximum likelihood estimates of the LAND model. We can easily extend this to mixtures of LANDs: Following the derivation of the standard Gaussian mixture model \cite{Bishop:2006:PRM:1162264}, our objective function for inferring  the parameters of the LAND mixture model is formulated as follows
\begin{equation}
\label{eq:moland_obj}
\psi(\bs{\Theta})
 = \sum_{k=1}^K \sum_{n=1}^N  r_{nk} \left[ \frac{1}{2}\inner{\Log{\bs{\mu}_k}{\b{x}_n}}{\bs{\Sigma}_k^{-1}\Log{\bs{\mu}_k}{\b{x}_n}} + \log(\C(\bs{\mu}_k,\bs{\Sigma}_k)) - \log(\pi_k) \right],
\end{equation}
where $\bs{\Theta}= \{\bs{\mu}_k,\bs{\Sigma}_k\}_{k=1}^K$ , $r_{nk} = \frac{\pi_k p_{\M}(\b{x}_n ~|~\bs{\mu}_k,\bs{\Sigma}_k)}{\sum_{l=1}^K \pi_l p_{\M}(\b{x}_n ~|~\bs{\mu}_l,\bs{\Sigma}_l)}$ is the probability that $\b{x}_n$ is generated by the $k^{\text{th}}$ component, and $\sum_{k=1}^K \pi_k = 1,~\pi_k \geq 0$. The corresponding EM algorithm is in Appx. \ref{appendix:algorithms}.

\section{Experiments}

In this section we present both synthetic and real experiments to demonstrate the advantages of the LAND. We compare our model with both the Gaussian mixture model (GMM), and a mixture of LANDs using least squares (LS) estimators (\ref{eq:def_mu}, \ref{eq:def_cov}). Since the latter are not maximum likelihood estimates we use a Riemannian $K$-means algorithm to find cluster centers. In all experiments we use $S=3000$ samples in the Monte Carlo integration. This choice is investigated empirically in the supplements. Furthermore, we choose $\sigma$ as small as possible, while ensuring that the manifold is smooth enough that geodesics can be computed numerically.

\subsection{Synthetic Data Experiments}

\begin{wrapfigure}{r}{0.35\textwidth}
  \vspace{-10pt}
  \centering
  % \begin{minipage}{0.35\textwidth}
%  \includegraphics[scale = 0.28]{images/result_arc-eps-converted-to.pdf}
  \includegraphics[width = 0.33\textwidth]{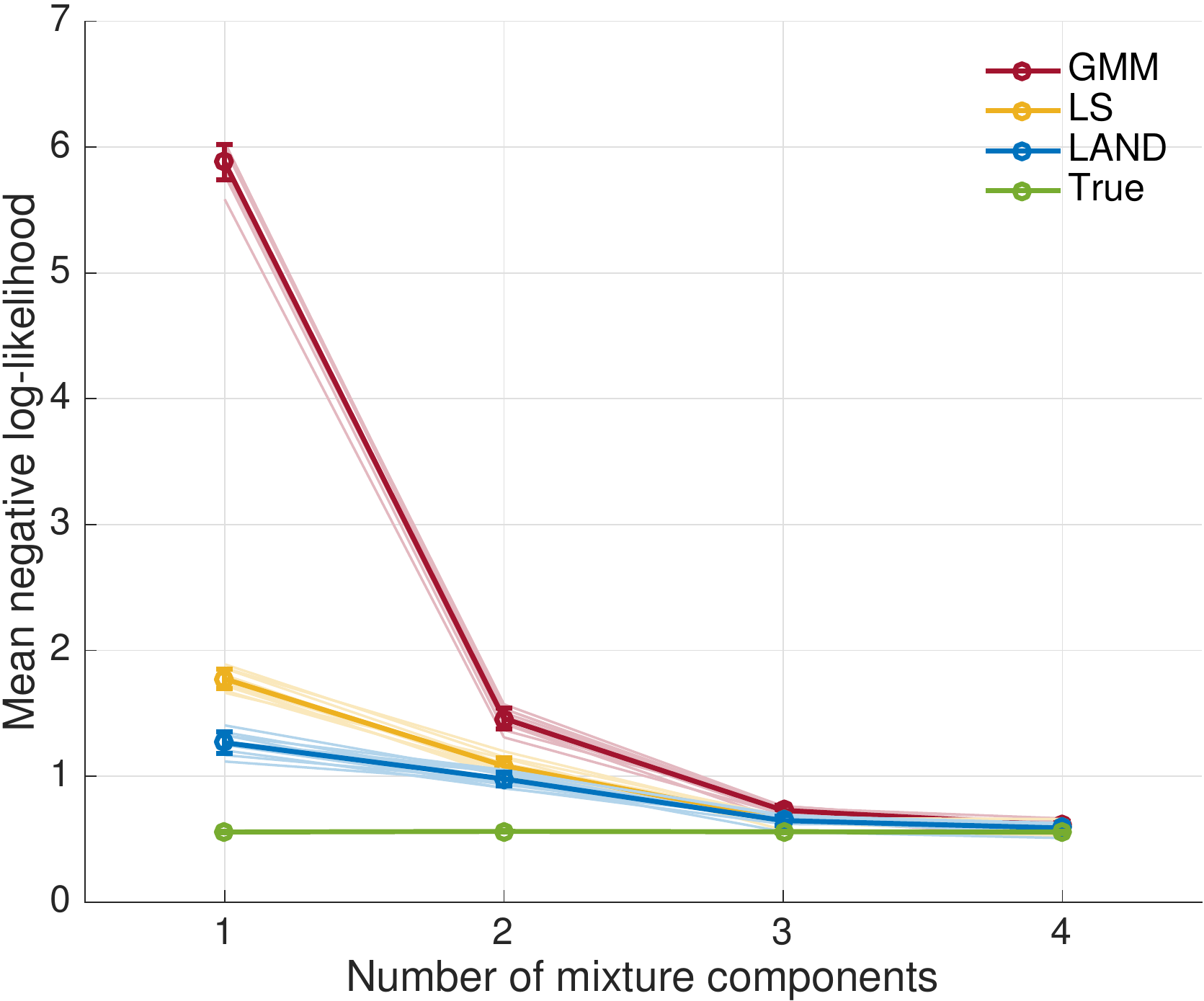}
  \caption{The mean negative log-likelihood experiment.}
  \label{fig:synthetic_1_likelihood}
  \vspace{-0pt}
\end{wrapfigure}
\begin{figure}[h]
    \centering
    \begin{subfigure}[b]{0.31\textwidth}
        \includegraphics[width=\textwidth]{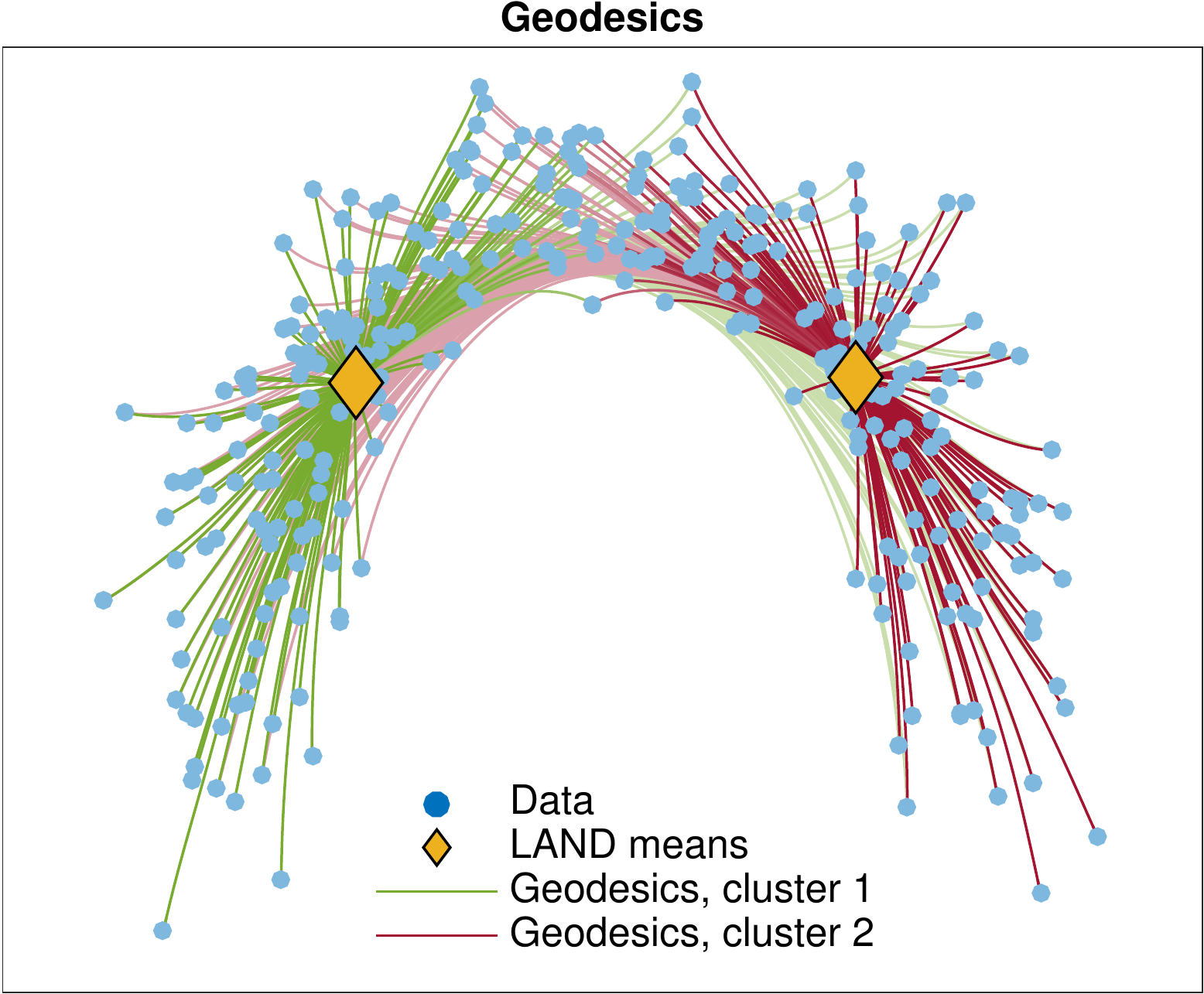}
    \end{subfigure}
    \quad
    \begin{subfigure}[b]{0.31\textwidth}
        \includegraphics[width=\textwidth]{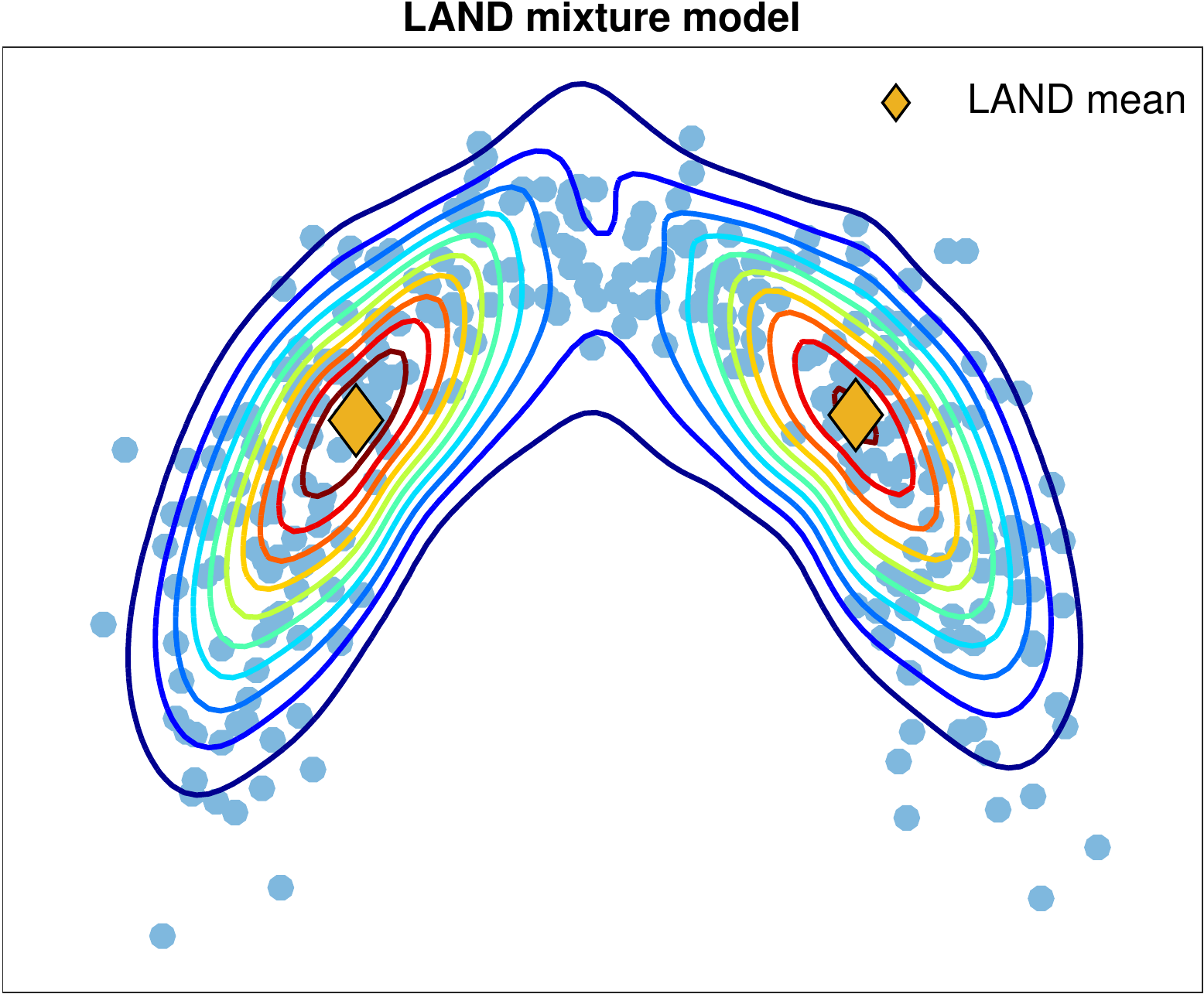}
    \end{subfigure}
    \quad
    \begin{subfigure}[b]{0.31\textwidth}
        \includegraphics[width=\textwidth]{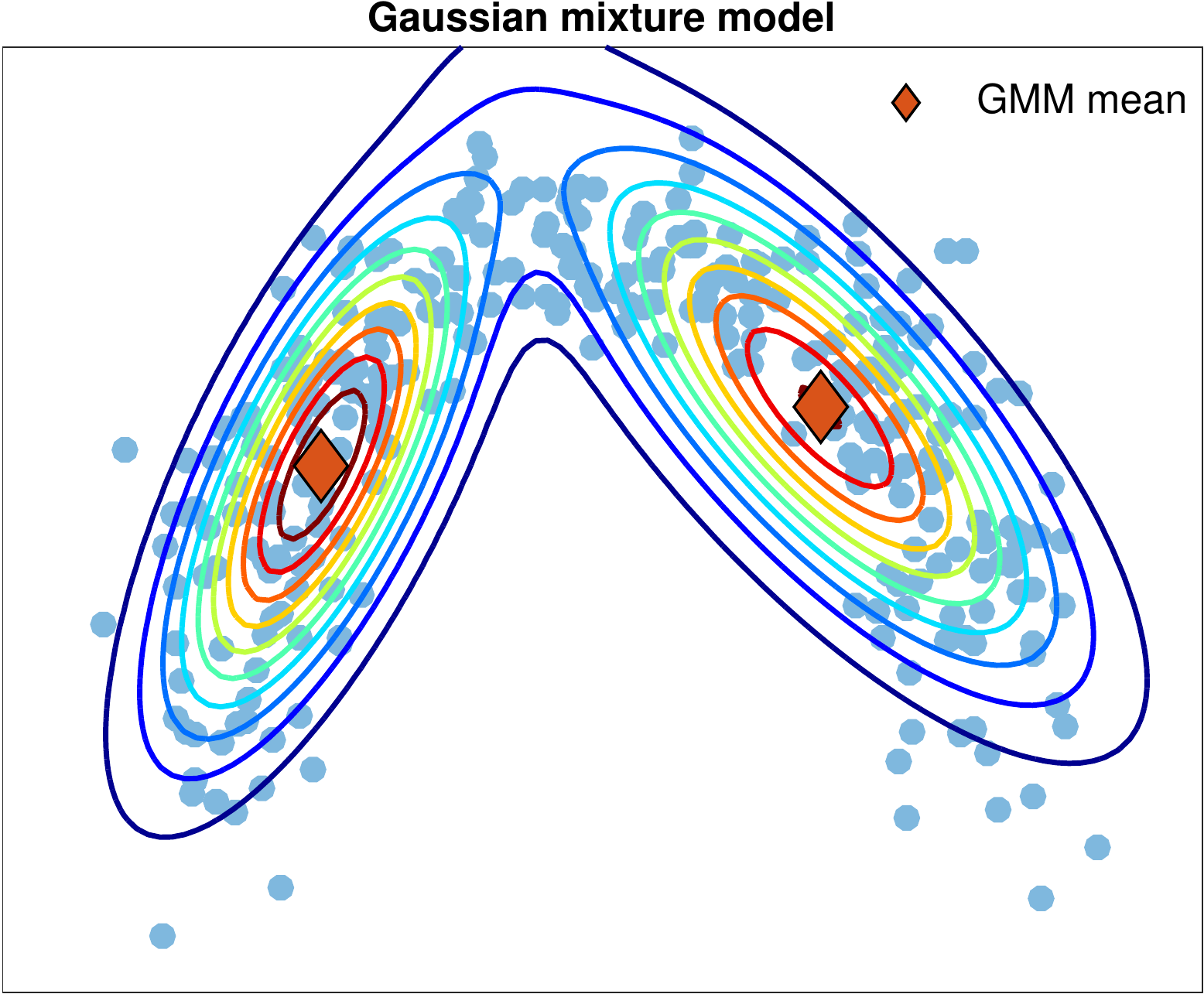}
    \end{subfigure}
    
    \caption{Synthetic data and the fitted models. \textit{Left}: the given data, the intensity of the geodesics represent the responsibility of the point to the corresponding cluster. \textit{Center}: the contours of the LAND mixture model. \textit{Right}: the contours of the Gaussian mixture model.}
\label{fig:synthetic_1_contours}
\end{figure}

As a first experiment, we generate a nonlinear data-manifold by sampling from a mixture of 20 Gaussians positioned along a half-ellipsoidal curve (see left panel of Fig.~\ref{fig:synthetic_1_contours}).
We generate 10 datasets with 300 points each, and fit for each dataset the three models with $K=1,\dots,4$ number of components. Then, we generate 10000 samples from each fitted model, and we compute the mean negative log-likelihood of the true generative distribution using these samples. Fig.~\ref{fig:synthetic_1_likelihood} shows that the LAND learns faster the underlying true distribution, than the GMM. Moreover, the LAND perform better than the least squares estimators, which overestimates the covariance. In  the supplements we show, using the standard AIC and BIC criteria, that the optimal LAND is achieved for $K=1$, while for the least squares estimators and the GMM, the optimal is achieved for $K=3$ and $K=4$ respectively.

In addition, in Fig.~\ref{fig:synthetic_1_contours} we show the contours for the LAND and the GMM for $K=2$. There, we can observe that indeed, the LAND adapts locally to the data and reveals their underlying nonlinear structure. This is particularly evident near the ``boundaries'' of the data-manifold.

We extend this experiment to a clustering task (see left panel of Fig.~\ref{fig:synthetic_2} for data).
The center and right panels of Fig.~\ref{fig:synthetic_2} show the contours of the LAND and Gaussian mixtures, and it is evident that the LAND is substantially better at capturing non-ellipsoidal clusters. Due to space limitations, we move further illustrative experiments to Appx. \ref{appendix:experiments} and continue with real data.

%For the second demonstration, we consider the clustering problem. We generated two datasets, with two clusters each dataset, and each cluster has 300 points, based on the previous true distribution. The task is to find the underlying distributions of the clusters. From the results Fig.~\ref{fig:synthetic_2} we observe that actually, the LAND is able to find the underlying distributions of the clusters, while the GMM fails even to find the correct means of the distributions.  \gear{Due to space limitations, more illustrative experiments, as well as the result for the IMM model, can be found in the supplementary material.}

\begin{figure}[h]
\centering
    \begin{subfigure}[b]{0.31\textwidth}
        \includegraphics[width=\textwidth]{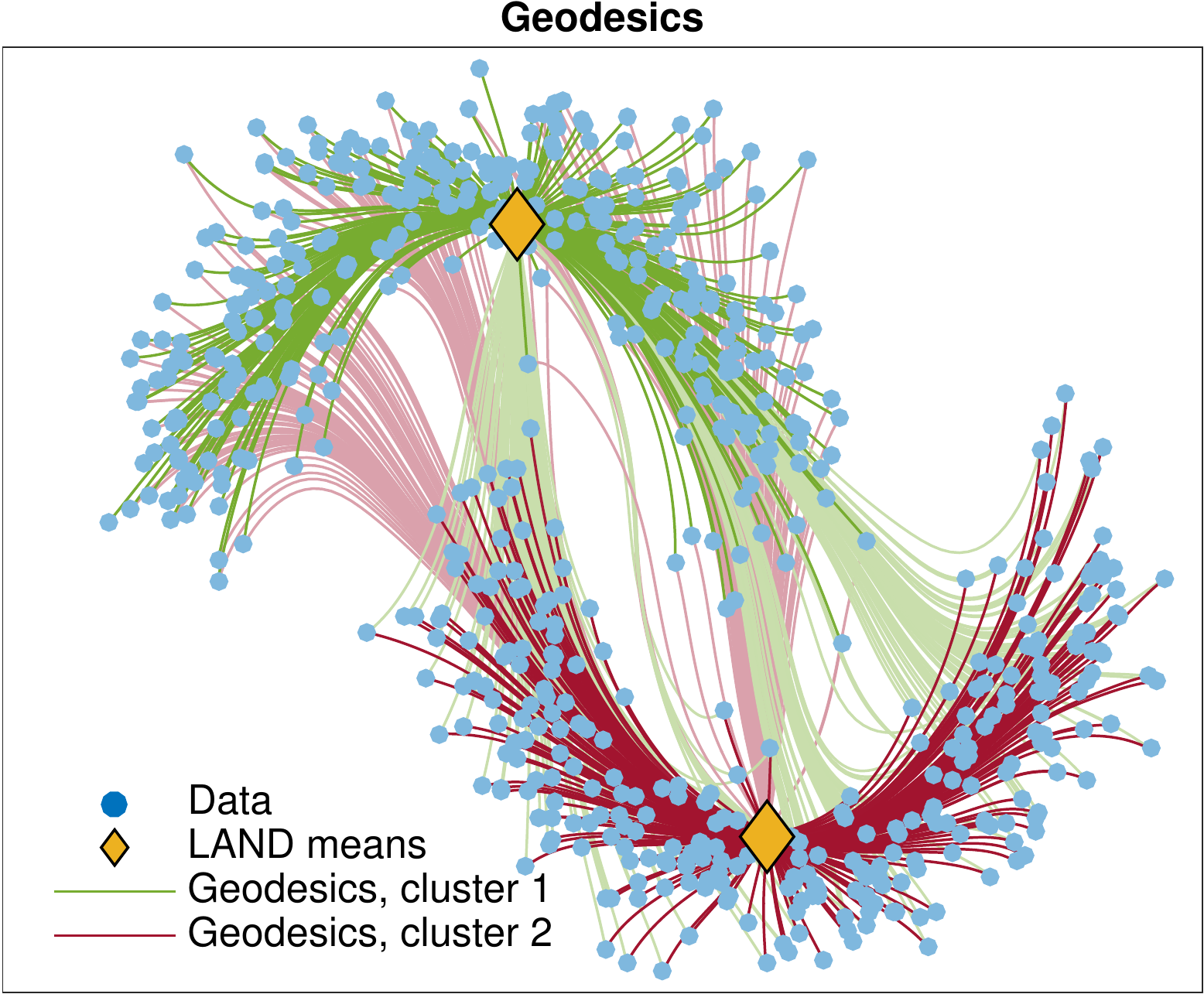}
    \end{subfigure}
    \quad
    \begin{subfigure}[b]{0.31\textwidth}
        \includegraphics[width=\textwidth]{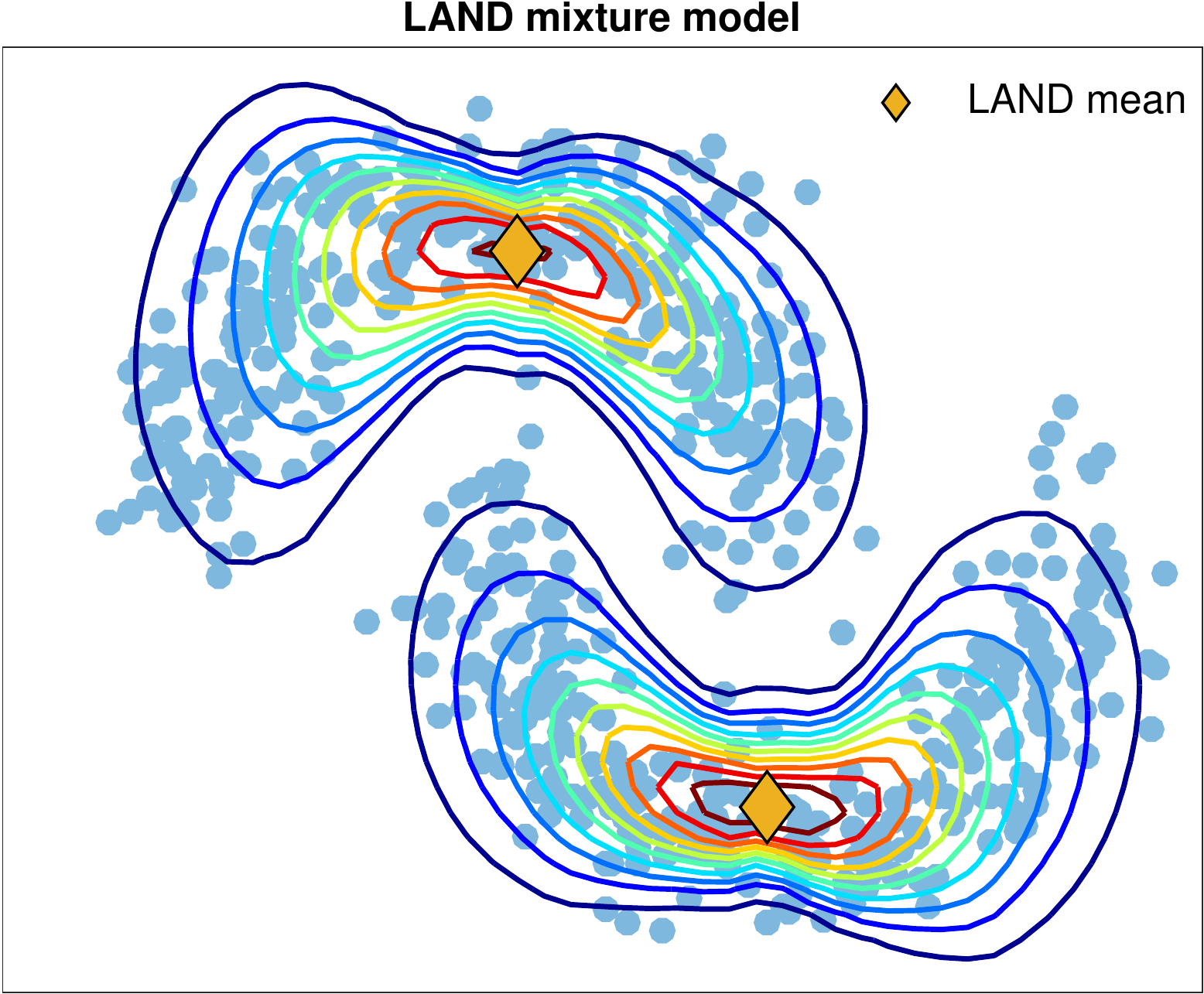}
    \end{subfigure}
    \quad    
    \begin{subfigure}[b]{0.31\textwidth}
        \includegraphics[width=\textwidth]{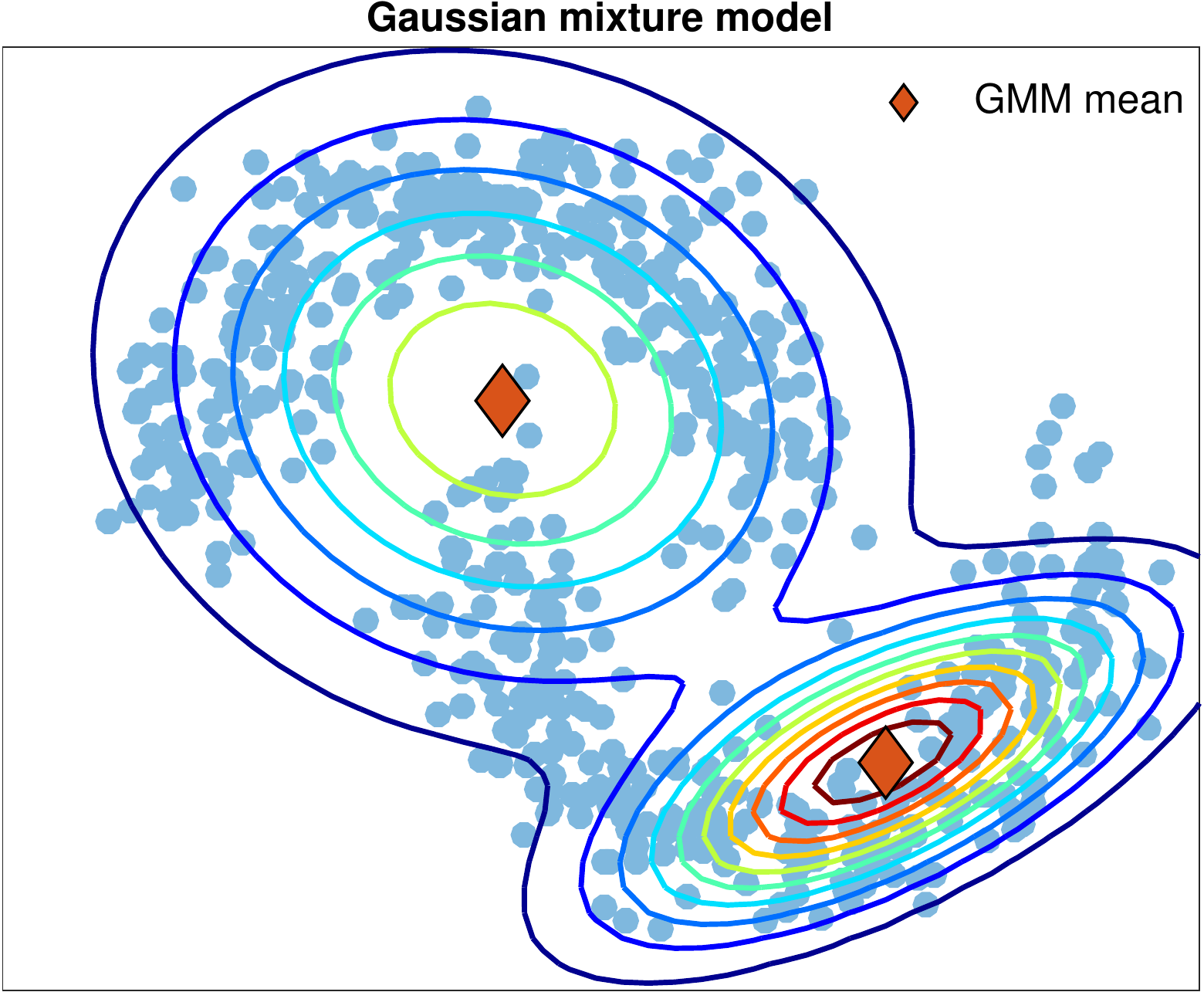}
    \end{subfigure}
    
    \vspace*{0.1cm}
    
    \begin{subfigure}[b]{0.31\textwidth}
        \includegraphics[width=\textwidth]{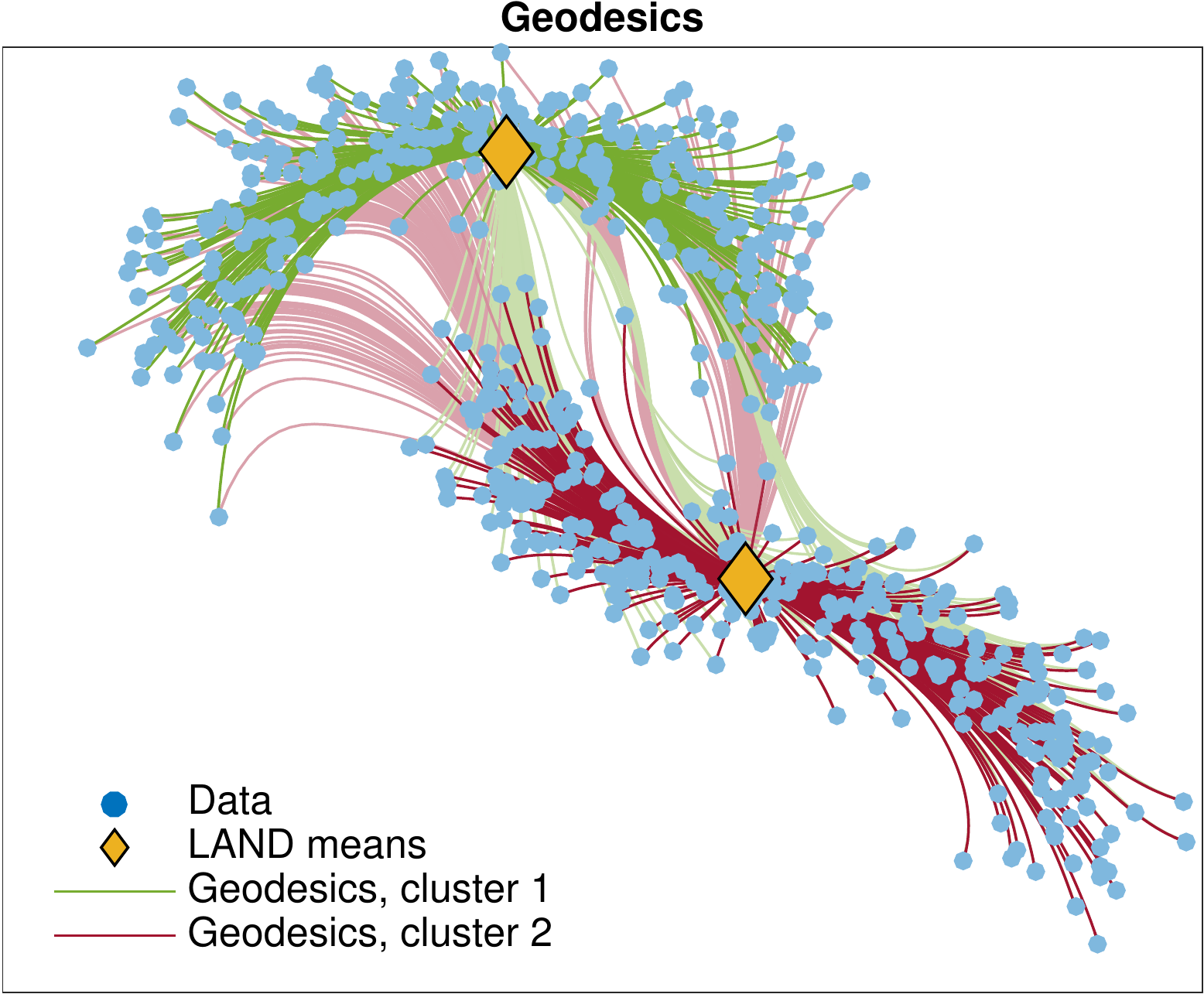}
    \end{subfigure}
    \quad
    \begin{subfigure}[b]{0.31\textwidth}
        \includegraphics[width=\textwidth]{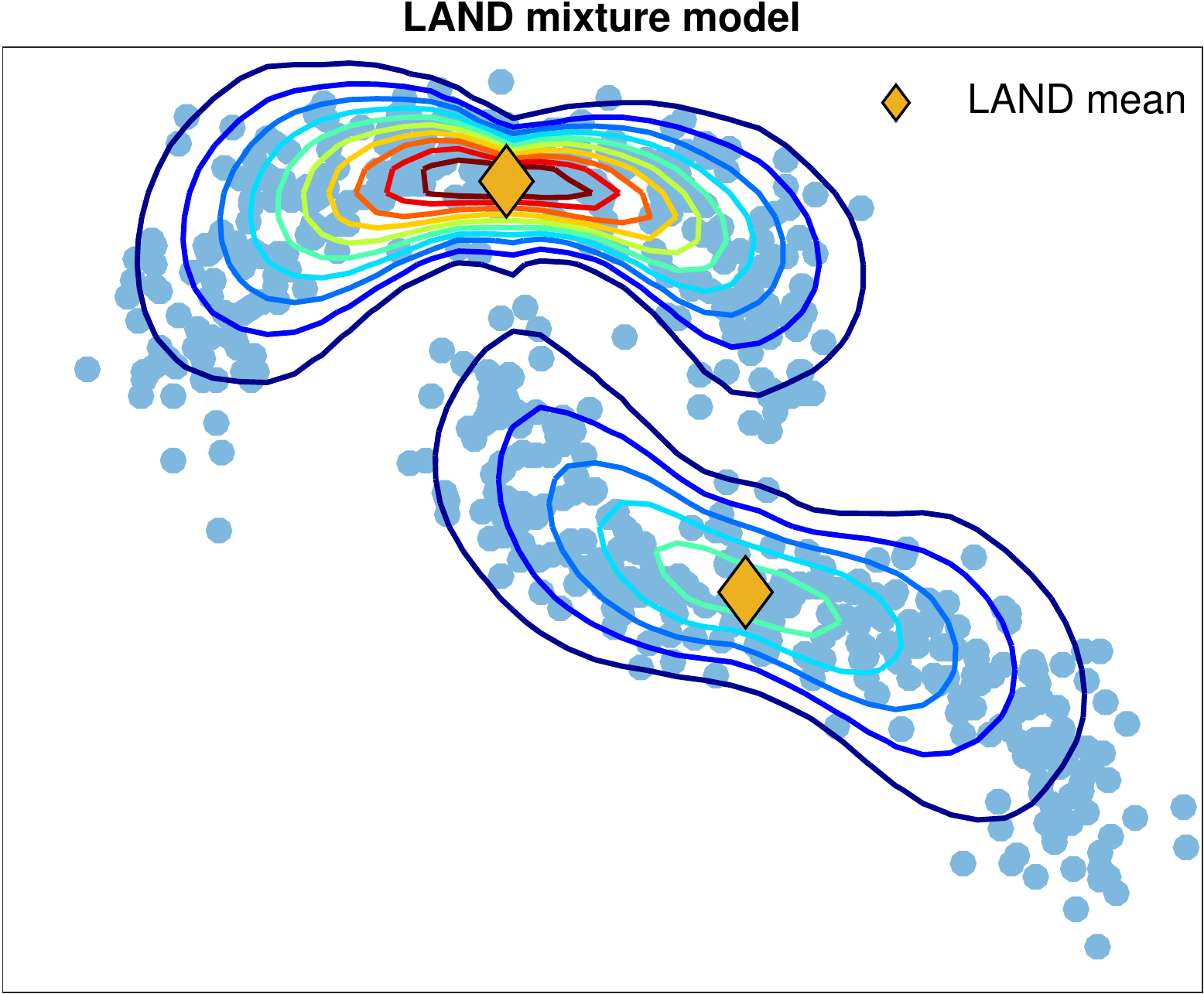}
    \end{subfigure}
    \quad
    \begin{subfigure}[b]{0.31\textwidth}
        \includegraphics[width=\textwidth]{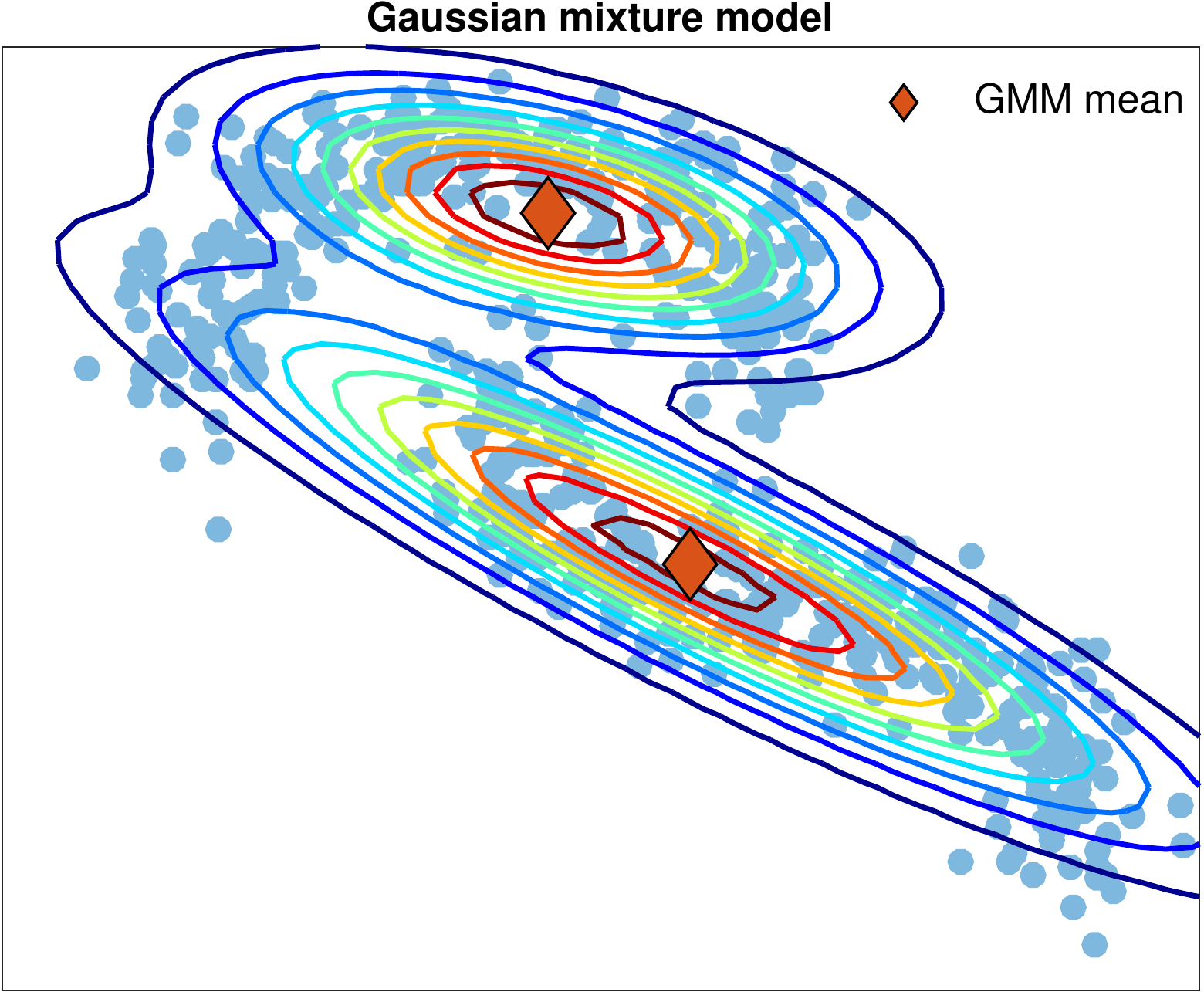}
    \end{subfigure}
    
    \caption{The clustering problem for two synthetic datasets. \textit{Left}: the given data, the intensity of the geodesics represent the responsibility of the point to the corresponding cluster. \textit{Center}: the LAND mixture model. \textit{Right}: the Gaussian mixture model.}\label{fig:synthetic_2}
\end{figure}

\subsection{Modeling Sleep Stages}

%\begin{figure}[t]
%\centering
%\smartdiagramset{
%  back arrow disabled=true,
%}
%\smartdiagram[flow diagram:horizontal]
%{
%  EEG Signal, Short time Fourier Transform, Spectrum magnitudes, Non-Negative Matrix Factorization, Features Vectors
%}
%
%\caption{The feature extraction procedure.}
%\label{eeg:feature_extraction}
%\end{figure}

We consider electro-encephalography (EEG) measurements of human sleep from 10 subjects, part of the PhysioNet database \cite{imtiaz:01,PhysioNet,Delorme_eeglab:an}. For each subject we get EEG measurements during sleep from two electrodes on the front and the back of the head, respectively. 
Measurements are sampled at $f_s = 100$Hz, and for each 30 second window a so-called sleep stage label is assigned from the set  $\{1,2,3,4,\text{REM, awake}\}$. Rapid eye movement (REM) sleep is particularly interesting, characterized by having EEG patterns similar to the awake state but with a complex physiological pattern, involving e.g.,  reduced muscle tone, rolling eye movements and erection \cite{purvesneuroscience}. Recent evidence points to the importance of REM sleep for memory consolidation \cite{boyce2016causal}. Periods in which the sleeper is awake are typically happening in or near REM intervals. Thus we here consider the characterization of sleep in terms of three categories REM, awake, and non-REM, the latter a merger of sleep stages $1-4$.
%(In order to extract reasonable features from the EEG of one subject, we followed the following procedure. We divided the 30s widow to 3 smaller windows of 10s, and we computed the spectrogram of the signal for windows of size $N_s = 1000$ samples, with overlap of $50\%$, and we discarded the windows on the change of the sleep stage. Then, using the $log(1+\abs{f})$ of the spectrum for each window we got the data matrix. Finally, we run 10 times non-negative matrix factorization (CITE?) for 5 factors in order to get finally the data matrix for the clustering.)

\begin{wrapfigure}{r}{0.35\textwidth}
  \vspace{+5pt}
  \centering
  % \begin{minipage}{0.35\textwidth}
%  \includegraphics[scale = 0.3]{images/sleep_data_exp-eps-converted-to.pdf}
\includegraphics[width = 0.33\textwidth]{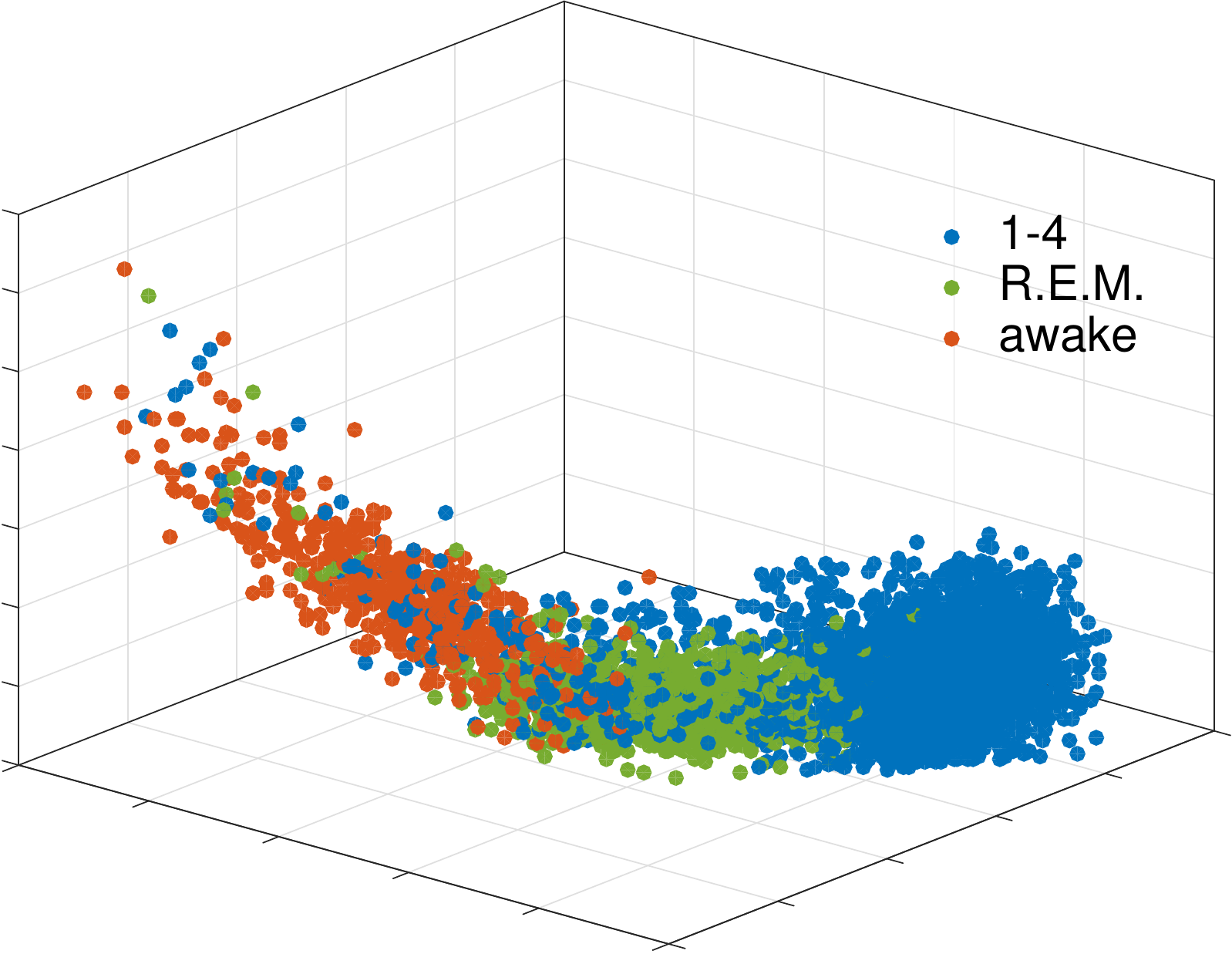}
  \caption{The 3 leading factors for subject ``s151''.}
  \label{fig:sleep_sigma_curve}
  \vspace{-5pt}
\end{wrapfigure}

We extract features from EEG measurements as follows: for each subject we subdivide the 30 second windows to 10 seconds, and apply a short-time-Fourier-transform to the EEG signal of the frontal electrode with $50\%$ overlapping windows. From this we compute the log magnitude of the spectrum $\log(1+\abs{f})$ of each window. The resulting data  matrix is decomposed using Non-Negative Matrix Factorization (10 random starts) into five factors, and we use the coefficients as 5$D$ features.  In Fig.~\ref{fig:sleep_sigma_curve} we illustrate the nonlinear manifold structure based on a three factor analysis.

We perform clustering on the data and evaluate the alignment between cluster labels and sleep stages using the  F-measure \cite{marxer2008f}. The LAND depends on the parameter $\sigma$ to construct the metric tensor, and in this experiment it is less straightforward to select $\sigma$ because of significant intersubject variability. First,  we fixed $\sigma = 1$ for all the subjects. From the results in Table \ref{tab:eeg_results} we observe that for $\sigma = 1$ the LAND(1) generally outperforms the GMM and achieves much better alignment. To further illustrate the effect of $\sigma$ we fitted a LAND for $\sigma = [0.5,0.6,\dots,1.5]$ and present the best result achieved by the LAND. Selecting $\sigma$ this way leads indeed  to higher degrees of alignment further underlining that the conspicuous manifold structure and the rather compact sleep stage distributions in Fig.~\ref{fig:sleep_sigma_curve} are both captured better with the LAND representation than with a linear GMM.  

%\[
%error(C_1,\dots,C_k) = \frac{1}{N} \sum_{k=1}^K \sum_{n\in C_k}   \mathbf{1}_{Y_n \neq Y'_k}
%\]

\begin{table}[h]
    \caption{The F-measure result for 10 subjects (the closer to 1 the better).}

\centering    
\resizebox{1\textwidth}{!}{
    \begin{tabular}{ c  c  c  c  c  c  c  c  c  c  c}
    \toprule    
				& s001 & s011 & s042 & s062 & s081 & s141 & s151 & s161 & s162 & s191 \\ \midrule
			LAND(1) & $\b{0.831}$ & $\b{0.701}$ & $0.670$ & $\b{0.740}$ & $\b{0.804}$ & $0.870$ & $\b{0.820}$ & $\b{0.780}$ & $0.747$ & $\b{0.786}$ \\
            GMM	& $0.812$ & $0.690$ & $\b{0.675}$ & $0.651$ & $0.798$ & $0.870$ & $0.794$ & $0.775$ & $0.747$ & $0.776$ \\ \midrule
      		LAND & $\b{0.831}$ & $\b{0.716}$ & $\b{0.695}$ & $\b{0.740}$ & $\b{0.818}$ & $\b{0.874}$ & $\b{0.830}$ & $\b{0.783}$ & $\b{0.750}$ & $\b{0.787}$ \\
     \bottomrule
    \end{tabular}
    }
    \label{tab:eeg_results}
\end{table}

% 1 - 1.4 - 1.2 - 1 - 0.6 - 0.9 - 0.6 - 1.3 - 0.9 - 1.2

\section{Related Work}

We are not the first to consider Riemannian normal distributions, e.g.\ \citet{Pennec:JMIV:06} gives a theoretical analysis of the distribution, and \citet{Miaomiao:nips:2013} consider the Riemannian counterpart of probabilistic PCA. Both consider the scenario where the manifold is known a priori. We adapt the distribution to the ``manifold learning'' setting by constructing a Riemannian metric that adapts to the data. This is our overarching contribution.
  
Traditionally, manifold learning is seen as an \emph{embedding problem} where a low-dimensional representation of the data is sought. This is useful for visualization \citep{tenenbaum:global:2000,Roweis00:nonlineardimensionality,Scholkopf99:kernelprincipal,Belkin:2003:LED}, clustering \cite{Luxburg:2007:TSC}, semi-supervised learning \cite{Belkin:2006:MRG} and more. However, in embedding approaches, the relation between a new point and the embedded points are less well-defined, and consequently these approaches are less suited for building generative models. In contrast, the Riemannian approach gives the ability to measure continuous geodesics that follow the structure of the data. This makes the learned Riemannian manifold a suitable space for a generative model.

%\hauberg{Make link to kernel methods.}

\citet{simoserra:bmvc:2014} consider mixtures of Riemannian normal distributions on manifolds that are known a priori. Structurally, their EM algorithm is similar to ours, but they do not account for the normalization constants for different mixture components. Consequently, their approach is inconsistent with the probabilistic formulation. \citet{straub2015dptgmm} consider data on spherical manifolds, and further consider a Dirichlet process prior for determining the number of components. Such a prior could also be incorporated in our model. The key difference to our work is that we consider learned manifolds as well as the following complications.

\section{Discussion}

In this paper we have introduced a parametric locally adaptive normal distribution. The idea is to replace the Euclidean distance in the ordinary normal distribution with a locally adaptive nonlinear distance measure. In principle, we learn a non-parametric metric space, by constructing a smoothly changing metric that induces a Riemannian manifold, where we build our model. As such, we propose a parametric model over a non-parametric space.

The non-parametric space is constructed using a local metric that is the inverse of a local covariance matrix. Here locality is defined via a Gaussian kernel, such that the manifold learning can be seen as a form of kernel smoothing. This indicates that our scheme for learning a manifold might not scale to high-dimensional input spaces. In these cases it may be more practical to learn the manifold probabilistically \cite{Tosi:UAI:2014} or as a mixture of metrics \cite{hauberg:nips:2012}. This is feasible as the LAND estimation procedure is agnostic to the details of the learned manifold as long as exponential and logarithm maps can be evaluated.

Once a manifold is learned, the LAND is simply a Riemannian normal distribution. This is a natural model, but more intriguing, it is a theoretical interesting model since it is the maximum entropy distribution for a fixed mean and covariance \cite{Pennec:JMIV:06}. It is generally difficult to build locally adaptive distributions with maximum entropy properties, yet the LAND does this in a fairly straight-forward manner. This is, however, only a partial truth as the distribution depends on the non-parametric space. The natural question, to which we currently do not have an answer, is whether a suitable maximum entropy manifold exist? 

Algorithmically, we have proposed a maximum likelihood estimation scheme for the LAND. This combines a gradient-based optimization with a scalable Monte Carlo integration method. Once exponential and logarithm maps are available, this procedure is surprisingly simple to implement. We have demonstrated the algorithm on both real and synthetic data and results are encouraging. We almost always improve upon a standard Gaussian mixture model as the LAND is better at capturing the local properties of the data.

We note that both the manifold learning aspect and the algorithmic aspect of our work can be improved. It would be of great value to learn the parameter $\sigma$ used for smoothing the Riemannian metric, and in general, more adaptive learning schemes are of interest. Computationally, the bottleneck of our work is evaluating the logarithm maps. This may be improved by specialized solvers, e.g.\ probabilistic solvers \cite{hennig:aistats:2014}, or manifold-specific heuristics.

The ordinary normal distribution is a key element in many machine learning algorithms. We expect that many fundamental generative models can be extended to the ``manifold'' setting simply by replacing the normal distribution with a LAND. Examples of this idea include Na\"ive Bayes, Linear Discriminant Analysis, Principal Component Analysis and more. Finally we note that standard hypothesis tests also extend to Riemannian normal distributions \cite{Pennec:JMIV:06} and hence also to the LAND.

\textbf{Acknowledgements}. LKH was funded in part by the Novo Nordisk Foundation Interdisciplinary Synergy Program 2014, 'Biophysically adjusted state-informed cortex stimulation (BASICS)'. SH was funded in part by the Danish Council for Independent Research, Natural Sciences.

%\newpage
\clearpage
{\small
\bibliography{land_arxiv_v3}

\begin{thebibliography}{23}
\providecommand{\natexlab}[1]{#1}
\providecommand{\url}[1]{\texttt{#1}}
\expandafter\ifx\csname urlstyle\endcsname\relax
  \providecommand{\doi}[1]{doi: #1}\else
  \providecommand{\doi}{doi: \begingroup \urlstyle{rm}\Url}\fi

\bibitem[Belkin and Niyogi(2003)]{Belkin:2003:LED}
M.~Belkin and P.~Niyogi.
\newblock {Laplacian Eigenmaps for Dimensionality Reduction and Data
  Representation}.
\newblock \emph{Neural Computation}, 15\penalty0 (6):\penalty0 1373--1396, June
  2003.

\bibitem[Belkin et~al.(2006)Belkin, Niyogi, and Sindhwani]{Belkin:2006:MRG}
M.~Belkin, P.~Niyogi, and V.~Sindhwani.
\newblock {Manifold Regularization: A Geometric Framework for Learning from
  Labeled and Unlabeled Examples}.
\newblock \emph{JMLR}, 7:\penalty0 2399--2434, Dec. 2006.

\bibitem[Bishop(2006)]{Bishop:2006:PRM:1162264}
C.~M. Bishop.
\newblock \emph{{Pattern Recognition and Machine Learning (Information Science
  and Statistics)}}.
\newblock Springer-Verlag New York, Inc., Secaucus, NJ, USA, 2006.

\bibitem[Boyce et~al.(2016)Boyce, Glasgow, Williams, and
  Adamantidis]{boyce2016causal}
R.~Boyce, S.~D. Glasgow, S.~Williams, and A.~Adamantidis.
\newblock {Causal evidence for the role of REM sleep theta rhythm in contextual
  memory consolidation}.
\newblock \emph{Science}, 352\penalty0 (6287):\penalty0 812--816, 2016.

\bibitem[Delorme and Makeig(2004)]{Delorme_eeglab:an}
A.~Delorme and S.~Makeig.
\newblock {EEGLAB}: an open source toolbox for analysis of single-trial {EEG}
  dynamics including independent component analysis.
\newblock \emph{J. Neurosci. Methods}, page~21, 2004.

\bibitem[do~Carmo(1992)]{docarmo:1992}
M.~do~Carmo.
\newblock \emph{Riemannian Geometry}.
\newblock Mathematics (Boston, Mass.). Birkh{\"a}user, 1992.

\bibitem[Goldberger et~al.(2000 (June 13))Goldberger, Amaral, Glass, Hausdorff,
  Ivanov, Mark, Mietus, Moody, Peng, and Stanley]{PhysioNet}
A.~L. Goldberger, L.~A.~N. Amaral, L.~Glass, J.~M. Hausdorff, P.~C. Ivanov,
  R.~G. Mark, J.~E. Mietus, G.~B. Moody, C.-K. Peng, and H.~E. Stanley.
\newblock {PhysioBank, PhysioToolkit, and PhysioNet: Components of a New
  Research Resource for Complex Physiologic Signals}.
\newblock \emph{Circulation}, 101\penalty0 (23):\penalty0 e215--e220, 2000
  (June 13).

\bibitem[Hauberg(2016)]{hauberg:tpami:princurve}
S.~Hauberg.
\newblock {Principal Curves on Riemannian Manifolds}.
\newblock \emph{IEEE Transactions on Pattern Analysis and Machine Intelligence
  (TPAMI)}, 2016.

\bibitem[Hauberg et~al.(2012)Hauberg, Freifeld, and Black]{hauberg:nips:2012}
S.~Hauberg, O.~Freifeld, and M.~J. Black.
\newblock {A Geometric Take on Metric Learning}.
\newblock In \emph{Advances in Neural Information Processing Systems (NIPS)
  25}, pages 2033--2041, 2012.

\bibitem[Hennig and Hauberg(2014)]{hennig:aistats:2014}
P.~Hennig and S.~Hauberg.
\newblock {Probabilistic Solutions to Differential Equations and their
  Application to Riemannian Statistics}.
\newblock In \emph{Proceedings of the 17th international Conference on
  Artificial Intelligence and Statistics (AISTATS)}, volume~33, 2014.

\bibitem[Imtiaz and Rodriguez-Villegas(2015)]{imtiaz:01}
S.~A. Imtiaz and E.~Rodriguez-Villegas.
\newblock {An open-source toolbox for standardized use of PhysioNet Sleep EDF
  Expanded Database}.
\newblock In \emph{2015 37th Annual International Conference of the IEEE
  Engineering in Medicine and Biology Society (EMBC)}, pages 6014--6017, Aug
  2015.

\bibitem[Karcher(1977)]{karcher:1977}
H.~Karcher.
\newblock Riemannian center of mass and mollifier smoothing.
\newblock \emph{Communications on Pure and Applied Mathematics}, 30\penalty0
  (5):\penalty0 509--541, 1977.

\bibitem[Luxburg(2007)]{Luxburg:2007:TSC}
U.~Luxburg.
\newblock {A Tutorial on Spectral Clustering}.
\newblock \emph{Statistics and Computing}, 17\penalty0 (4):\penalty0 395--416,
  Dec. 2007.

\bibitem[Marxer et~al.(2008)Marxer, Purwins, and Hazan]{marxer2008f}
R.~Marxer, H.~Purwins, and A.~Hazan.
\newblock An f-measure for evaluation of unsupervised clustering with
  non-determined number of clusters.
\newblock \emph{Report of the EmCAP project (European Commission FP6-IST)},
  pages 1--3, 2008.

\bibitem[Pennec(2006)]{Pennec:JMIV:06}
X.~Pennec.
\newblock {Intrinsic Statistics on {R}iemannian Manifolds: Basic Tools for
  Geometric Measurements}.
\newblock \emph{Journal of Mathematical Imaging and Vision}, 25\penalty0
  (1):\penalty0 127--154, July 2006.

\bibitem[Purves et~al.()Purves, Augustine, Fitzpatrick, Hall, LaMantia,
  McNamara, and White]{purvesneuroscience}
D.~Purves, G.~Augustine, D.~Fitzpatrick, W.~Hall, A.~LaMantia, J.~McNamara, and
  L.~White.
\newblock \emph{Neuroscience, 2008}.
\newblock De Boeck, Sinauer, Sunderland, Mass.

\bibitem[Roweis and Saul(2000)]{Roweis00:nonlineardimensionality}
S.~T. Roweis and L.~K. Saul.
\newblock Nonlinear dimensionality reduction by locally linear embedding.
\newblock \emph{Science}, 290:\penalty0 2323--2326, 2000.

\bibitem[Sch{\"o}lkopf et~al.(1999)Sch{\"o}lkopf, Smola, and
  Müller]{Scholkopf99:kernelprincipal}
B.~Sch{\"o}lkopf, A.~Smola, and K.-R. Müller.
\newblock Kernel principal component analysis.
\newblock In \emph{Advances in Kernel Methods - Support Vector Learning}, pages
  327--352, 1999.

\bibitem[Simo-Serra et~al.(2014)Simo-Serra, Torras, and
  Moreno-Noguer]{simoserra:bmvc:2014}
E.~Simo-Serra, C.~Torras, and F.~Moreno-Noguer.
\newblock {G}eodesic {F}inite {M}ixture {M}odels.
\newblock In \emph{Proceedings of the British Machine Vision Conference}. BMVA
  Press, 2014.

\bibitem[Straub et~al.(2015)Straub, Chang, Freifeld, and
  Fisher~III]{straub2015dptgmm}
J.~Straub, J.~Chang, O.~Freifeld, and J.~W. Fisher~III.
\newblock A {D}irichlet {P}rocess {M}ixture {M}odel for {S}pherical {D}ata.
\newblock In \emph{International Conference on Artificial Intelligence and
  Statistics (AISTATS)}, 2015.

\bibitem[Tenenbaum et~al.(2000)Tenenbaum, de~Silva, and
  Langford]{tenenbaum:global:2000}
J.~B. Tenenbaum, V.~de~Silva, and J.~C. Langford.
\newblock {A Global Geometric Framework for Nonlinear Dimensionality
  Reduction}.
\newblock \emph{Science}, 290\penalty0 (5500):\penalty0 2319, 2000.

\bibitem[Tosi et~al.(2014)Tosi, Hauberg, Vellido, and Lawrence]{Tosi:UAI:2014}
A.~Tosi, S.~Hauberg, A.~Vellido, and N.~D. Lawrence.
\newblock {Metrics for Probabilistic Geometries}.
\newblock In \emph{The Conference on Uncertainty in Artificial Intelligence
  (UAI)}, July 2014.

\bibitem[Zhang and Fletcher(2013)]{Miaomiao:nips:2013}
M.~Zhang and P.~Fletcher.
\newblock {Probabilistic Principal Geodesic Analysis}.
\newblock In \emph{Advances in Neural Information Processing Systems (NIPS)
  26}, pages 1178--1186, 2013.

\end{thebibliography}
}

\clearpage
\appendix
\section*{Appendix}

\textbf{Notation}: all points $\b{x}\in\R^D$ are considered as column vectors, and they are denoted with bold lowercase characters. $\S_{++}^D$ represents the set of symmetric $D\times D$ positive definite matrices. The learned Riemannian manifold is denoted $\M$, and its tangent space at point $\b{x}\in\M$ is denoted $\tangent{\bs{\mu}}$.

We present for convenience the domain and co-domain of the following often used terms. Note that $\tangent{\bs{\mu}}$ is $\R^D$.

\begin{table}[!ht]
\setlength{\tabcolsep}{20pt}
\centering
\begin{tabular}{l l}
$\bs{\gamma}(t):[0,1] \rightarrow \M$ & $\Exp{\bs{\mu}}{\b{v}}:\M \times \tangent{\b{x}} \rightarrow \M$ \\ \rule{0pt}{3ex}

$\b{M}(\b{x}):\M \rightarrow \S^D_{++}$ & $\Log{\bs{\mu}}{\b{x}}:\M \times \M \rightarrow \tangent{\b{x}}$ 
\end{tabular}
\end{table}

\section{Estimating the Normalization Constant}

The locally adaptive normal distribution is defined as
\begin{equation}
p_{\M}(\b{x} ~|~ \bs{\mu},\bs{\Sigma}) = \frac{1}{\C(\bs{\mu},\bs{\Sigma})}\exp\left(-\frac{1}{2} \inner{\Log{\bs{\mu}}{\b{x}}}{\bs{\Sigma}^{-1}\Log{\bs{\mu}}{\b{x}}}\right), \quad \b{x} \in \M.
\end{equation}
Therefore, the normalization constant is equal to
\begin{align}
\label{eq:normalization_constant}
& \int_{\M} p_{\M}(\b{x} ~|~\bs{\mu},\bs{\Sigma}) \dif{d} \M(\b{x}) = 1 \Rightarrow \\
& \int_{\M} \frac{1}{\C(\bs{\mu},\bs{\Sigma})}\exp\left(-\frac{1}{2} \inner{\Log{\bs{\mu}}{\b{x}}}{\bs{\Sigma}^{-1}\Log{\bs{\mu}}{\b{x}}}\right) \dif{d} \M(\b{x}) = 1
\Rightarrow \\
\C(\bs{\mu},\bs{\Sigma}) = &\int_{\M} \exp\left(-\frac{1}{2} \inner{\Log{\bs{\mu}}{\b{x}}}{\bs{\Sigma}^{-1}\Log{\bs{\mu}}{\b{x}}}\right) \dif{d} \M(\b{x})\\
= &\int_{\D(\bs{\mu})} \sqrt{\abs{\b{M}(\Exp{\bs{\mu}}{\b{v}})}} \exp\left(-\frac{1}{2} \inner{\Log{\bs{\mu}}{\Exp{\bs{\mu}}{\b{v}}}}{\bs{\Sigma}^{-1}\Log{\bs{\mu}}{\Exp{\bs{\mu}}{\b{v}}}}\right) \dif{d} \b{v}\\
=&\int_{\tangent{\bs{\mu}}} m(\bs{\mu},\b{v}) \exp\left(-\frac{1}{2} \inner{\b{v}}{\bs{\Sigma}^{-1}\b{v}}\right) \dif{d} \b{v} \\
= &\int_{\tangent{\bs{\mu}}} m(\bs{\mu},\b{v}) \frac{\Z}{\Z} \exp\left(-\frac{1}{2} \inner{\b{v}}{\bs{\Sigma}^{-1}\b{v}}\right) \dif{d} \b{v}\\
=&\Z\cdot \E_{\N(0,\bs{\Sigma})}[m(\bs{\mu},\b{v})] \simeq \frac{\Z}{S} \sum_{s=1}^S m(\bs{\mu},\b{v}_s), \qquad \text{where} \quad \b{v}_s \sim \N(0,\bs{\Sigma}).
\end{align}
To simplify notation we have define the $m(\bs{\mu},\b{v}) = \sqrt{\abs{\b{M}(\Exp{\bs{\mu}}{\b{v}})}}$ and $\Z = \sqrt{(2\pi)^D \abs{\bs{\Sigma}}}$. The integral is then estimated with a Monte-Carlo technique.

\clearpage

\section{Steepest Descent Direction for the Mean} \label{appendix:gradient_mu}

The objective function is differentiable with respect to $\bs{\mu}$ with
\begin{align}
\parder{}{\bs{\mu}} \inner{\Log{\bs{\mu}}{\b{x}_n}}{\bs{\Sigma}^{-1}\Log{\bs{\mu}}{\b{x}_n}} = -2 \bs{\Sigma}^{-1}\Log{\bs{\mu}}{\b{x}_n}
\end{align}
Then the gradient of the objective function $\phi(\bs{\mu},\bs{\Sigma})$ is equal to
\begin{align}
&\grad{\bs{\mu}}\phi(\bs{\mu},\bs{\Sigma}) = %\\
\parder{}{\bs{\mu}} \left[ \frac{1}{2N} \sum_{n=1}^N \inner{\Log{\bs{\mu}}{\b{x}_n}}{\bs{\Sigma}^{-1}\Log{\bs{\mu}}{\b{x}_n}}
+
\log(\C(\bs{\mu},\bs{\Sigma}))
\right]\\
& = -\frac{1}{N} \bs{\Sigma}^{-1} \sum_{n=1}^N \Log{\bs{\mu}}{\b{x}_n} + \frac{1}{\C(\bs{\mu},\bs{\Sigma})}\int_{\M} \parder{}{\bs{\mu}} \left[   \exp\left(-\frac{1}{2} \inner{\Log{\bs{\mu}}{\b{x}}}{\bs{\Sigma}^{-1}\Log{\bs{\mu}}{\b{x}}} \right) \right]\dif{d} \M(\b{x}) \\
& = -\frac{\bs{\Sigma}^{-1} }{N} \sum_{n=1}^N \Log{\bs{\mu}}{\b{x}_n} + \frac{ \bs{\Sigma}^{-1}}{\C(\bs{\mu},\bs{\Sigma})} \int_{\M} \Log{\bs{\mu}}{\b{x}} \exp\left(-\frac{1}{2} \inner{\Log{\bs{\mu}}{\b{x}}}{\bs{\Sigma}^{-1}\Log{\bs{\mu}}{\b{x}}} \right) \dif{d} \M(\b{x}) \\
& = -\frac{1}{N} \bs{\Sigma}^{-1} \sum_{n=1}^N \Log{\bs{\mu}}{\b{x}_n} + \frac{ \bs{\Sigma}^{-1}}{\C(\bs{\mu},\bs{\Sigma})} \int_{\tangent{\bs{\mu}}} m(\bs{\mu},\b{v})\b{v} \exp\left(-\frac{1}{2} \inner{\b{v}}{\bs{\Sigma}^{-1}\b{v}} \right) \dif{d} \b{v} \\
& = -\bs{\Sigma}^{-1}\left[ \frac{1}{N}\sum_{n=1}^N \Log{\bs{\mu}}{\b{x}_n} - \frac{\Z}{\C(\bs{\mu},\bs{\Sigma})\cdot S}\sum_{s=1}^S m(\bs{\mu},\b{v}_s) \b{v}_s\right].
\label{eq:gradient_mu}
\end{align}

This gradient is highly dependent on the condition number of the covariance matrix $\bs{\Sigma}$, which makes the gradient unstable. We therefore consider the steepest descent direction.

We start by showing the general steepest descent direction.

\begin{align} 
\b{d}^*
&= \argmin_{\b{d}\in\R^D} \{ \inner{\grad{\bs{\mu}}\phi}{\b{d}} ~|~ \norm{\b{d}}_{M} = 1 \} \qquad \quad \quad \quad \quad \inner{\b{d}}{M\b{d}} = 1 \Rightarrow M = A^\t A\\
&= A^{-1} \argmin_{\b{x}\in\R^D} \{ \inner{\grad{\bs{\mu}}\phi}{A^{-1}\b{x}} ~|~ \norm{\b{x}}_2 = 1 \} \quad \quad \quad \inner{A\b{d}}{A\b{d}} = 1 \Rightarrow \b{x} = A\b{d}\\
\label{eq:steepest_optimization}
&=A^{-1} \argmin_{\b{x}\in\R^D} \{ \inner{A^{-\t}\grad{\bs{\mu}}\phi }{\b{x}} ~|~ \norm{\b{x}}_2 = 1\}. \quad \quad \quad \inner{\b{x}}{\b{x}} = 1 ~\text{and}~ \b{d} = A^{-1} \b{x}\\
\end{align}

Using the Cauchy-Schwarz inequality $(-\norm{\b{x}}_2 \norm{\b{y}}_2 \leq \inner{\b{x}}{\b{y}})$ for the optimization problem (\ref{eq:steepest_optimization}), we get that the minimizer is equal to
\begin{equation}
\label{eq:steepest_minimizer}
-\norm{A^{-\t}\grad{\bs{\mu}}\phi}_2 \norm{\b{x}}_2 \leq \inner{A^{-\t}\grad{\bs{\mu}}\phi}{\b{x}} \Rightarrow \b{x}^* = -\frac{A^{-\t}\grad{\bs{\mu}}\phi}{\norm{A^{-\t}\grad{\bs{\mu}}\phi}_2},
\end{equation}
and thus, by plugging the result of (\ref{eq:steepest_minimizer}) in to (\ref{eq:steepest_optimization}), we get that the steepest descent direction is
\begin{equation}
\label{eq:steepest_descent_general}
\begin{split}
\b{d}^* = -\frac{A^{-1}A^{-\t}\grad{\bs{\mu}}\phi }{ \norm{A^{-1}A^{-\t}\grad{\bs{\mu}}\phi}_2} &= -\frac{(A^\t A)^{-1}\grad{\bs{\mu}}\phi }{\sqrt{\inner{A^{-\t}\grad{\bs{\mu}}\phi}{\inner{A^{-\t}\grad{\bs{\mu}}\phi }} }} = -\frac{M^{-1}\grad{\bs{\mu}}\phi}{\sqrt{\inner{\grad{\bs{\mu}}\phi}{M^{-1}\grad{\bs{\mu}}\phi}}} \\
\Rightarrow \b{d}^* &= -\frac{M^{-1}\grad{\bs{\mu}}\phi}{\norm{M^{-1}\grad{\bs{\mu}}\phi}_M }.
\end{split}
\end{equation}

In our case, the $M = \bs{\Sigma}^{-1}$, and thus, we get that the steepest descent direction of the objective function of the LAND model is
\begin{equation}
\label{eq:steepest_descent}
\b{d}^* = \frac{1}{N} \sum_{n=1}^N \Log{\bs{\mu}}{\b{x}_n}  - \frac{\Z}{\C(\bs{\mu},\bs{\Sigma})\cdot S} \sum_{s=1}^S m(\bs{\mu},\b{v}_s)\b{v}_s,
\end{equation}
where we omit the denominator $\norm{\grad{\bs{\mu}}\phi(\bs{\mu},\bs{\Sigma})}_2$, since this is just a scaling factor, which will be captured by the stepsize. This avoid problems that appears due to large condition numbers of $\bs{\Sigma}$.

\section{Gradient Direction for the Covariance} \label{appendix:gradient_Sigma}

We decompose the $\bs{\Sigma}^{-1} = \b{A}^\t \b{A}$. In addition, we rewrite the inner product as follows
\begin{align}
\inner{\Log{\bs{\mu}}{\b{x}_n}}{\b{A}^\t\b{A} \Log{\bs{\mu}}{\b{x}_n}} =& ~tr(\Log{\bs{\mu}}{\b{x}_n}^\t \b{A}^\t\b{A} \Log{\bs{\mu}}{\b{x}_n}) \\
= &~tr(\b{A} \Log{\bs{\mu}}{\b{x}_n} \Log{\bs{\mu}}{\b{x}_n}^\t \b{A}^\t) \\ 
\Rightarrow \parder{}{\b{A}} [ tr(\b{A} \Log{\bs{\mu}}{\b{x}_n} &\Log{\bs{\mu}}{\b{x}_n}^\t \b{A}^\t) ] = 2 \b{A} \Log{\bs{\mu}}{\b{x}_n}\Log{\bs{\mu}}{\b{x}_n}^\t,
\end{align}
where $tr(\cdot)$ is the trace operator. Then the gradient of the objective with respect the matrix $\b{A}$ is

\begin{align}
\grad{\b{A}}\phi(\bs{\mu},\bs{\Sigma}) 
& =\parder{}{\b{A}} \left[ \frac{1}{2N}\sum_{n=1}^N \inner{\Log{\bs{\mu}}{\b{x}_n}}{\b{A}^\t\b{A}\Log{\bs{\mu}}{\b{x}_n}} + \log(\C(\bs{\mu},\bs{\Sigma}))\right]\\
& = \frac{1}{2N}2 \b{A}\sum_{n=1}^N \Log{\bs{\mu}}{\b{x}_n} \Log{\bs{\mu}}{\b{x}_n}^\t \\
&+ \frac{1}{\C(\bs{\mu},\bs{\Sigma})}\int_{\M} \parder{}{\b{A}}\left[ \exp\left(-\frac{1}{2} \inner{\Log{\bs{\mu}}{\b{x}}}{\b{A}^\t\b{A}\Log{\bs{\mu}}{\b{x}}}\right) \right] \dif{d}\M(\b{x})\\
& = \frac{1}{N}\b{A}\sum_{n=1}^N \Log{\bs{\mu}}{\b{x}_n} \Log{\bs{\mu}}{\b{x}_n}^\t \\
& - \frac{\b{A} }{\C(\bs{\mu},\bs{\Sigma})} \int_{\M} \Log{\bs{\mu}}{\b{x}}\Log{\bs{\mu}}{\b{x}}^\t \exp\left(-\frac{1}{2} \inner{\Log{\bs{\mu}}{\b{x}}}{\b{A}^\t\b{A}\Log{\bs{\mu}}{\b{x}}}\right) \dif{d}\M(\b{x})\\
& = \frac{1}{N}\b{A}\sum_{n=1}^N \Log{\bs{\mu}}{\b{x}_n} \Log{\bs{\mu}}{\b{x}_n}^\t  \\
& - \frac{\b{A}}{\C(\bs{\mu},\bs{\Sigma})} \int_{\tangent{\bs{\mu}}} m(\bs{\mu},\b{v}) \b{v}\b{v}^\t \exp\left(-\frac{1}{2} \inner{\b{v}}{\bs{\Sigma}^{-1}\b{v}} \right) \dif{d}\b{v}.
\end{align}

Finally, treating the integral as an expectation problem and using Monte Carlo integration, we get that the gradient is
\begin{equation}
\label{eq:grad_A}
\grad{\b{A}}\phi(\bs{\mu},\bs{\Sigma}) = \b{A}\left[ \frac{1}{N} \sum_{n=1}^N \Log{\bs{\mu}}{\b{x}_n} \Log{\bs{\mu}}{\b{x}_n}^\t - \frac{\Z}{\C(\bs{\mu},\bs{\Sigma})\cdot S} \sum_{s=1}^S m(\bs{\mu},\b{v}_s)\b{v}_s\b{v}_s^\t
\right].
\end{equation}

\section{Gradients for the LAND Mixture Model}

Similarly the LAND mixture model are
\begin{align}
&\grad{\bs{\mu}_k}\psi(\bs{\Theta}) = -\bs{\Sigma}^{-1}_k\left[ \sum_{n=1}^N r_{nk} \Log{\bs{\mu}_k}{\b{x}_n} - \frac{\Z \cdot R_k }{\C_k(\bs{\mu}_k,\bs{\Sigma}_k)\cdot S} \sum_{s=1}^S m(\bs{\mu}_k,\b{v}_s) \b{v}_s \right] \label{eq:grad_mu_mm} \\
&\grad{\b{A}_k}\psi(\bs{\Theta})  = \b{A}_k \left[ \sum_{n=1}^N r_{nk} \Log{\bs{\mu}_k}{\b{x}_n} \Log{\bs{\mu}_k}{\b{x}_n}^\t  - \frac{\Z\cdot R_k}{\C_k(\bs{\mu}_k,\bs{\Sigma}_k)\cdot S} \sum_{s=1}^S m(\bs{\mu}_k,\b{v}_s)\b{v}_s \b{v}_s^\t \right] \label{eq:grad_A_mm}
\end{align}
where $R_k = \sum_{n=1}^N r_{nk}$, and the responsibilities $r_{nk} = \frac{\pi_k p_{\M}(\b{x}_n ~|~\bs{\mu}_k,\bs{\Sigma}_k)}{\sum_{l=1}^K \pi_l p_{\M}(\b{x}_n ~|~\bs{\mu}_l,\bs{\Sigma}_l)}$.

\newpage

\section{Algorithms} \label{appendix:algorithms}

In this section we present the algorithms for: 1) estimating the normalization constant $\C(\bs{\mu},\bs{\Sigma})$, 2) maximum likelihood estimation of the LAND, and 3) fitting the LAND mixture model.

\begin{algorithm}[!ht]
	\caption{The estimation of the normalization constant $\C(\bs{\mu},\bs{\Sigma})$}
      \algsetup{indent=2em}
        \begin{algorithmic}[1]
         	\REQUIRE{the given data $\{\b{x}_n\}_{n=1}^N$, the $\bs{\mu},~\bs{\Sigma}$, the number of samples $S$}
         	\ENSURE {the estimated $\hat{\C}({\bs{\mu}},{\bs{\Sigma}})$}
         	\STATE{sample $S$ tangent vectors $\b{v}_s \sim \N(0,\bs{\Sigma})$ on $\tangent{\bs{\mu}}$}
			 \STATE{map the $\b{v}_s$ on $\M$ as $\b{x}_s = \Exp{\bs{\mu}}{\b{v}_s},~ s=1,\dots,S$}
			 \STATE{compute the normalization constant $\hat{\C}(\bs{\mu},\bs{\Sigma}) = \frac{\Z}{S} \sum_{s=1}^S \sqrt{ \abs{\b{M} ( \b{x}_s ) }} $}
        \end{algorithmic}
\end{algorithm}

\begin{algorithm}[!ht]
	\caption{LAND maximum likelihood}
      \algsetup{indent=2em}
        \begin{algorithmic}[1]
         	\REQUIRE{the data $\{\b{x}_n\}_{n=1}^N$, stepsize $\alpha_{\bs{\mu}},\alpha_A$, tolerance $\epsilon$}
         	\ENSURE {the estimated $\hat{\bs{\mu}},~\hat{\bs{\Sigma}},~\hat{\C}(\hat{\bs{\mu}},\hat{\bs{\Sigma}})$}
         	\STATE{initialize $\bs{\mu}^0,\bs{\Sigma}^0$ and $t\leftarrow0$}
			 \REPEAT
			 	\STATE{estimate $\C(\bs{\mu}^t,\bs{\Sigma}^t)$ using Eq.~\ref{eq:normalization_constant} }
			 \STATE{compute $d_{\bs{\mu}}\phi(\bs{\mu}^t,\bs{\Sigma}^t)$ using Eq.~\ref{eq:steepest_descent}}
			 \STATE{$\bs{\mu}^{t+1} \leftarrow \Exp{\bs{\mu}^t}{\alpha_{\bs{\mu}}d_{\bs{\mu}}\phi(\bs{\mu}^t,\bs{\Sigma}^t) }$}
			\STATE{estimate $\C(\bs{\mu}^{t+1},\bs{\Sigma}^t)$ using Eq.~\ref{eq:normalization_constant} }
			\STATE{compute $\grad{\b{A}}\phi(\bs{\mu}^{t+1},\bs{\Sigma}^t)$ using Eq.~\ref{eq:grad_A}}
			\STATE{$\b{A}^{t+1} \leftarrow \b{A} - \alpha_\b{A} \grad{\b{A}}\phi(\bs{\mu}^{t+1},\bs{\Sigma}^t)$}
			\STATE{$\bs{\Sigma}^{t+1} \leftarrow [(\b{A}^{t+1})^\t \b{A}^{t+1}|^{-1}$ } 
			\STATE{$t \leftarrow t+1$}
			 \UNTIL{$\norm{\phi(\bs{\mu}^{t+1},\bs{\Sigma}^{t+1}) - \phi(\bs{\mu}^{t},\bs{\Sigma}^{t})}_2^2 \leq \epsilon$}
        \end{algorithmic}
\end{algorithm}

\begin{algorithm}[!ht]
	\caption{LAND mixture model}
      \algsetup{indent=2em}
        \begin{algorithmic}[1]
         	\REQUIRE{the data $\{\b{x}_n\}_{n=1}^N$, $\{\alpha_{\bs{\mu}_k},\alpha_{\b{A}_k}\}_{k=1}^K$, tolerance $\epsilon$}
         	\ENSURE {the estimated $\{ \hat{\bs{\mu}}_k,~\hat{\bs{\Sigma}}_k,~\hat{\C}_k,~\hat{\pi}_k \}_{k=1}^K$}
         	\STATE{initialize the $\{ \bs{\mu}^0_k,~\bs{\Sigma}^0_k,~\C^0_k,~\pi^0_k \}_{k=1}^K$ and $t\leftarrow 0$}
			 \REPEAT
			 	\STATE{\textbf{Expectation step}:}
			 	\STATE{compute the responsibilities $r_{nk} = \frac{\pi_k p_{\M}( \b{x}_n ~|~ \bs{\mu}_k,~\bs{\Sigma}_k)}{\sum_{t = 1}^K \pi_t p_{\M}( \b{x}_n ~|~ \bs{\mu}_t,~\bs{\Sigma}_t)}$ }
			 	\STATE{\textbf{Maximization step}:}
			 	\FOR{$k=1,\dots,K$}
			 	\STATE{estimate $\C_k(\bs{\mu}_k^t,\bs{\Sigma}_k^t)$ using Eq.~\ref{eq:normalization_constant} }
			 \STATE{compute from Eq.~\ref{eq:grad_mu_mm} the $d_{\bs{\mu}}\phi(\bs{\mu}^t_k,\bs{\Sigma}^t_k)$}
			 \STATE{$\bs{\mu}^{t+1}_k \leftarrow \Exp{\bs{\mu}^t_k}{\alpha_{\bs{\mu}_k} d_{\bs{\mu}}\phi(\bs{\mu}^t_k,\bs{\Sigma}^t_k)}$}
			 	\STATE{estimate $\C_k(\bs{\mu}_k^t,\bs{\Sigma}_k^t)$ using Eq.~\ref{eq:normalization_constant} }
			\STATE{compute from Eq.~\ref{eq:grad_A_mm} the $\grad{\b{A}_k}\phi(\bs{\mu}^{t+1}_k,\bs{\Sigma}^t_k)$}
			\STATE{$\b{A}^{t+1}_k \leftarrow \b{A}^t_k - \alpha_{\b{A}_k} \grad{\b{A}}\phi(\bs{\mu}_k^{t+1},\bs{\Sigma}_k^t)$}
			\STATE{$\bs{\Sigma}_k^{t+1} \leftarrow [(\b{A}_k^{t+1})^\t \b{A}_k^{t+1}]^{-1}$}
			 \STATE{$\pi_k = \frac{1}{N}\sum_{n=1}^N r_{nk}$} 
			\ENDFOR
			\STATE{$t \leftarrow t+1$}
			 \UNTIL{$\norm{\psi(\bs{\Theta}^{t+1}) -\psi(\bs{\Theta}^{t}) }_2^2 \leq \epsilon$}
        \end{algorithmic}
\end{algorithm}

\subsection{Stepsize Selection}

The LAND objective is expensive to evaluate due to the dependency on $\Log{\mu}{\b{x}_n}$. This imply that a line-search is infeasible for selecting a stepsize. Thus, we use the following common trick. Each stepsize is given in the start of the algorithm. If the objective increased after an update, we reduce the corresponding stepsize as $\alpha = 0.75\cdot \alpha$, and if the objective reduced, then $\alpha = 1.1\cdot \alpha$.

\subsection{Initialization Issues} \label{appendix:initialization}

The initialization of the LAND is important, as well as for the mixture model. We discuss two different initializations plus one specifically for the mixture model.

\begin{enumerate}

\item \textbf{Random}: we initialize the LAND mean with a random point on the manifold. The initial covariance is the empirical covariance of the tangent vectors. This initialization can be used also for the mixture model, with $K$ random starting points. Then, we cluster the points and the covariances are initialized using empirical estimators.

\item \textbf{Least Squares}: we initialize the LAND with the intrinsic least squares mean, and the covariance with the empirical estimator. This initialization can be used also for the mixture model, using the extension of the $k$-means on Riemannian manifolds, and then, the points of each cluster for the empirical covariances.

\item \textbf{GMM}: we initialize the LAND mixture model centres with the result of the GMM. For the empirical covariances, we use the points that belong to each cluster from the GMM solution.

\end{enumerate}

\subsection{Stopping criterion}

Our objective function is non-convex, thus, as stopping criterion we use the change of the objective value. In particular, we stop the optimization when $\norm{\phi(\bs{\mu}^{t+1},\bs{\Sigma}^{t+1}) - \phi(\bs{\mu}^{t},\bs{\Sigma}^{t})}_2^2 \leq \epsilon$, for some $\epsilon$ given by the user. The same stopping criterion is used for the mixture model.

\section{Experiments} \label{appendix:experiments}

In this section we provide additional illustrative experiments.

\subsection{Estimating the Normalization Constant} \label{appendix:constant_estimation}

\begin{figure}[!ht]
\centering
	\includegraphics[scale=0.3]{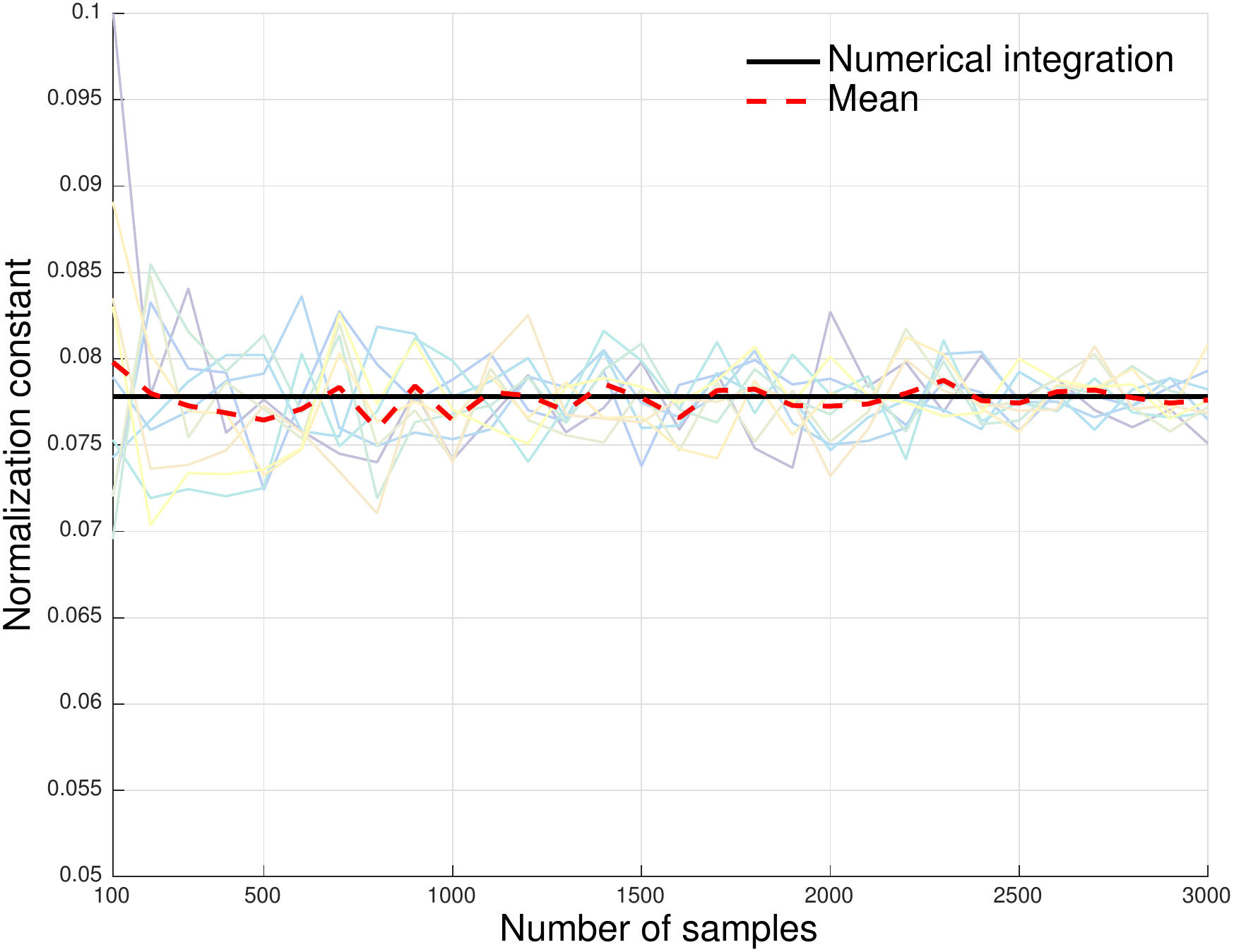}
	\caption{The normalization constant estimation for different sample sizes $S$. The black line denotes the trapezoidal numerical integral, and the dashed red line the mean value of the estimators using our proposed method.}
	\label{fig:c_behaviour}
\end{figure}

In order to show the consistency of the normalization constant estimation with respect to the number of samples, we conduct the following experiment. We used the data from the first synthetic experiment of the paper. Then for a grid $100\times 100$ on the $\tangent{\bs{\mu}}$ we computed the corresponding $m(\bs{\mu},\b{v})$ values. Thus, we computed the numerical integral on the tangent space using trapezoidal numerical integration. Then we estimated the normalization constant using our approach for sample sizes $S = 100:100:3000$ and for 10 different runs. From the result in Fig.~\ref{fig:c_behaviour} we observe that the numerical scheme we provide, approximates well the normalization constant that we computed numerically.

\subsection{MNIST digit 1 data}

\begin{figure}[!ht]
    \centering
    \begin{subfigure}{0.31\textwidth}
        \includegraphics[width=\textwidth]{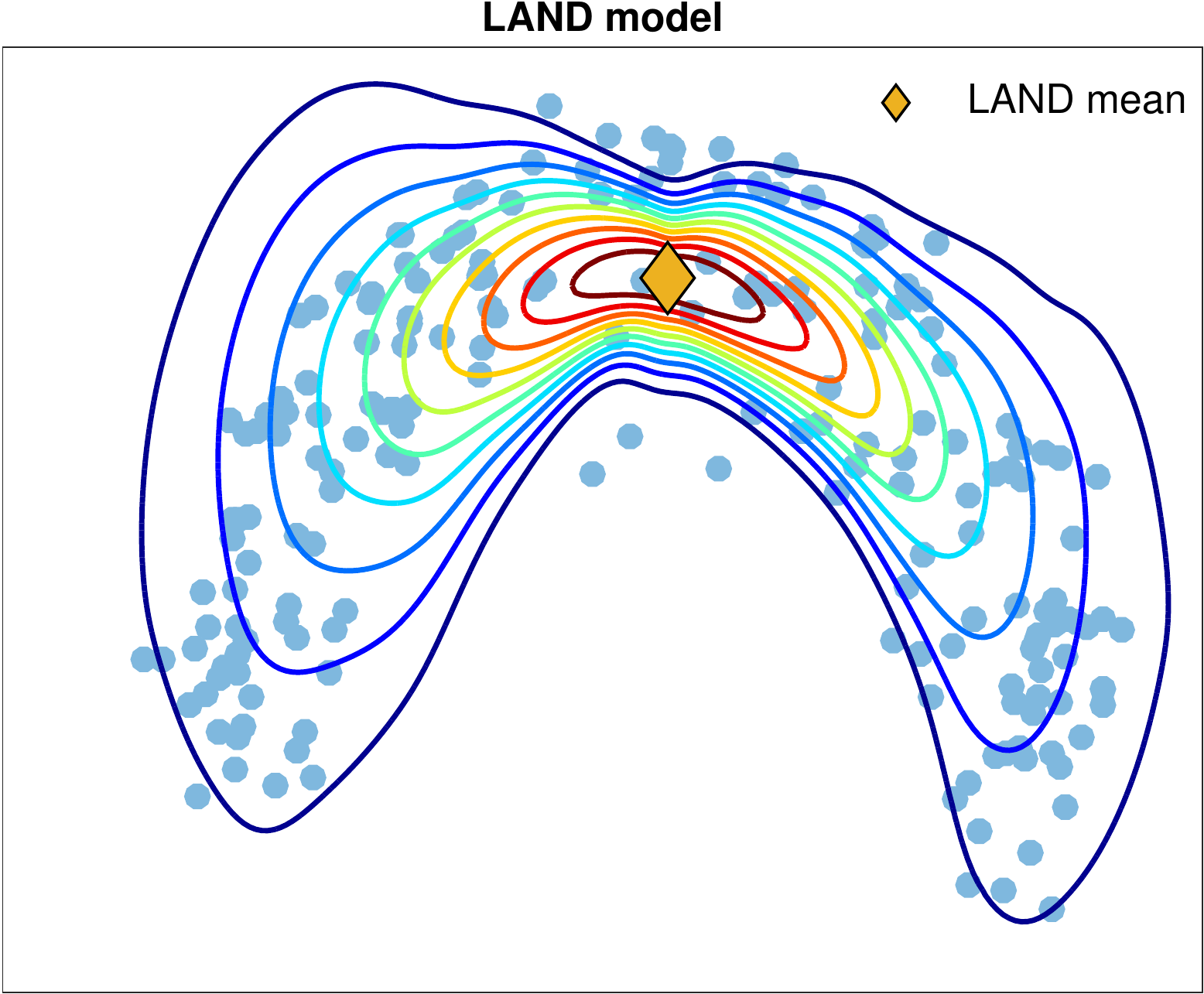}
    \end{subfigure}
    \quad
    \begin{subfigure}{0.31\textwidth}
        \includegraphics[width=\textwidth]{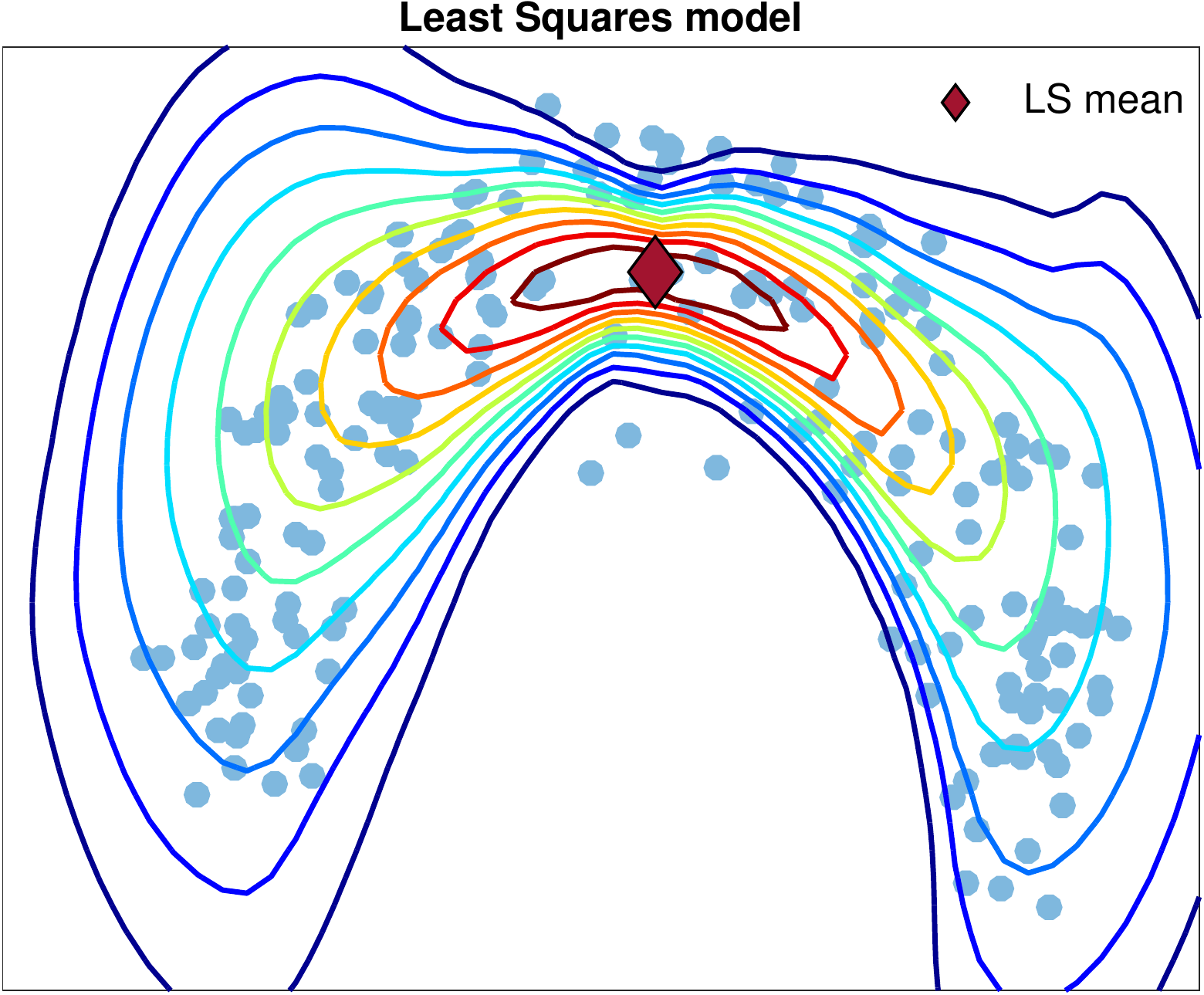}
    \end{subfigure}
    \quad
    \begin{subfigure}{0.31\textwidth}
        \includegraphics[width=\textwidth]{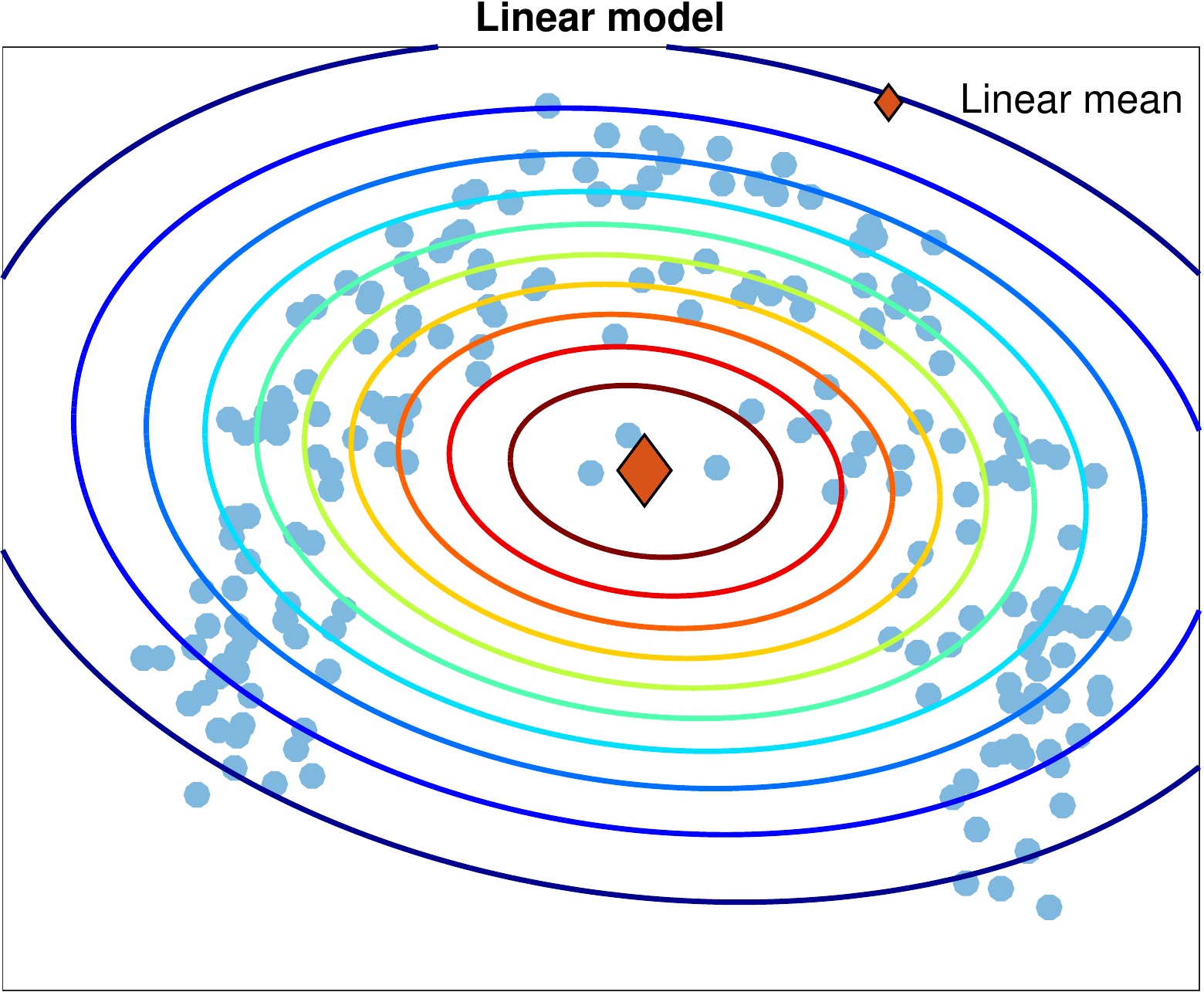}
    \end{subfigure}
    \caption{The MNIST digit 1 projected onto the 2 first principal components experiment. \textit{Left}: the LAND model approximates efficiently the data distribution. \textit{Center}: the least squares model approximates the distribution, but it overfits the given data. \textit{Right}: the normal distribution has poor performance due to the linear distance measure.}
	\label{fig:mnist}    
\end{figure}

In this experiment we used the digit 1 from the MNIST dataset. We sample 200 points and using PCA we projected them onto the first 2 principal components. Then we fitted LAND, a least squares model, and a normal distribution. From the result Fig.~\ref{fig:mnist} we observe that the LAND model approximates efficiently the underlying distribution of the data. Also, the least squares model has a similar performance, since it takes under consideration the underlying manifold. However, is obvious that it overfits the given data, and gives significant probability to low density areas. On the other hand, the linear model has poor performance, due to the linear distance measure.

\subsection{The Sleep Stages Experiment}

\begin{figure}[!ht]
\centering
	\begin{subfigure}{0.31\textwidth}
		\includegraphics[width=\textwidth]{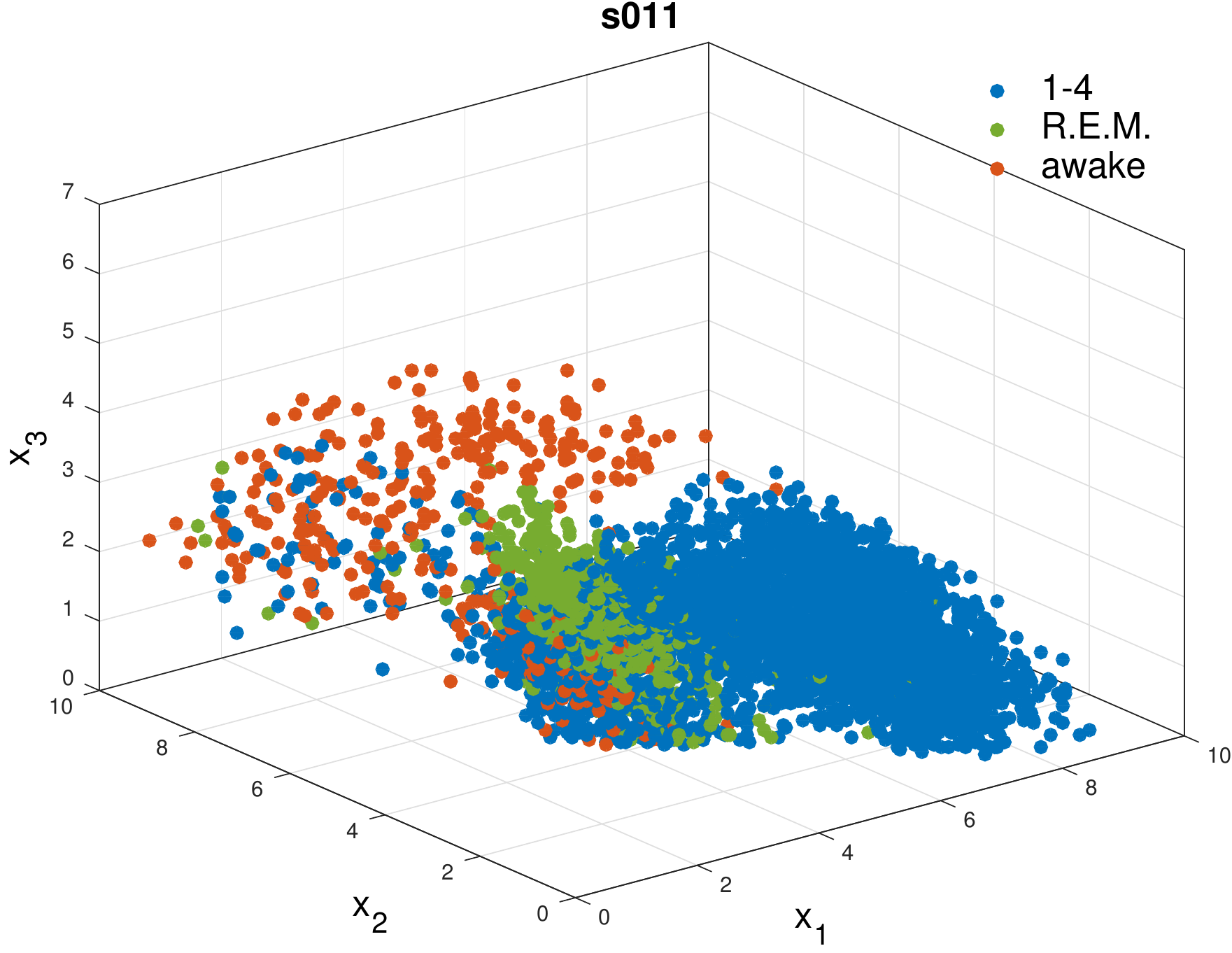}
	\end{subfigure}
	~
	\begin{subfigure}{0.31\textwidth}
		\includegraphics[width=\textwidth]{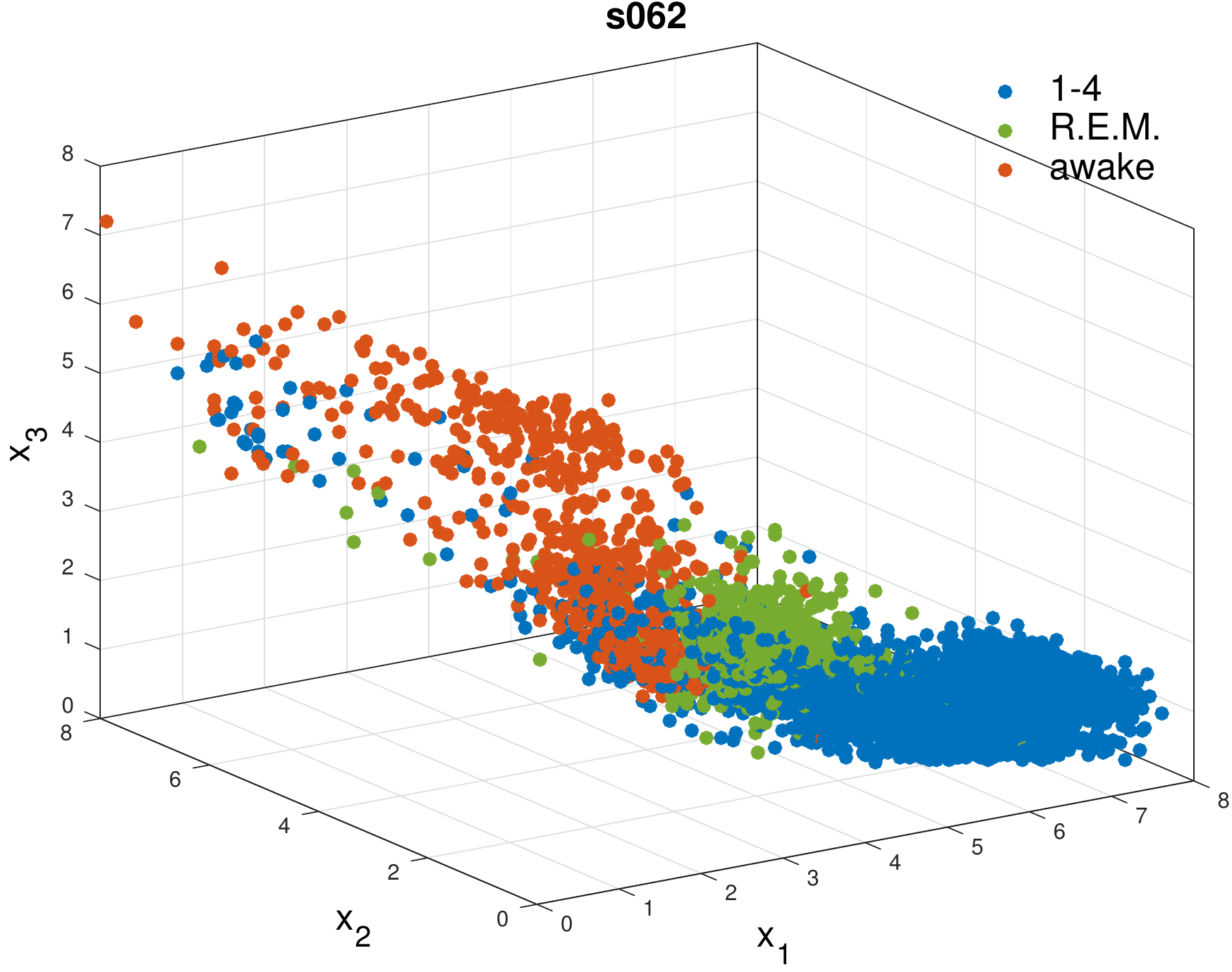}
	\end{subfigure}
	~
	\begin{subfigure}{0.31\textwidth}
		\includegraphics[width=\textwidth]{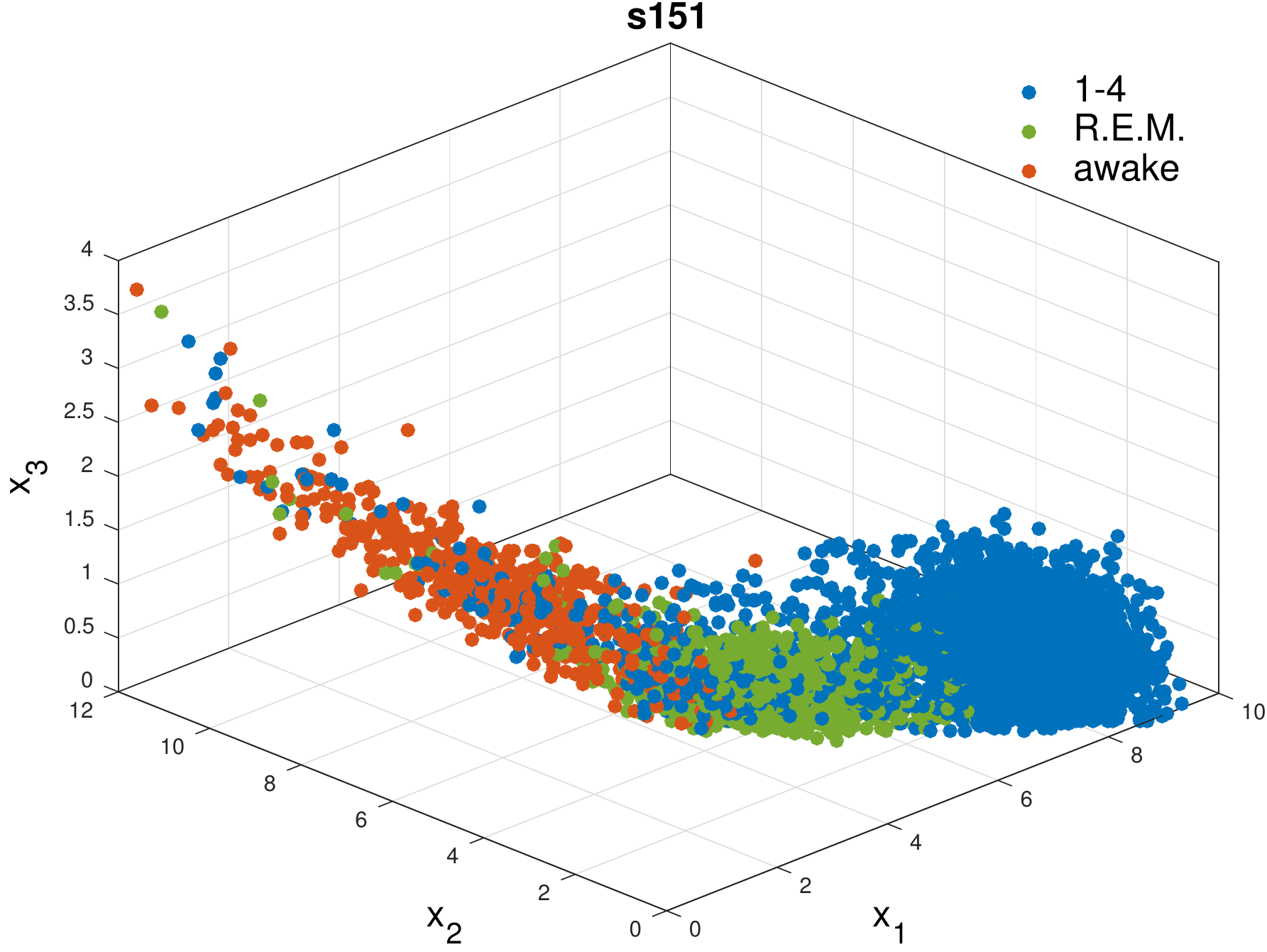}
	\end{subfigure}

	\begin{subfigure}{0.31\textwidth}
		\includegraphics[width=\textwidth]{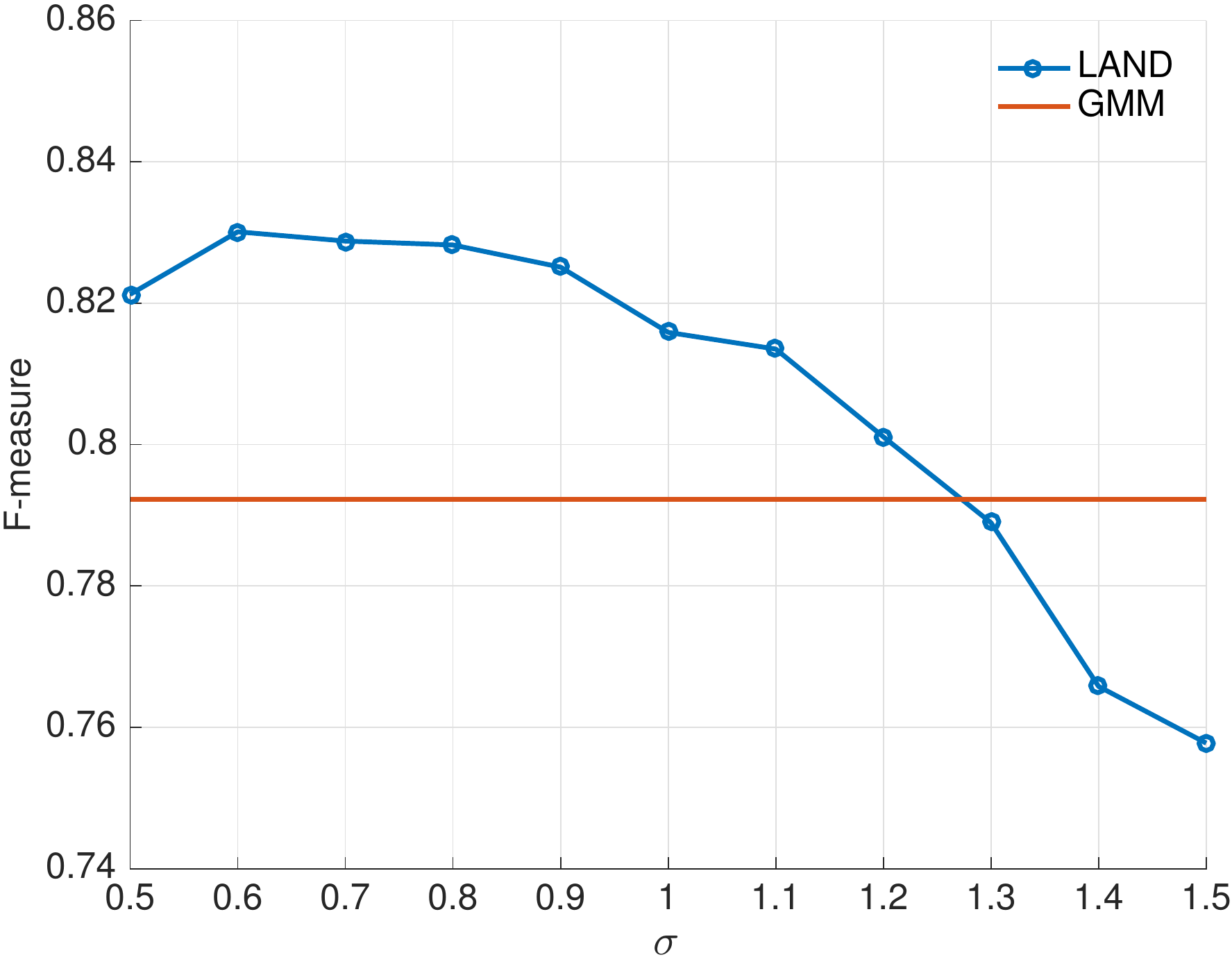}
	\end{subfigure}

\caption{\textit{Top row}: The given data, after the feature extraction procedure for 3 factors for three subjects. \textit{Bottom row}: the F-measure for different values of $\sigma$ (subject ``s151'').}
	\label{fig:sleep_features}
\end{figure}

Here we present the feature extraction result for 3 factors. From the Fig.~\ref{fig:sleep_features} we observe that actually, the derived data have a manifold structure. Moreover, we see that the characteristics of the data i.e., the EEG measurement, varies a lot between the 3 subjects.

\subsection{The Clustering Problem for the Synthetic Data}

Due to space limitations, we were not able to present the result of the least squares mixture model for the clustering problem of the synthetic data, thus, we present here the result. From the Fig.~\ref{fig:synthetic_2} we observe that indeed the LAND can approximates efficiently the underlying distributions of the clusters. Even thought the least squares mixture model takes under consideration the underlying structure of the data, it fails to reveal precisely the distributions of the clusters. Thus, we argue that our maximum likelihood estimates are better than the least squares estimates.  On the other hand, the GMM fails even to find the correct means of the distributions.

\begin{figure}[!ht]
\centering
    \begin{subfigure}[b]{0.31\textwidth}
        \includegraphics[width=\textwidth]{images/twomoons_land-eps-converted-to.pdf}
    \end{subfigure}
    \quad
    \begin{subfigure}[b]{0.31\textwidth}
        \includegraphics[width=\textwidth]{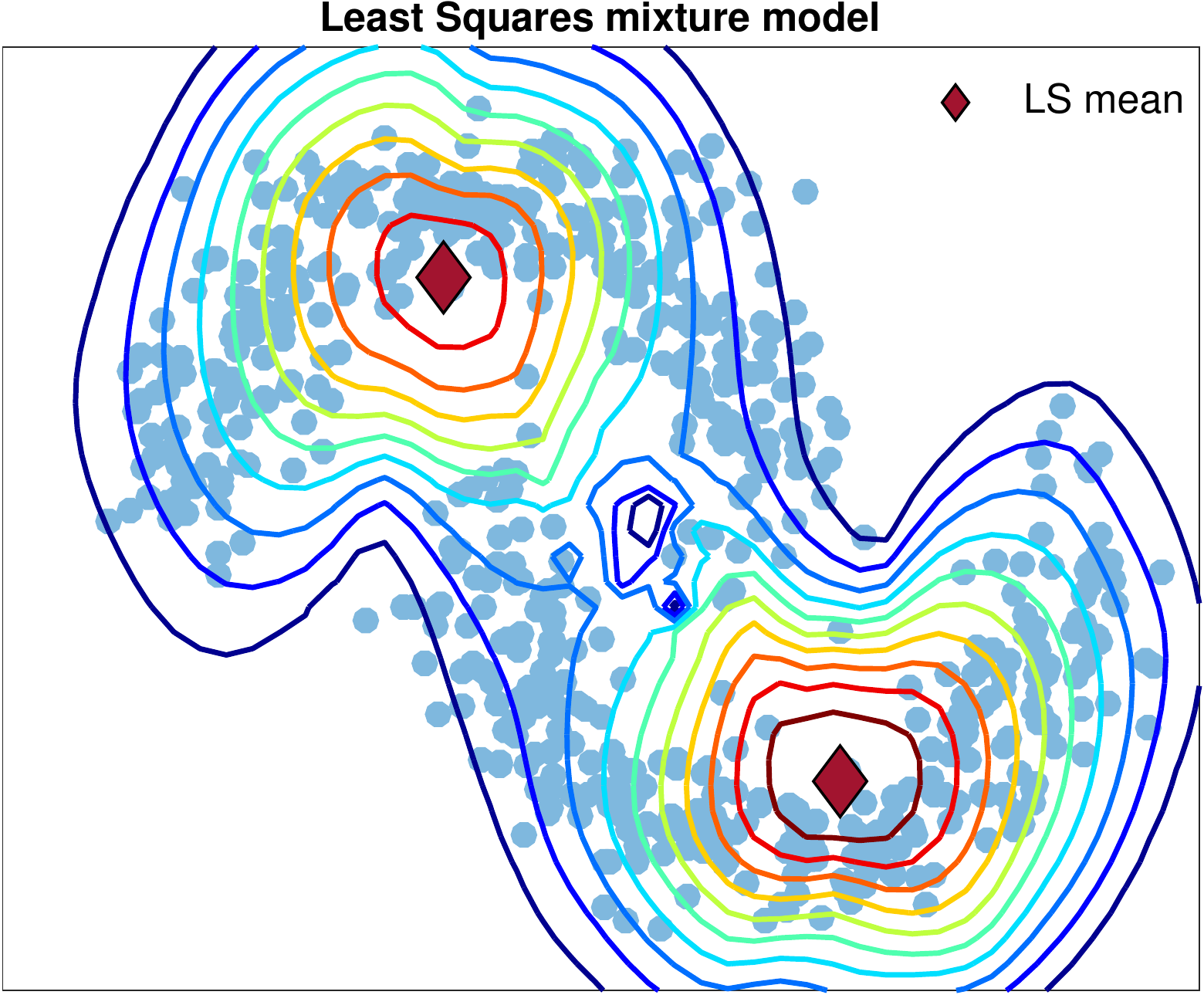}
    \end{subfigure}
    \quad    
    \begin{subfigure}[b]{0.31\textwidth}
        \includegraphics[width=\textwidth]{images/twomoons_linear-eps-converted-to.pdf}
    \end{subfigure}
    
    \vspace*{0.1cm}
    
    \begin{subfigure}[b]{0.31\textwidth}
        \includegraphics[width=\textwidth]{images/wave_arc_land-eps-converted-to.pdf}
    \end{subfigure}
    \quad
    \begin{subfigure}[b]{0.31\textwidth}
         \includegraphics[width=\textwidth]{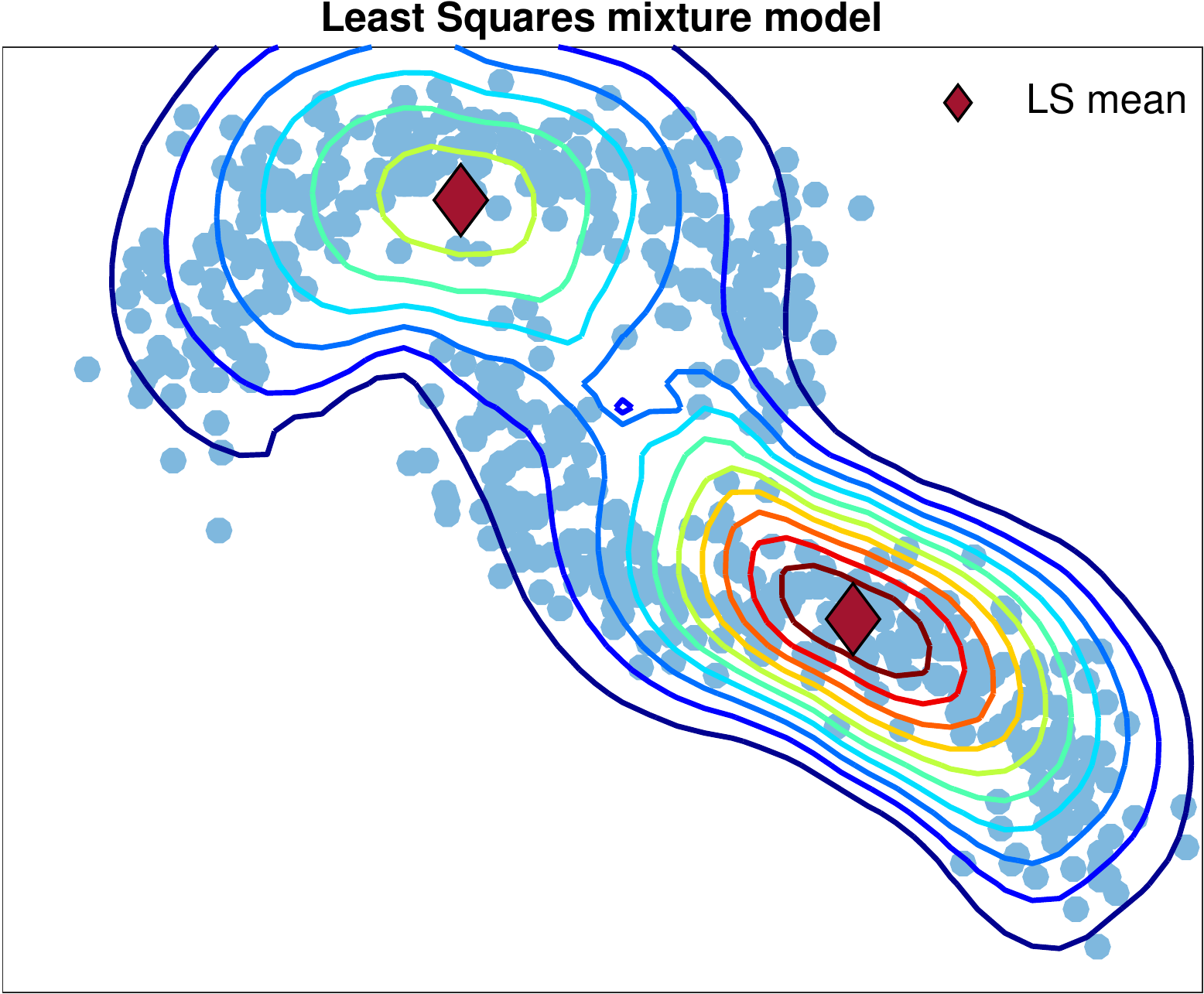}
    \end{subfigure}
    \quad
    \begin{subfigure}[b]{0.31\textwidth}
        \includegraphics[width=\textwidth]{images/wave_arc_linear-eps-converted-to.pdf}
    \end{subfigure}
    
    \caption{The clustering problem for two synthetic datasets. \textit{Left}: the LAND mixture model approximates efficiently the underlying distributions of the clusters. \textit{Center}: the least squares fails to reveal precisely the distributions of the clusters. \textit{Right}: the GMM due to the linear distance measure fails even to find the correct means of the distributions.}
\label{fig:synthetic_2}
\end{figure}

\subsection{The Contour Plots for the Synthetic Data}

\begin{figure}[!ht]
\centering
    \begin{subfigure}[b]{0.31\textwidth}
        \includegraphics[width=\textwidth]{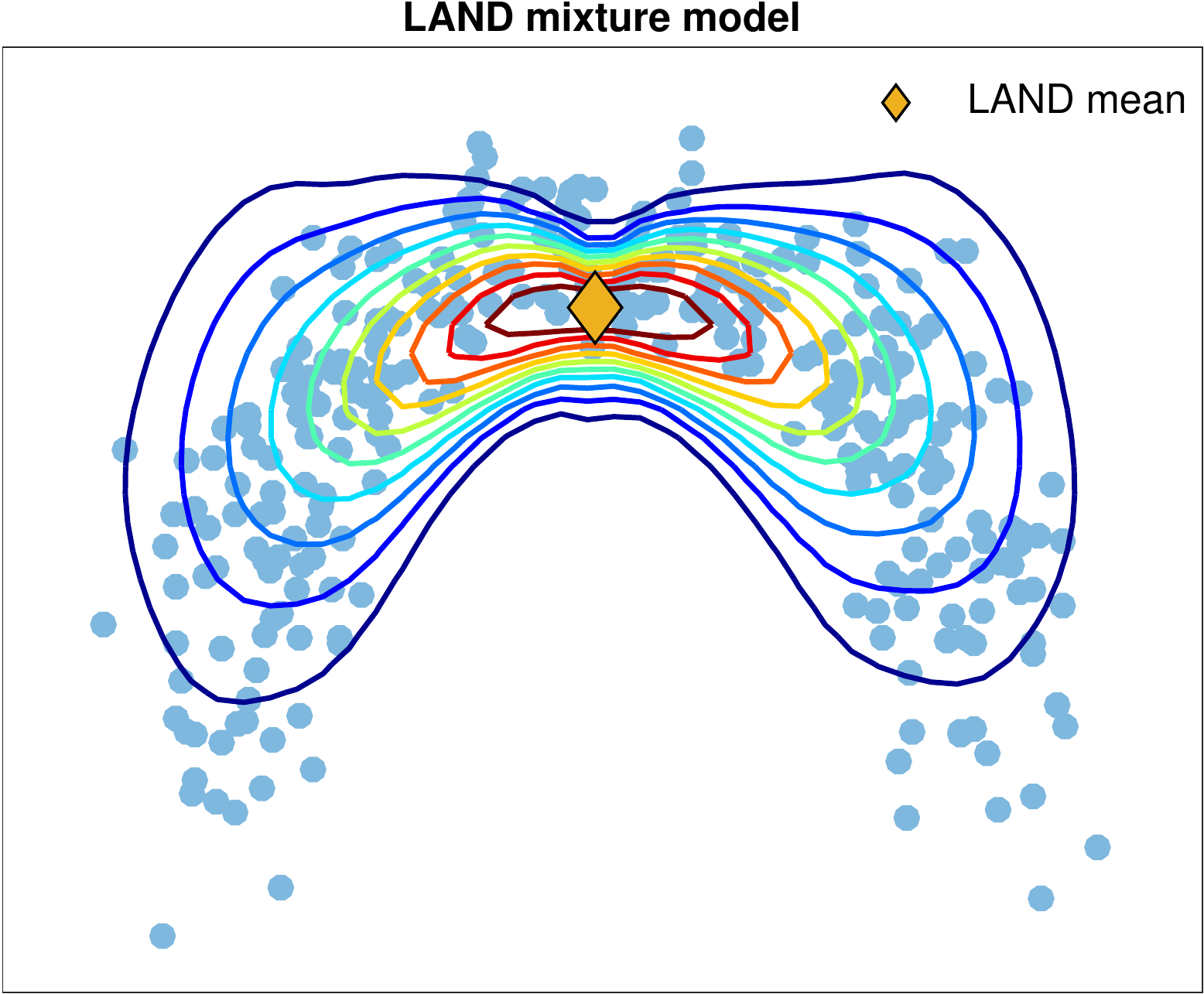}
    \end{subfigure}
    \quad
    \begin{subfigure}[b]{0.31\textwidth}
        \includegraphics[width=\textwidth]{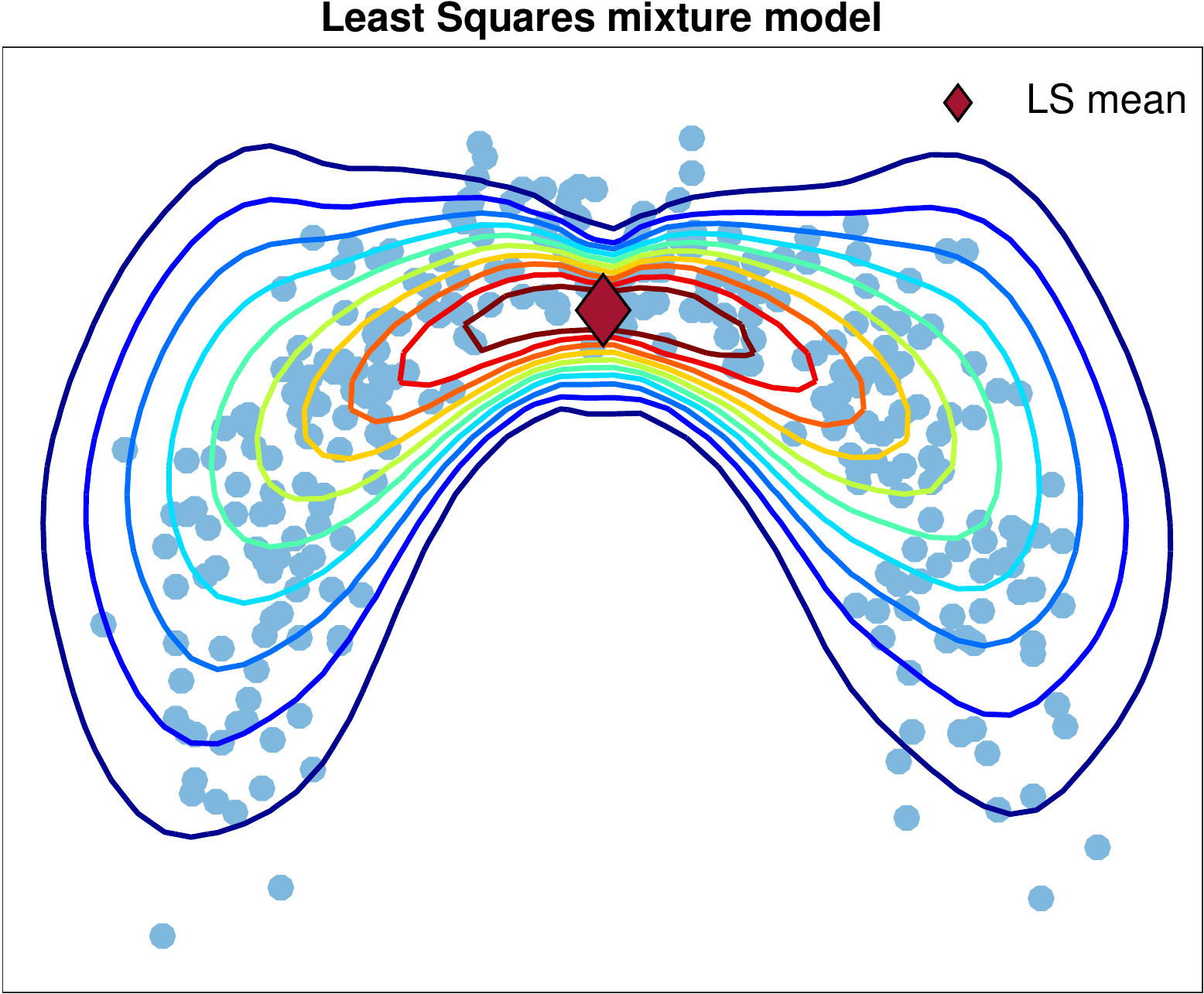}
    \end{subfigure}
    \quad    
    \begin{subfigure}[b]{0.31\textwidth}
        \includegraphics[width=\textwidth]{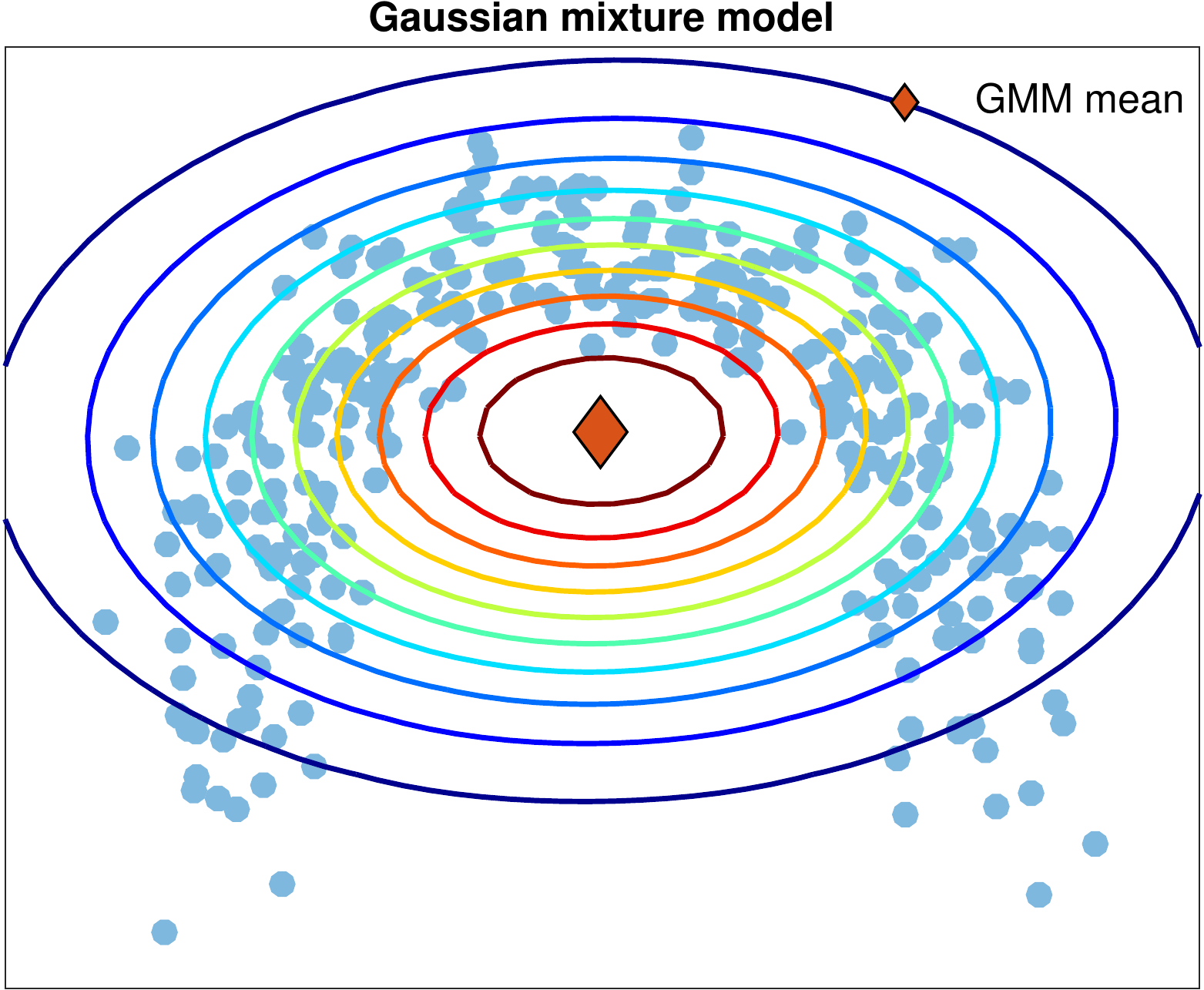}
    \end{subfigure}
    
	\vspace*{0.1cm}
    
    \begin{subfigure}[b]{0.31\textwidth}
        \includegraphics[width=\textwidth]{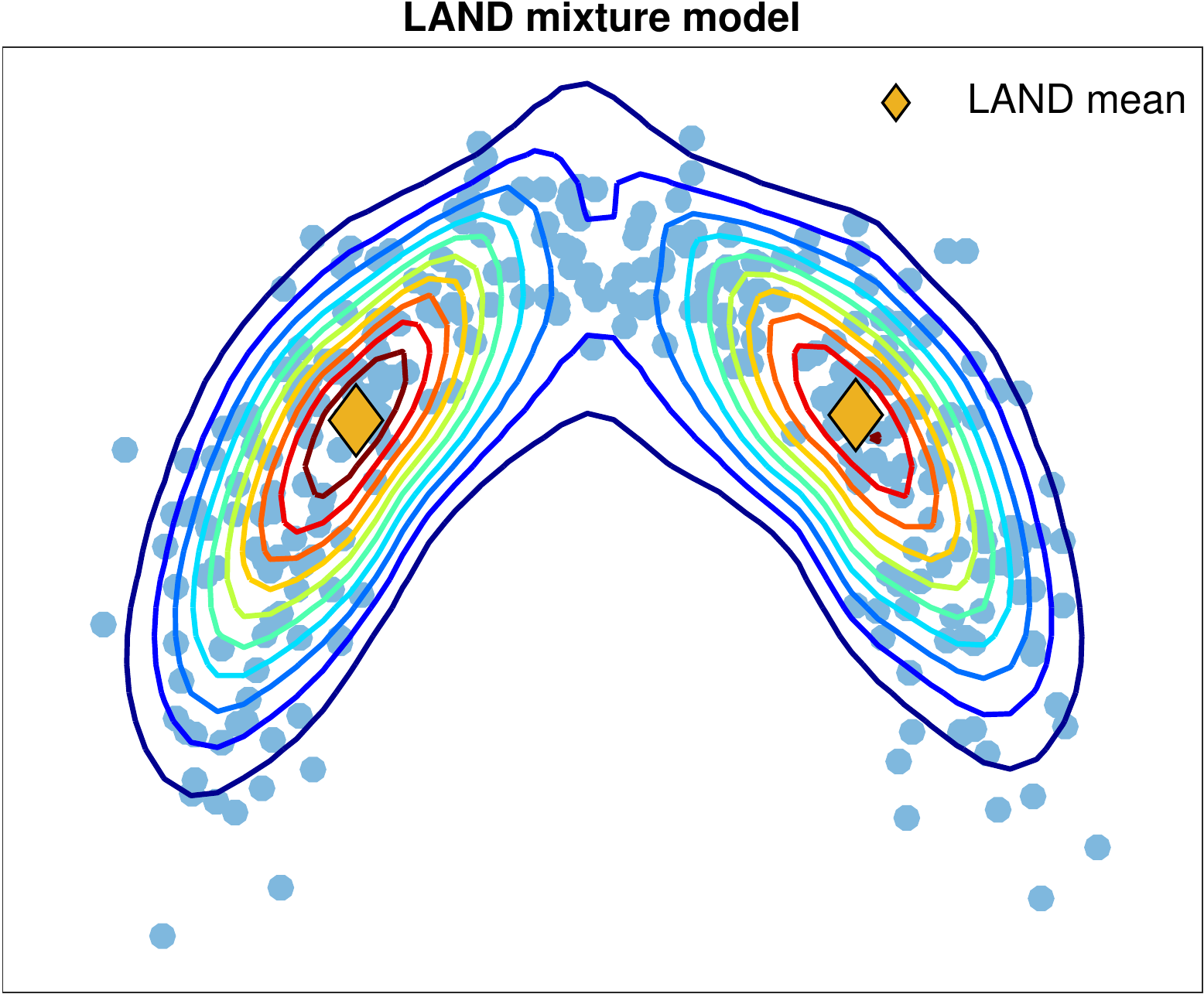}
    \end{subfigure}
    \quad
    \begin{subfigure}[b]{0.31\textwidth}
        \includegraphics[width=\textwidth]{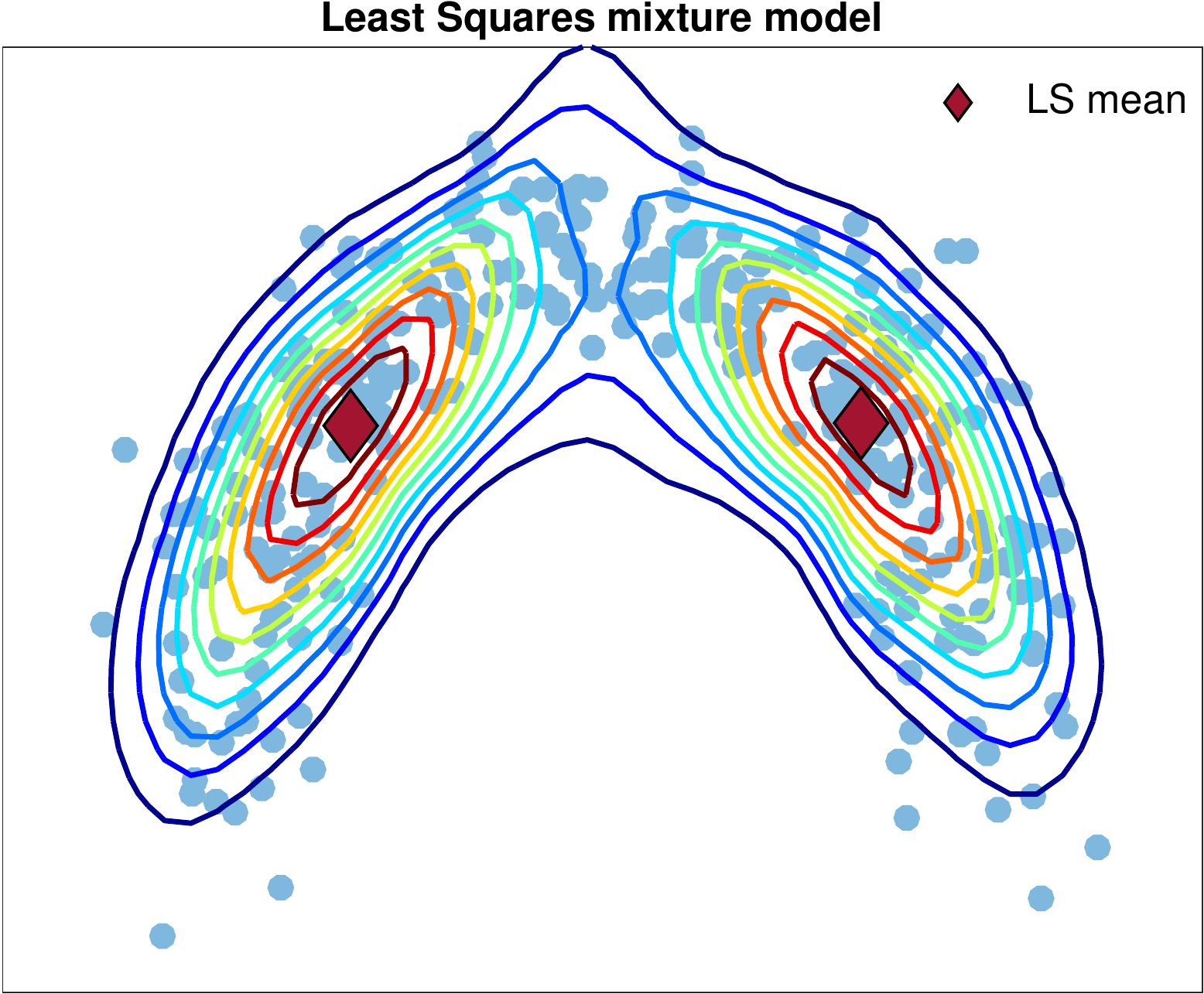}
    \end{subfigure}
    \quad    
    \begin{subfigure}[b]{0.31\textwidth}
        \includegraphics[width=\textwidth]{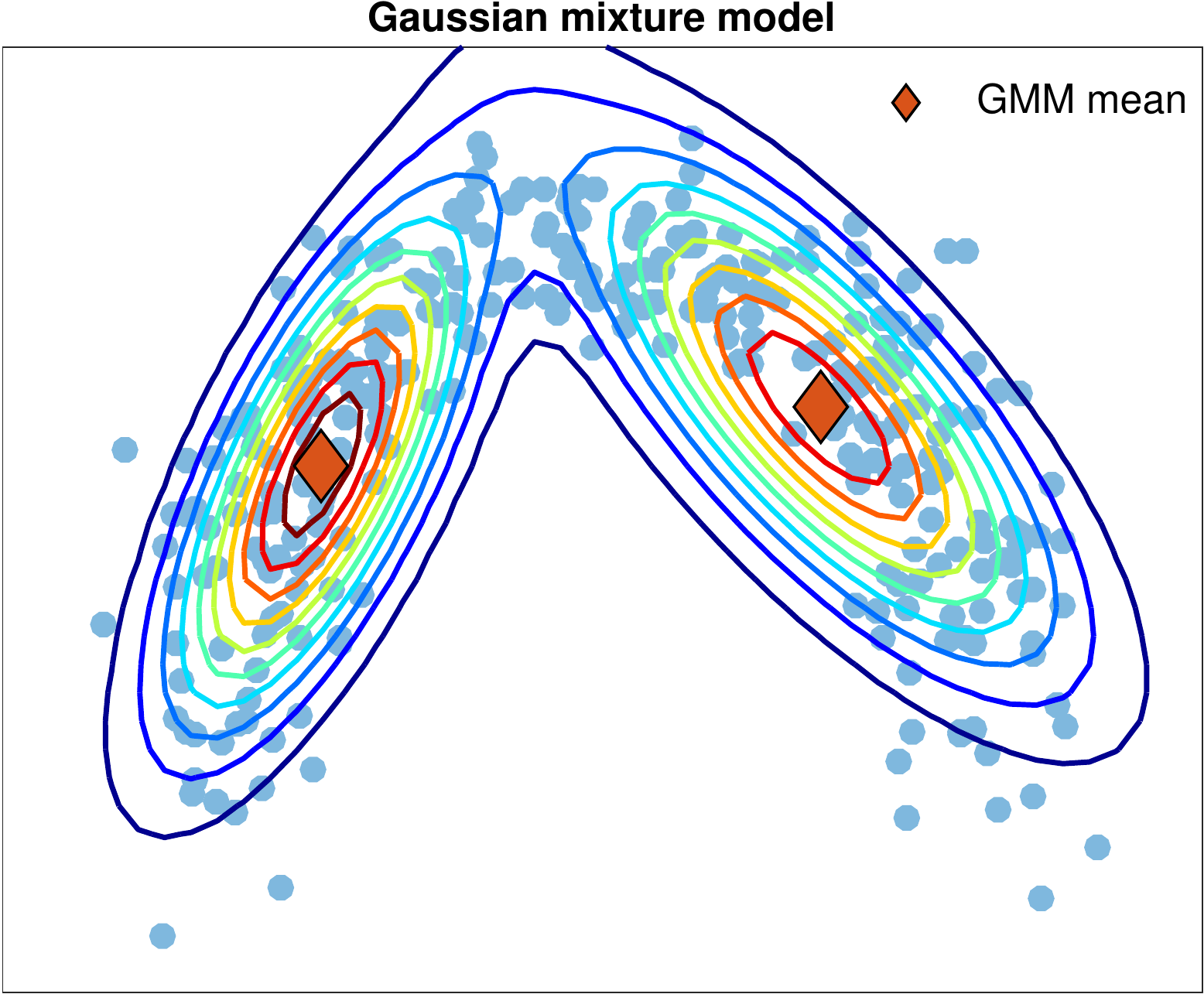}
    \end{subfigure}
    
	\vspace*{0.1cm}
    
    \begin{subfigure}[b]{0.31\textwidth}
        \includegraphics[width=\textwidth]{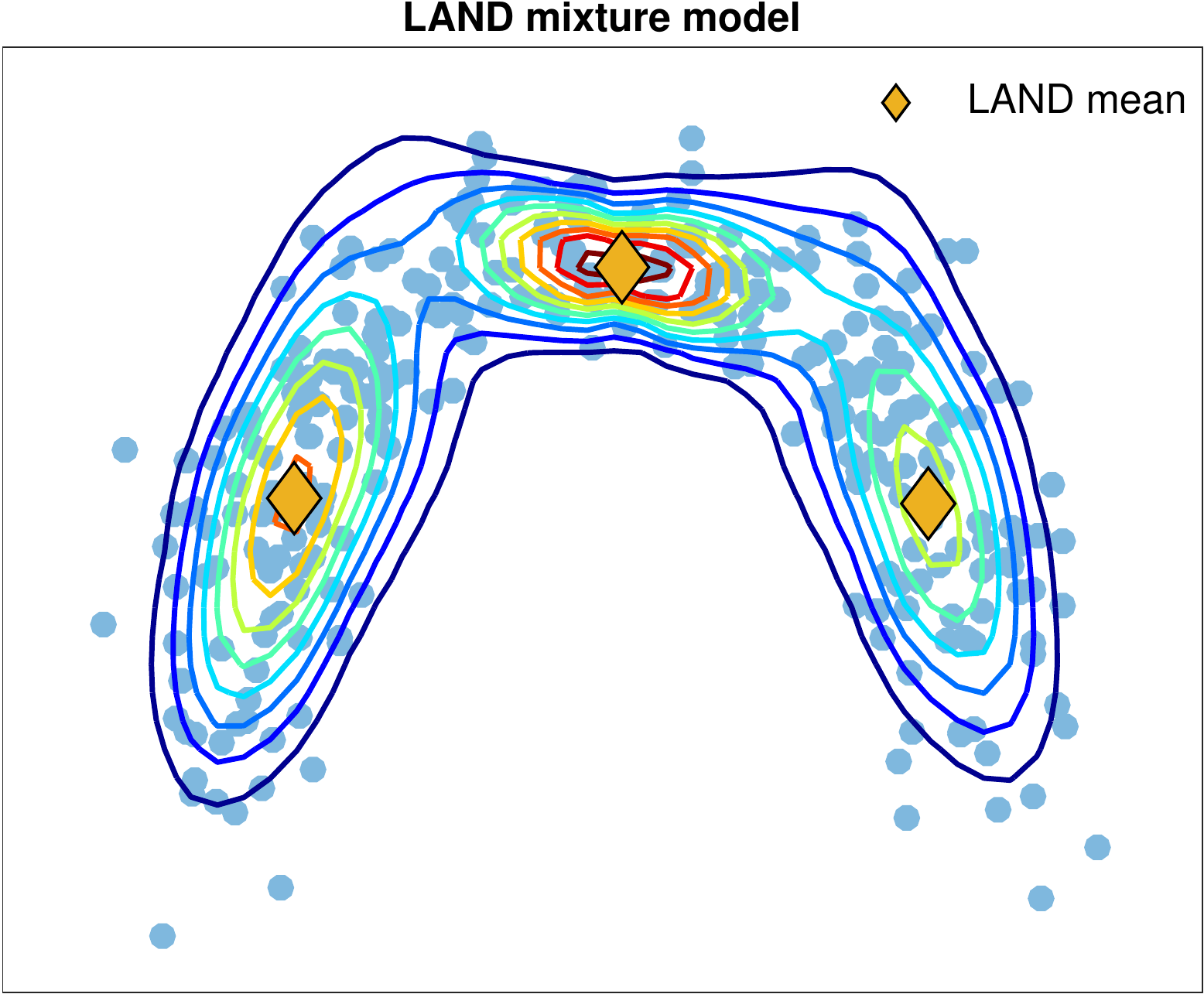}
    \end{subfigure}
    \quad
    \begin{subfigure}[b]{0.31\textwidth}
        \includegraphics[width=\textwidth]{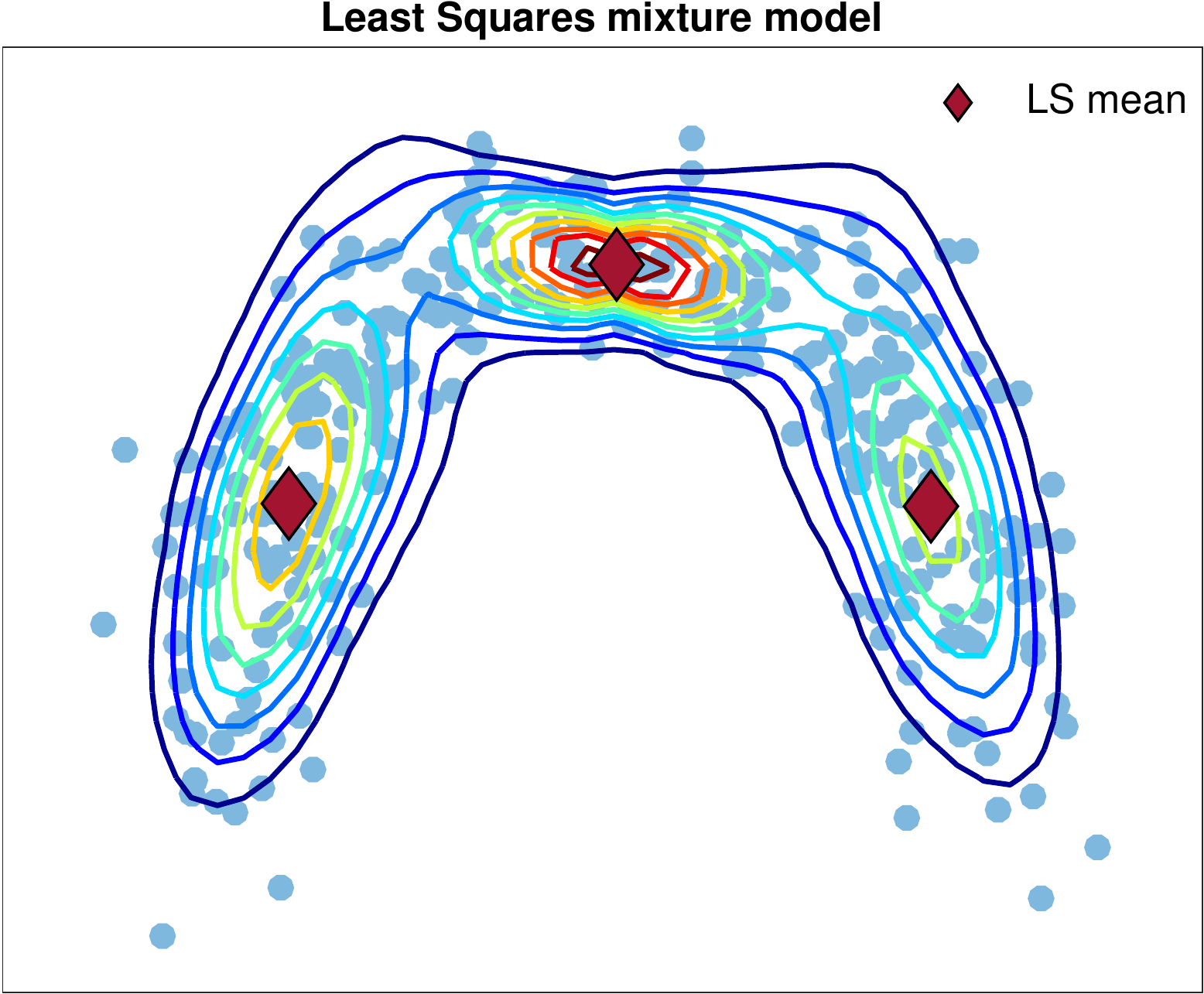}
    \end{subfigure}
    \quad    
    \begin{subfigure}[b]{0.31\textwidth}
        \includegraphics[width=\textwidth]{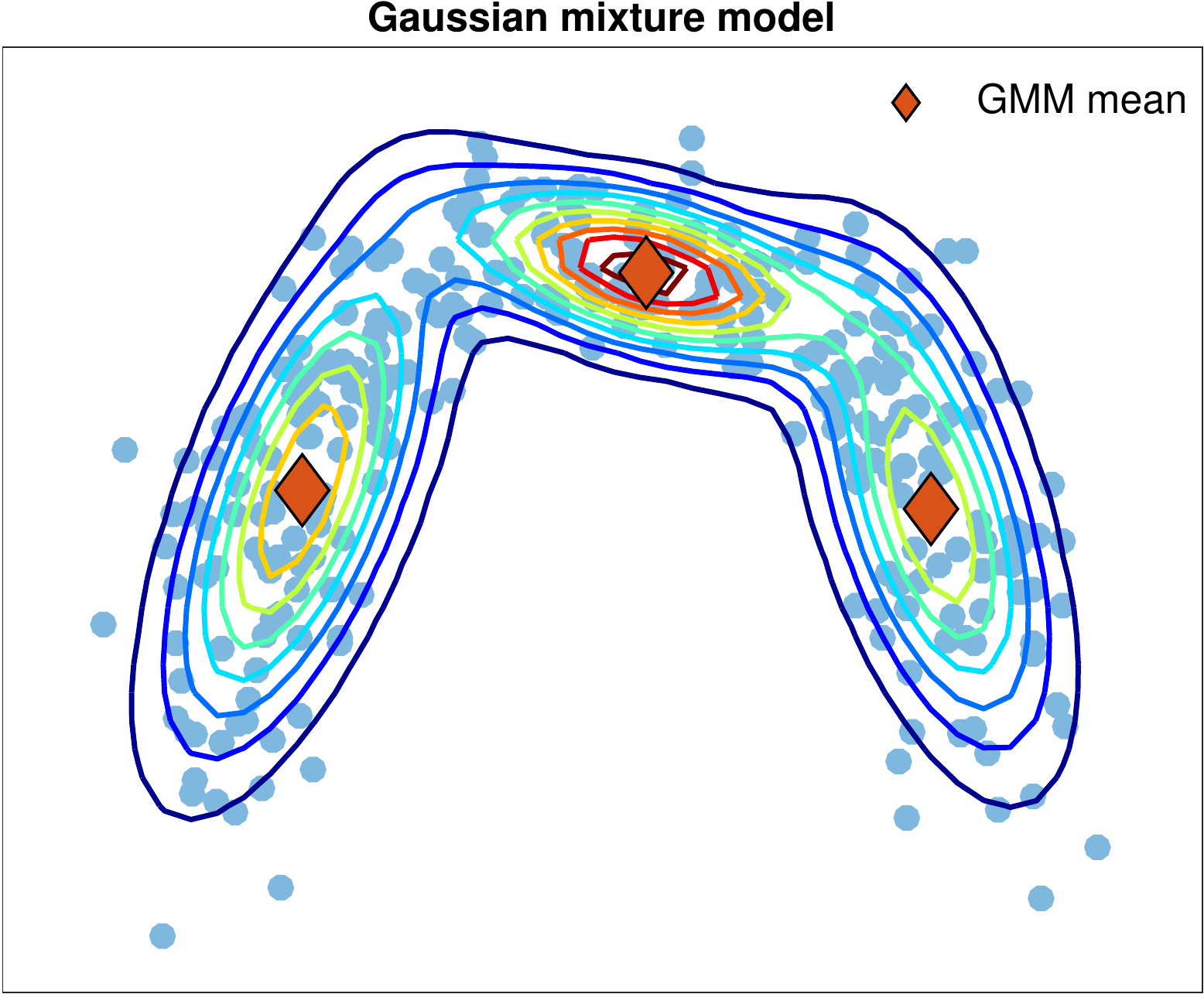}
    \end{subfigure}
    
	\vspace*{0.1cm}
	
    \begin{subfigure}[b]{0.31\textwidth}
        \includegraphics[width=\textwidth]{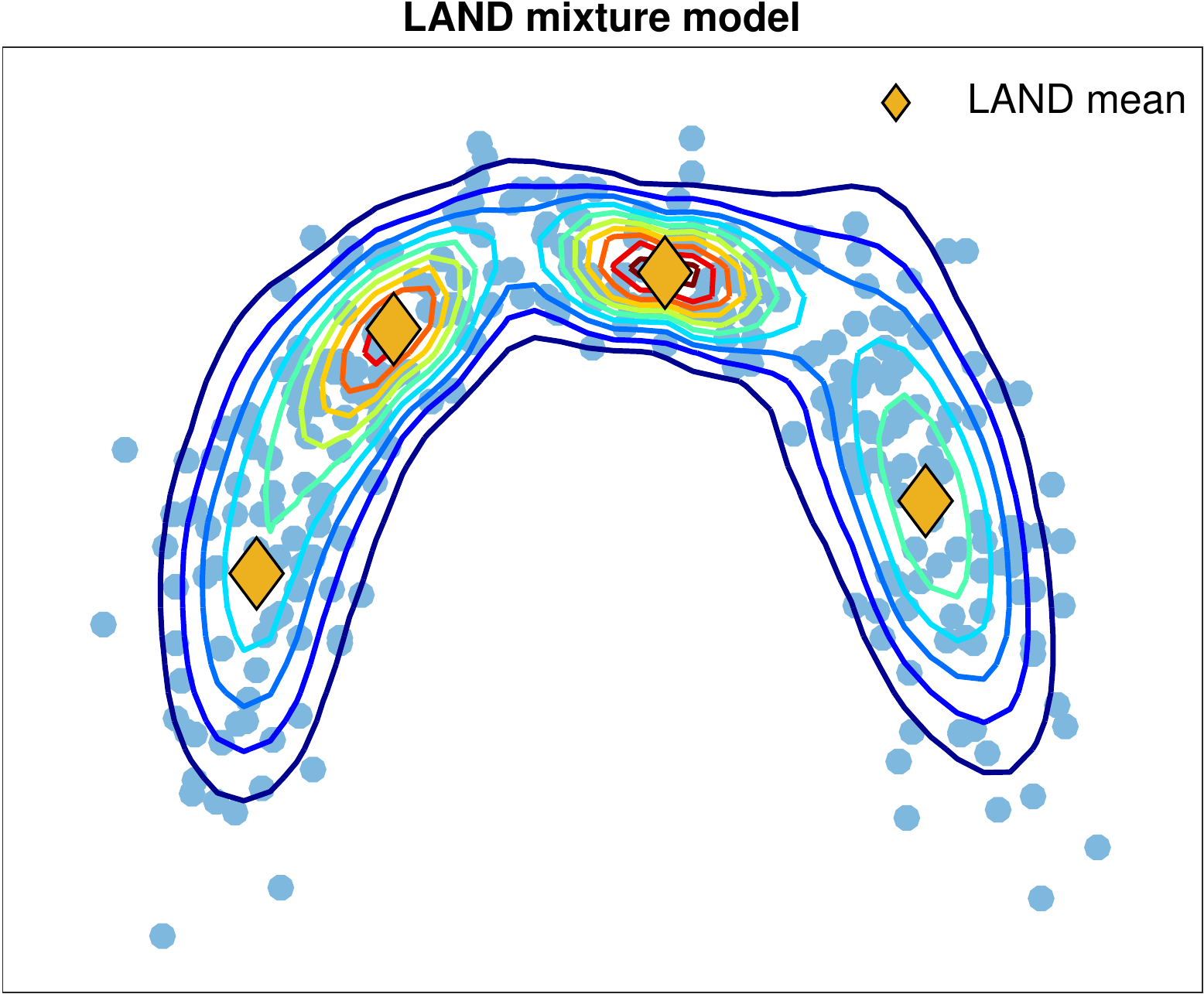}
    \end{subfigure}
    \quad
    \begin{subfigure}[b]{0.31\textwidth}
        \includegraphics[width=\textwidth]{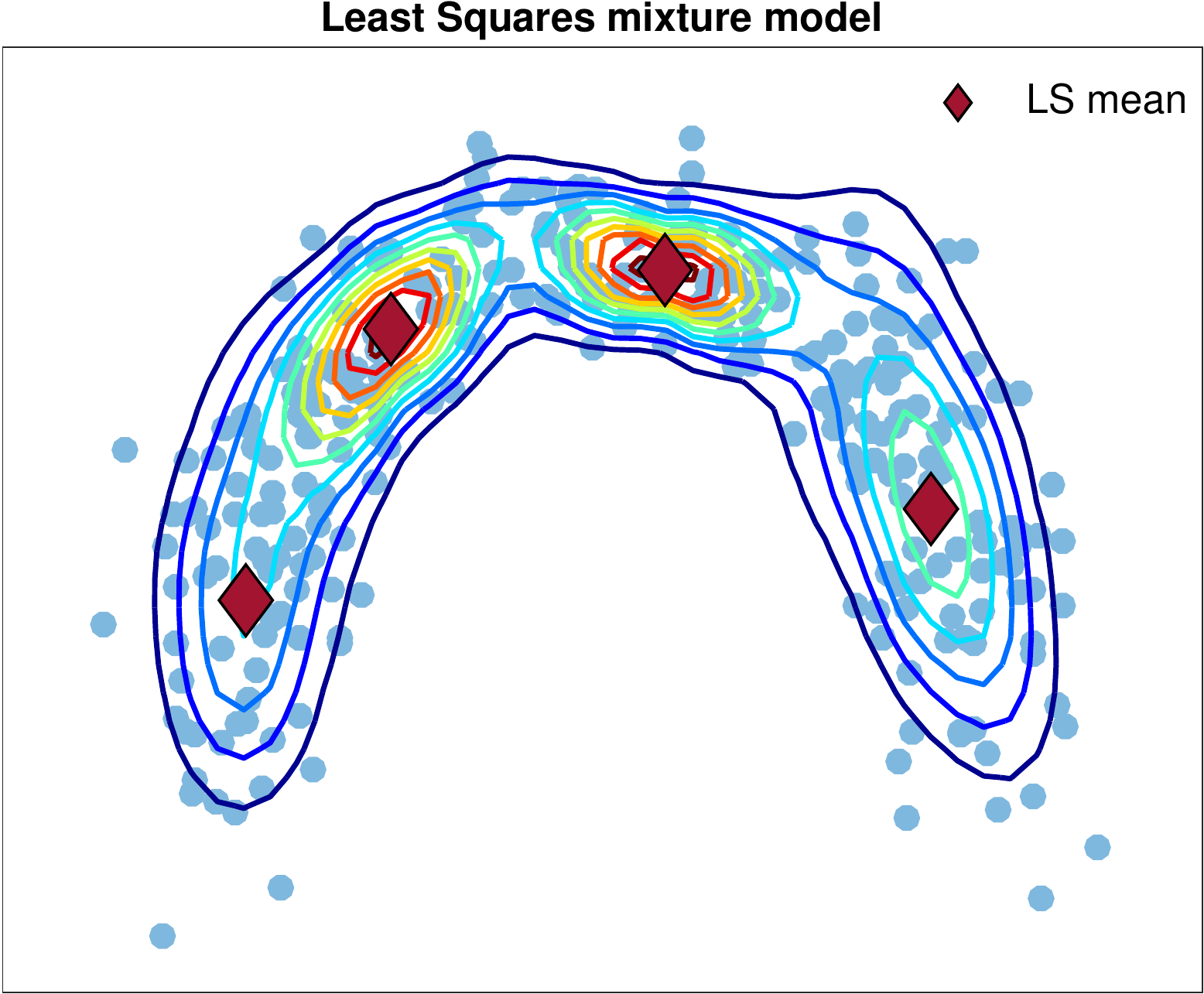}
    \end{subfigure}
    \quad    
    \begin{subfigure}[b]{0.31\textwidth}
        \includegraphics[width=\textwidth]{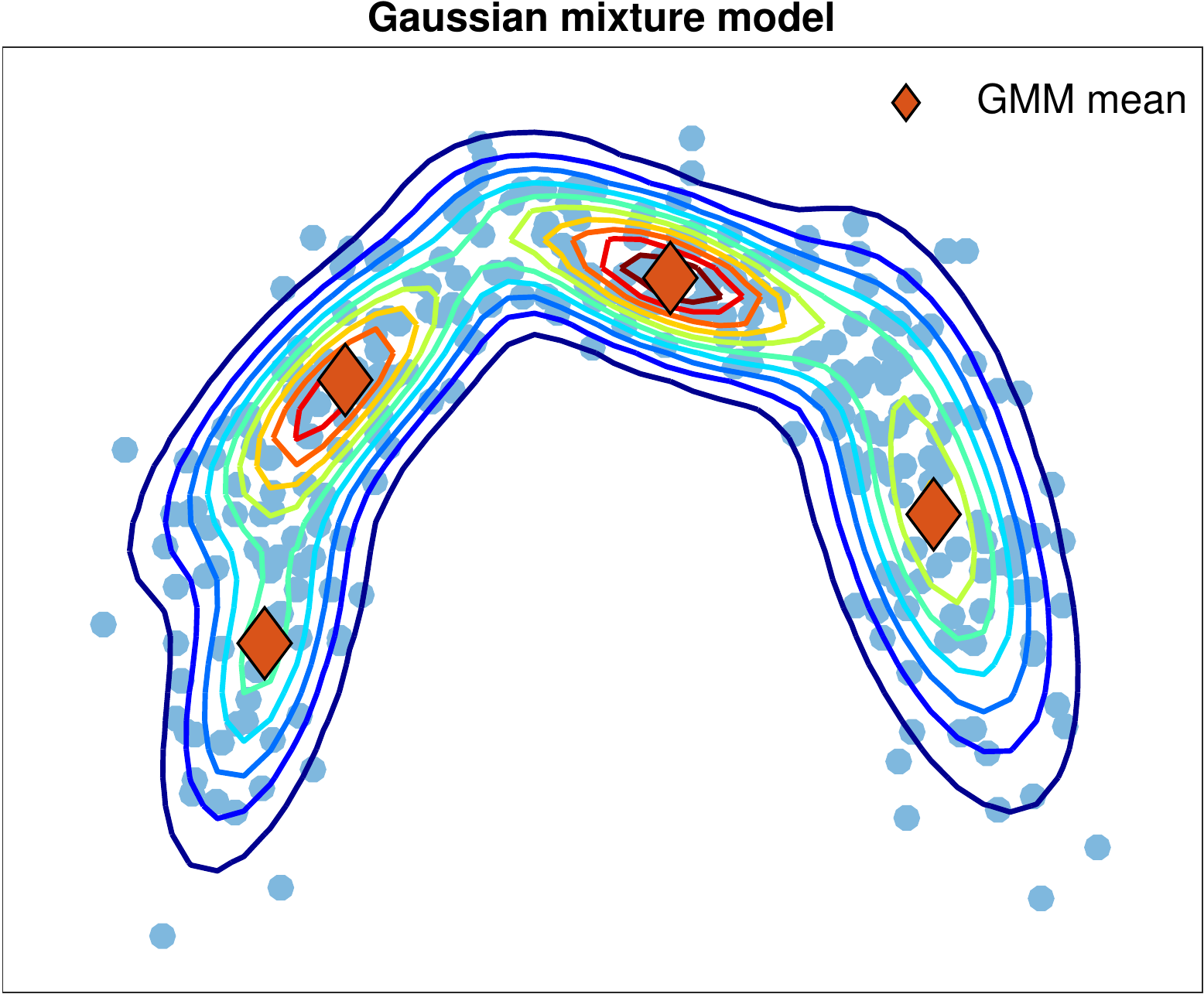}
    \end{subfigure}	
	
    \caption{Synthetic data and the fitted models. From top to bottom we present the results for $K=1,2,3,4$, respectively. \emph{Left}: the contours of the LAND mixture model. \emph{Center}: the contours of the least squares mixture model. \emph{Right}: the contours of the Gaussian mixture model.}
\label{fig:synthetic_1}
\end{figure}

Additionally to the results presented in the main paper, in Fig. \ref{fig:synthetic_1} we present the contours of all the fitted models and for all the numbers of components, where the advantages of the LAND are obvious. Especially, when $K=1$ we observe that the LAND approximates well the underlying distribution, while even though the least squares estimator reveals the nonlinearity of the distribution, as we discussed in the paper the covariance overfits the given data.

Furthermore, when $K$ increases the LAND components locally become almost linear Gaussians, since the geodesics will almost be straight lines. However, even in this case the LAND mixture model is more flexible than the Gaussian mixture model, see the result for $K=4$. Also, the LAND does not overfit the given data, as the least squares mixture model does, since the probability mass is more concentrated around the means, see the result for $K=2$.

\subsection{Motion Capture Data}

\begin{figure}[!ht]
    \centering
    \begin{subfigure}{0.31\textwidth}
        \includegraphics[width=\textwidth]{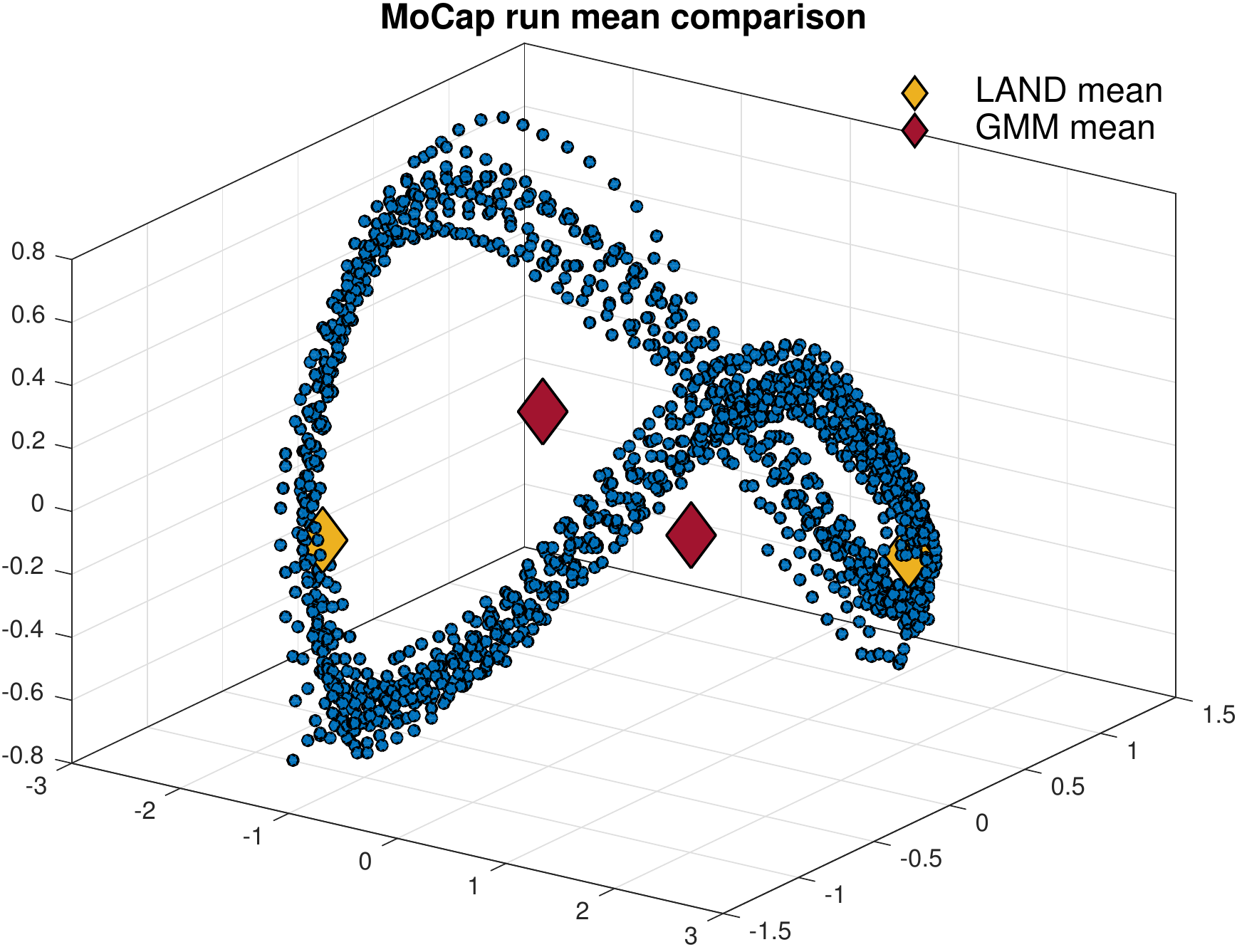}
    \end{subfigure}
	\quad
    \begin{subfigure}{0.31\textwidth}
        \includegraphics[width=\textwidth]{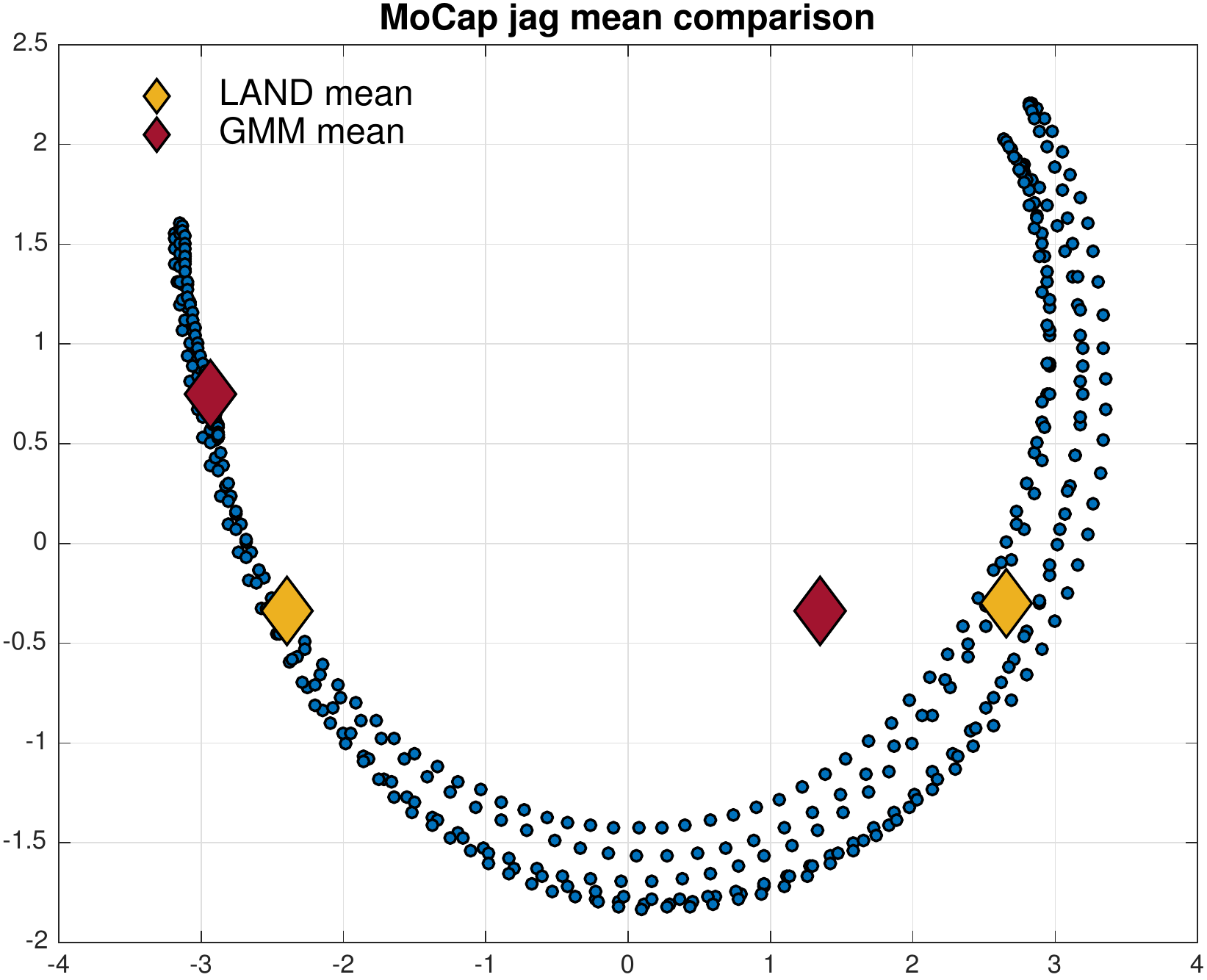}
    \end{subfigure}
    \caption{Motion capture experiment. \textit{Left}: the experiment for the run sequence. \textit{Right}: the experiment for the jumping jag sequence.}
    \label{fig:mocap_result}
\end{figure}

We conducted an experiment using motion capture data from \emph{CMU Motion Capture Database}\footnote{http://mocap.cs.cmu.edu/}. Specifically, we picked two movements motion: 16 from subject 22 (jumping jag), and the subject 9 (run). Each data point corresponds to a human pose. We projected the data onto the first 2 and 3 principal and we fitted a LAND mixture model and a Gaussian mixture model for $K=2$. From the results in Fig.~\ref{fig:mocap_result} we see that the LAND means fall inside the data, while the GMM means are actually outside of the manifold.

\subsection{Scalability of Geodesic Computations}

\begin{figure}[!ht]
    \centering
        \includegraphics[width=0.31\textwidth]{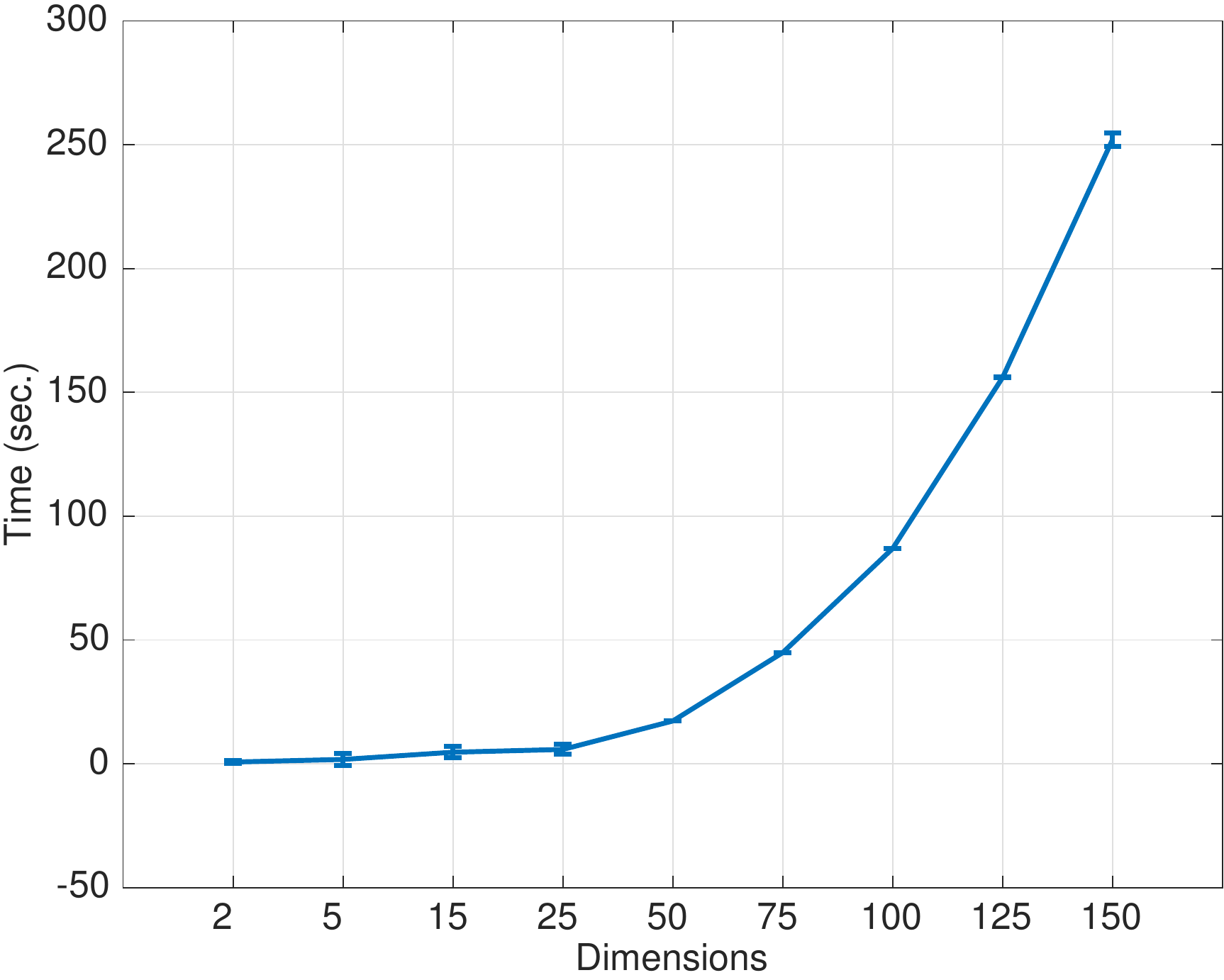}
        
    \caption{Scalability experiment.}
    \label{fig:scalability}
\end{figure}

A scalability concern is that the underlying ODEs are computationally
more demanding in high dimensions, and more specifically, we are interested in the logarithm map. We conducted a supplementary experiment on the MNIST
data, reporting the ODE solver running time as a function of input
dimensionality. In particular, we fix a point and we compute the running time of the logarithm map between this point and 20 random chosen points, for a set of the dimensions of the feature space. From the result in Fig. \ref{fig:scalability} we observe that the current implementation scales to approximately $50$ dimensions, where it becomes impractical.

\subsection{Model Selection}

We used the standard AIC and BIC criteria, 
\begin{align}
&BIC = -2\cdot \ln( L ) + \nu \cdot \ln(N)\\
&AIC = -2\cdot \ln( L ) + 2 \cdot \nu
\end{align}
where $L\in\R$ is the log-likelihood of the model, and $\nu \in \R$ is the number of free parameters. The optimal number of components $K$ can then be chosen to minimize either criteria. Note that the LAND and the GMM are not normalized under the same measure, so their likelihoods are not directly comparable. However, we can select the optimal $K$ for each method separately.

\begin{figure}[!ht]
    \centering
    \begin{subfigure}{0.31\textwidth}
        \includegraphics[width=\textwidth]{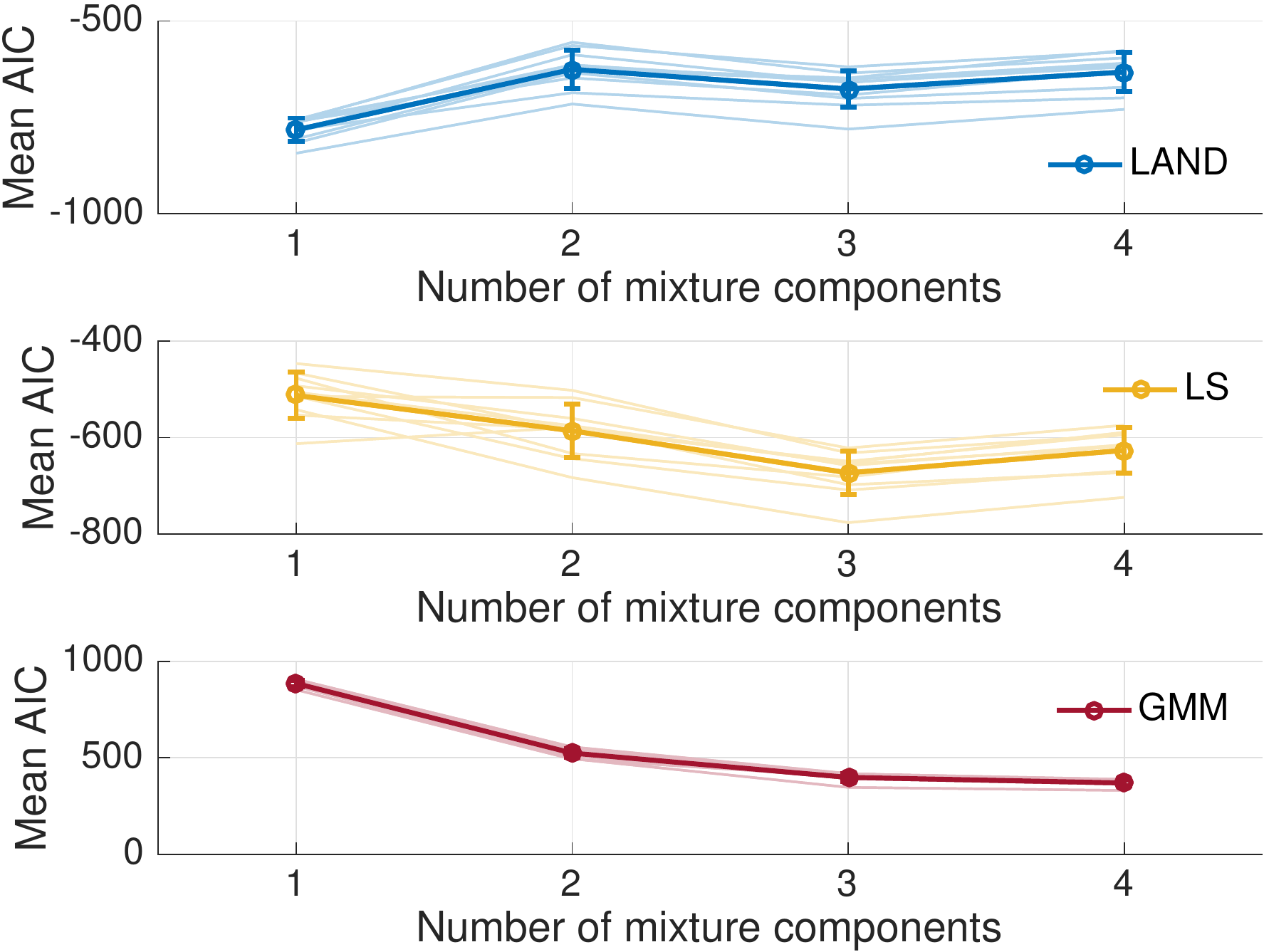}
    \end{subfigure}
	\quad
    \begin{subfigure}{0.31\textwidth}
        \includegraphics[width=\textwidth]{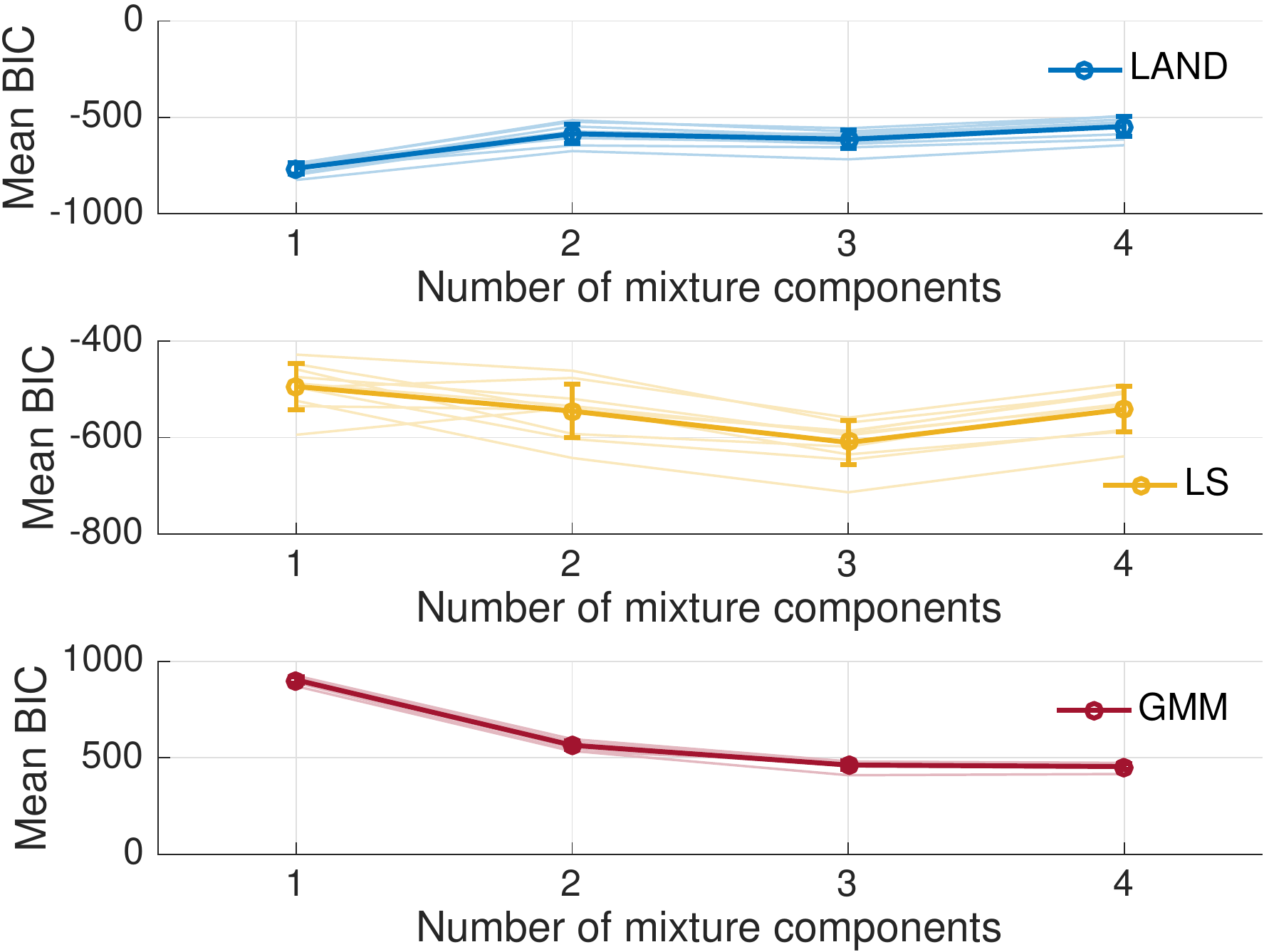}
    \end{subfigure}
    \caption{Model selection experiment. \textit{Left}: the AIC criterion. \textit{Right}: the BIC criterion.}
    \label{fig:aic_bic}
\end{figure}

We used the synthetic data from the first experiment in the paper. From the results in Fig. \ref{fig:aic_bic} we observe that the optimal LAND model is achieved for $K=1$, while the for the least squares estimators and the GMM, the optimal is achieved for $K=3$ and $K=4$ respectively. Thus, we argue that the less complex LAND model with only one component, is able to reveal the underlying distribution, while the other two methods need more components resulting to more complex models.

\end{document}